\documentclass[11pt]{article}

\usepackage{hyperref}
\usepackage{url}
\usepackage{subcaption}
\usepackage{pgffor}
\usepackage{float} 
\usepackage[dvipsnames]{xcolor}
\usepackage{titletoc}
\usepackage{etoc}

\usepackage{booktabs}
\usepackage{tabularx}
\usepackage{xcolor}
\definecolor{copyred}{HTML}{A11226}
\definecolor{oursblue}{HTML}{1B4F91}
\newcommand{\copiedcolor}[1]{\textcolor{copyred}{\textit{#1}}}
\newcommand{\ourscolor}[1]{\textcolor{oursblue}{#1}}
\usepackage{fontspec}
\newfontfamily\tamfont{NotoSansTamil-Regular.ttf}[
  Path = ./fonts/,
  Script = Tamil,
  BoldFont = NotoSansTamil-Bold.ttf
]
\newcommand{\ta}[1]{{\tamfont #1}}

\makeatletter
\let\SAVEDaddcontentsline\addcontentsline
\makeatother
\ifdefined\revon
    \newcommand{\new}[1]{\textcolor{blue}{#1}}
    \newcommand{\old}[1]{\textcolor{red}{\sout{#1}}}
    \usepackage[normalem]{ulem}
\else
    \newcommand{\new}[1]{#1}
    \newcommand{\old}[1]{}
\fi

\newcommand{\mytitle}{Contrast Matters: Understanding Robustness of In-Context Fine-Tuning to Target-Context Relatedness}

\usepackage[final]{acl}

\usepackage{times}
\usepackage{latexsym}

\usepackage[T1]{fontenc}

\usepackage[utf8]{inputenc}

\usepackage{microtype}

\usepackage{inconsolata}

\usepackage{graphicx}
\usepackage{algorithm}
\usepackage{algpseudocode}

\usepackage{latexsym}
\usepackage{amsmath}
\usepackage{dsfont}
\usepackage{amssymb}
\usepackage{wrapfig}
\usepackage{mathtools}
\usepackage{subcaption}
\usepackage{fvextra}
\DefineVerbatimEnvironment{wrapverb}{Verbatim}{breaklines=true}
\usepackage{multirow}
\usepackage{defs}
\RequirePackage[inline,shortlabels]{enumitem}
\setlist{nosep}

\title{\mytitle}

\usepackage[disable]{todonotes}

\author{
 \textbf{Deeptanshu Malu\textsuperscript{1*}} \quad
 \textbf{Deevyanshu Malu\textsuperscript{1*}} \quad
 \textbf{Aditya Nemiwal\textsuperscript{1}} \quad
 \textbf{Sunita Sarawagi\textsuperscript{1}}
\\
\textsuperscript{*}Equal Contribution \quad \textsuperscript{1}IIT Bombay
\\
 \texttt{\{deeptanshumalu, deevyanshumalu\}@cse.iitb.ac.in}
}

\newcommand{\fsft}{IC-Train}
\newcommand{\vxt}{\vx_*}
\newcommand{\vyt}{\vy_*}

\newcommand{\fsrand}{Random-Context}
\newcommand{\fssim}{Similar-Context}
\newcommand{\zs}{Zero-Context}
\newcommand{\fsour}{\ours}

\newcommand{\fsse}{\fssim}

\newcommand{\ours}{Contrastive-Context}
\renewcommand{\cM}{P_\theta}
\renewcommand{\choose}{\texttt{Choose}}

\newcommand{\simS}{\text{sim}}
\newcommand{\vyh}{\hat{\vy}}
\newcommand{\copyscore}{\text{Copy-score}}
\newcommand{\iclscore}{\text{ICL-score}}
\newcommand{\iwlscore}{\text{IWL-score}}

\begin{document}
\maketitle

\begin{abstract}
In-context fine-tuning (IC-Train), training an LLM with labeled examples in-context, is increasingly used in place of standard fine-tuning for domain adaptation and continual absorption of labeled data. We study the robustness of the in-context learning ability that emerges from such training: does the fine-tuned model perform well across test inputs whose in-context examples range from unrelated to nearly identical? Across 32 configurations spanning four open-source LLMs and eight test sets over machine translation, Text-to-SQL, and multilingual semantic parsing, we show that robustness hinges on an overlooked design choice: how in-context examples are selected relative to the target during training.
The two prevailing strategies turn out to be accurate over complementary parts of this spectrum: random contexts yield a model that gains little from related examples even when they are placed in its context, while retrieved similar contexts weaken accuracy on targets lacking close neighbors and raise the propensity to copy labels from context.
Probes tracking in-weights learning, in-context learning, and copying trace these failures to distinct training dynamics, and show that introducing contrast in target-context similarity both within a context and across batches, restores robustness across the entire spectrum.

\end{abstract}

\begin{figure}
    \centering
    \includegraphics[width=0.9\linewidth]{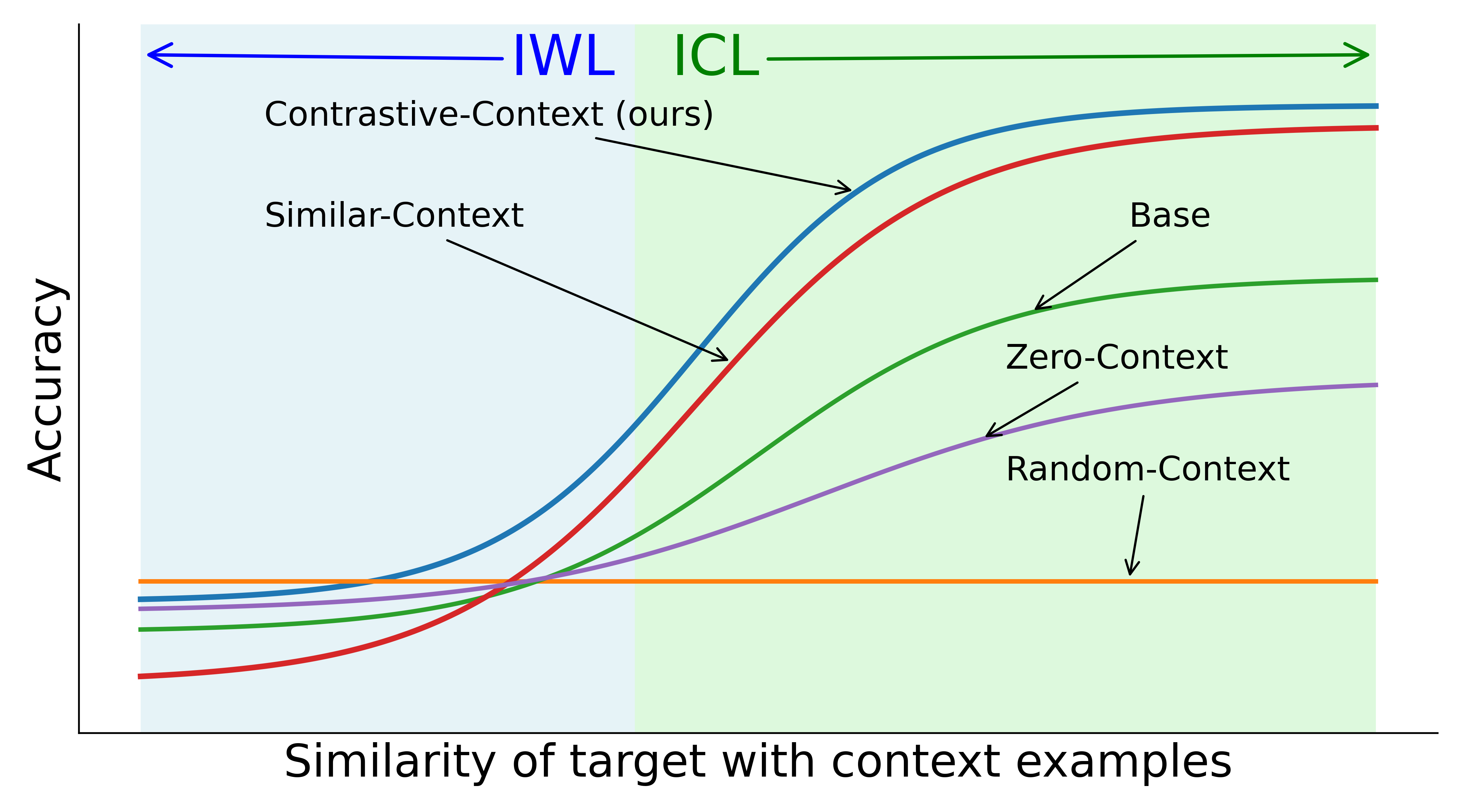}
    \caption{Schematic summary of main findings of the paper. On the X-axis is target to context similarity --- IWL important on left side and ICL on right side. 
    Fine-tuning with in-context example (\fsft) is sensitive to similarity of target to context examples chosen during training: random context leads to sharp drop in ICL, similar context does not develop IWL and instead is prone to blind copying, contrast in the context recovers both limitations.}
    \label{fig:placeholder}
\end{figure}
\section{Introduction}

Our goal is to fine-tune an LLM on a task with limited labeled data, while retaining the ability to continuously incorporate feedback in the form of new examples at test time, without further training.
Such adaptation should boost accuracy on inputs resembling the new examples, while retaining generalization to inputs without close neighbors. In-context learning (ICL) offers a natural mechanism for flexibly absorbing labeled data by placing test inputs alongside labeled examples. Pre-trained LLMs~\cite{brownicl} show strong ICL ability, yet prior studies~\cite{alves-etal-2023-steering,duan2024exploringICLandIT,wang2024losscontext,goyal2025contextparametric}, as well as our experiments, find that fine-tuning often diminishes it. A natural alternative is to fine-tune in ICL mode (IC-Train)~\cite{min-etal-2022-metaicl,alves-etal-2023-steering,bornschein2025finetuned,Zheng2024FinetuningLL,zebaze-etal-2025-context}, where the model is trained to predict a target given a few labeled examples in its context.

IC-Train is thus positioned to meet both goals at once: a model adapted to the task, and one that still knows what to do with a related example when it appears in context.  In this paper, we show that a critical but overlooked factor that determines how far IC-Train achieves these goals is how the in-context examples are selected relative to the target during training.  Most prior work treated example selection as implementation details, and sampled them at random and independent of the target~\cite{alves-etal-2023-steering,bornschein2025finetuned}, including in multi-task meta-training settings~\cite{min-etal-2022-metaicl}, a strategy we call Random-Context. A more recent line retrieves the top-k most similar examples for each target~\cite{zebaze-etal-2025-context,Zheng2024FinetuningLL}, which we call Similar-Context.  Prior studies compared these methods on aggregate accuracy, and showed modest differences.  

We study the robustness of a model trained with \fsft\ under varying grades of target-context relatedness, to cover the continuous spectrum of useful context available to a random test example during deployment.  Through several context distributions, across 32 (model, language-pair, test-set) configurations for machine translation, eleven Text-to-SQL databases, and three languages of multilingual semantic parsing,  we show that the two prevailing strategies are each accurate over only part of the spectrum, and they cover complementary parts of it. 
\begin{enumerate}[leftmargin=*]
    \item Random-Context produces a model whose accuracy is nearly flat in target-context similarity: it gains little from a highly related example placed in its context, and in the high-similarity region falls below standard zero-shot fine-tuning. 
    \item Similar-Context is competitive when close neighbors are available, but relatively weak when they are not, in most of our settings. Figure 1 summarizes these findings.
\end{enumerate}
To understand why, we design probes that track over the course of training, how much a model relies on in-weights learning (IWL) versus genuine input-conditioned use of context, together with a third diagnostic for the propensity to copy context labels irrespective of their relevance to the target. The probes reveal that: under Random-Context, ICL capability decays steadily with training, so the model comes to rely almost entirely on its weights; under Similar-Context, IWL develops more weakly and the propensity to blind copying mildly rises.
Table~\ref{tab:blind-copy-prompt} (in Appendix~\ref{sec:appendix:blind-copying}) presents an example of such blind copy.

Both diagnoses point to the importance of contrast in target-context similarity. A model never shown a related example in context has no reason to learn to use one. A model always trained with related examples learns to copy from context instead of acquiring the task in its weights, and when every context example is related, it has no reason to discriminate among them by relevance. We show that introducing contrast both across training batches and within a single context 
yields a model that is robust across the full similarity spectrum, retaining both IWL and ICL without resorting to copying. We refer to this recipe as Contrastive-Context. Our claim is not that any single ingredient is novel (mixtures of retrieval strategies are not new), but show that the similarity structure of IC-Train data determines how robust the resulting model is to varying degrees of target-context relatedness.

\xhdr{Contributions}
\begin{enumerate*}
    \item We highlight the importance of a fine-grained evaluation of models trained under \fsft\ to diverse degree of relatedness of contexts to target examples.
    \item We show that combining across-batch and within-context contrast in similarity yields robustness across the entire spectrum.
    \item We evaluate IC-Train across grades of target-context relatedness on 32 MT configurations plus Text-to-SQL and multilingual semantic parsing, and show that the two prevailing strategies are robust over complementary parts of the spectrum.
    \item We introduce probes for in-weights learning, in-context learning, and blind copying that trace these profiles to distinct training dynamics, and that can be applied to characterize any IC-trained model 
\end{enumerate*}

\newcommand{\ctx}{C}

\section{Problem Setup and Context Selection Strategies}
Let $\cM$ denote a pre-trained LLM to be finetuned on a task (e.g. low-resource translation). We are given a labeled dataset $D$ of $N$ pairs of inputs-outputs $\{(\vx_1,\vy_1),\ldots,(\vx_N,\vy_N)\}$ drawn from the   
underlying joint distribution $P(X,Y)$ of this task.  
Our goal is to train $\cM$ using $D$ so that its performance remains robust under these deployment settings: \begin{enumerate*}[(1)]
    \item As more labeled pairs are appended to $D$  (e.g., via user feedback), the model's accuracy should improve on test inputs with highly similar cases in $D$, \emph{without requiring further parameter updates.} 
    \item For test inputs lacking similar examples in $D$, accuracy should be no worse than that of a model trained in the standard zero-shot setting.
\end{enumerate*}  
A natural candidate to meet these goals is \fsft.
In standard fine-tuning, we maximize likelihood on each example independently as
 $   \E_{(\vx,\vy)\sim D} \log P_\theta(\vy|\vx)$.
\fsft\ samples $k+1$ labeled examples from $D$, place the first $k$ pairs as in-context examples while maximizing likelihood of target as:
\begin{equation}
\label{eq:fsft:generic}
\begin{split}
    \E_{(\vxt,\vyt)\sim D} \E_{\ctx = \{(\vx_i,\vy_i):1:k\}\sim D}\log P_\theta(\vyt | \ctx,\vxt)
\end{split}
\end{equation}
The above training explicitly trains the model to leverage in-context examples, thereby better preparing it to absorb any labeled data  during deployment.   
This paper shows that a crucial but often overlooked factor in \fsft\ is the inter-example similarity among the $k+1$ samples.  To make this explicit, we rewrite the \fsft\ objective in terms of the strategy $\choose(D,\vx)$, for choosing the $k$ examples from $D$ to accompany a target input 
$\vxt$:
\begin{equation}
\label{eq:fsft}
\begin{split}
       \E_{(\vxt,\vyt)} \E_{\ctx \sim \choose(D,\vxt)} \log P_\theta(\vyt | \ctx,\vxt)
\end{split}
\end{equation}
As we will show, $\choose(D,\vxt)$ critically influences whether \fsft\ strengthens or erodes our desired robustness properties.
%
 Most prior work studied \fsft\ with $k$ context examples chosen at random independent of $\vxt$.  We refer this method as \textbf{\fsrand}. A second strategy is to select as $C$ the  top-$k$ examples from $D$ most similar to $\vxt$, which  we call \textbf{\fssim}.  
 Section~\ref{sec:expt:accuracy} shows that 
 each of these is accurate over only part of the spectrum of target-context relatedness, and they cover complementary parts of it. 
%

\paragraph{\fsour:}  
Both regimes are missing contrast in target-context similarity, and we therefore construct training contexts that create contrast both among the examples within a context and across contexts. Contrast across batches comes from drawing a fraction $1-p$ of instances with random context, and fraction $\frac{p}{2}$ with similar context.  The remaining  $\frac{p}{2}$ creates contrast within a context: we sample one example weighed by similarity to the target, and the rest uniformly randomly. When the labeled pool is small, no  example may be close enough to the target during fine-tuning, so the model could lose the capability of harnessing highly related context examples.  We therefore augment $\epsilon$ fraction of the training instances (\fsrand\ and \fssim) with synthetic highly similar example by small perturbation of $\vxt$.  For NLP tasks, the perturbation is obtained by getting a paraphrase of $\vxt$.  A pseudocode of the overall method appears in Algorithm~\ref{alg:cc}.
\begin{algorithm}[h]
\small
\caption{\textsc{Contrastive-Context}: sampling one \fsft\ training instance}
\label{alg:cc}
\begin{algorithmic}[1]
\Require Labeled pool $D$; target $(\vxt,\vyt)\!\sim\!D$; context size $k$;
  contrast fraction $p$; augmentation fraction $\epsilon$; similarity $K(\cdot,\cdot)$
\Ensure Context $C$ of $k$ labeled pairs
\State $u \sim \mathrm{Unif}[0,1]$
\If{$u > p$} \Comment{prob.\ $1-p$}
  \State $m \gets \textsc{Random-Context}$
  \State $C \gets k$ pairs drawn uniformly from $D$
\ElsIf{$u > p/2$} \Comment{prob.\ $p/2$}
  \State $m \gets \textsc{Similar-Context}$
  \State $C \gets \textsc{TopK}(D, \vxt, k)$ under $K$
\Else \Comment{prob.\ $p/2$}
  \State $m \gets \textsc{Intra-Context Contrast}$
  \State $(\vx_s,\vy_s) \sim D$ with prob.\ $\propto K(\vxt,\vx_s)$
  \State $C \gets \{(\vx_s,\vy_s)\} \cup \{k\!-\!1$ pairs drawn uniformly$\}$
\EndIf
\If{$m \in \{\textsc{Random-Context}, \textsc{Similar-Context}\}$ \textbf{and} $\mathrm{Unif}[0,1] < \epsilon$}
  \Comment{synthetic highly-similar example}
  \State $\tilde{\vx} \gets \textsc{Paraphrase}(\vxt)$
  \State replace a random element of $C$ with $(\tilde{\vx},\vyt)$
\EndIf
\State \textbf{shuffle} $C$ \textbf{and return} $C$
\end{algorithmic}
\end{algorithm}

\newcommand{\metric}{\text{metric}}
\newcommand{\cK}{\mathcal{K}}
\newcommand{\vyhr}{\vyh_R}


\section{Evaluating \fsft\ Across Target-Context Relatedness}
\begin{figure*}[!t]
\centering
\begin{small}
\begin{tabular}{ccc}
    \includegraphics[width=0.32\textwidth]{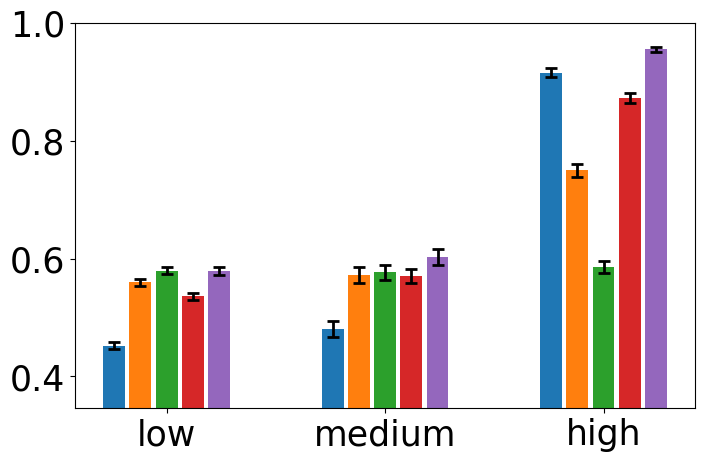} & \includegraphics[width=0.32\textwidth]{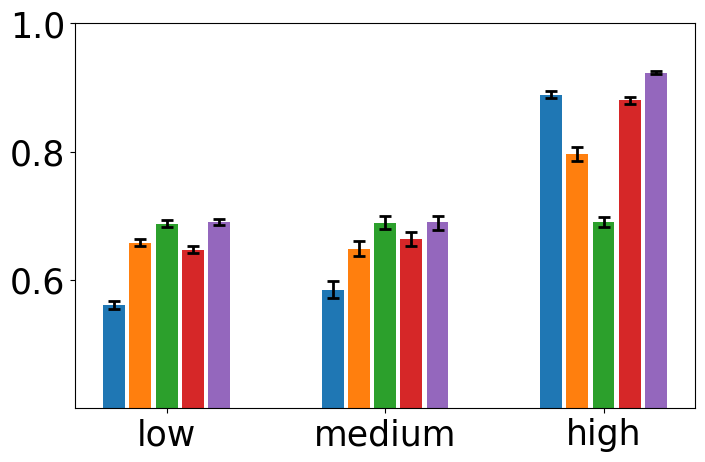} & \includegraphics[width=0.32\textwidth]{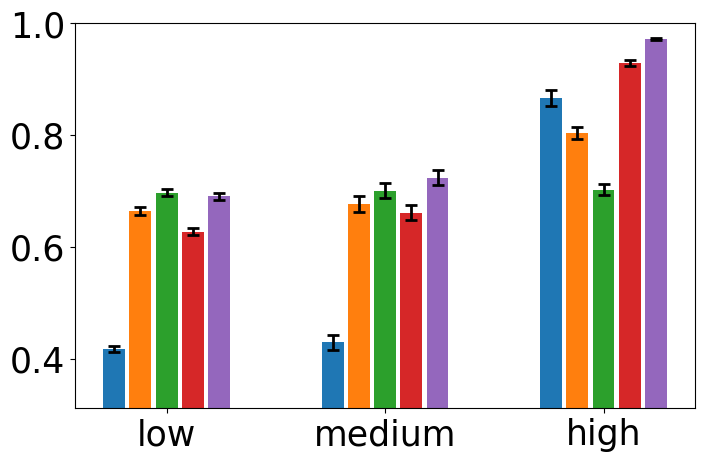} \\
    Llama1B-En-Lt-ID & Llama1B-En-Hi-ID & Mistral-En-Lt-ID \\
    \includegraphics[width=0.32\textwidth]{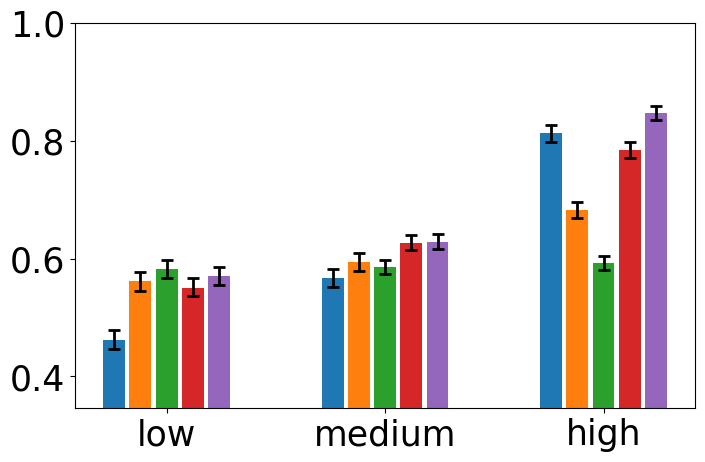} & \includegraphics[width=0.32\textwidth]{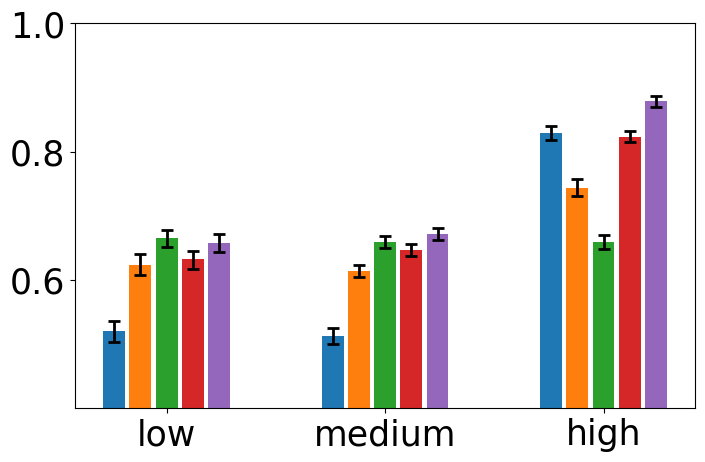} & \includegraphics[width=0.32\textwidth]{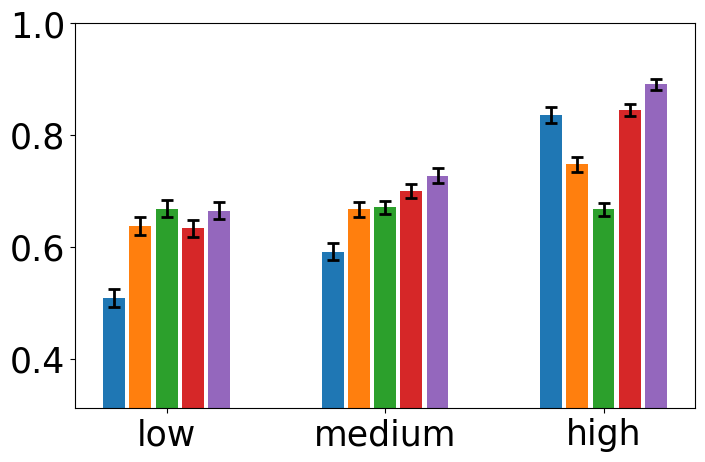} \\
    Llama1B-En-Lt-OOD & Llama1B-En-Hi-OOD & Mistral-En-Lt-OOD \\
    \includegraphics[width=0.32\textwidth]{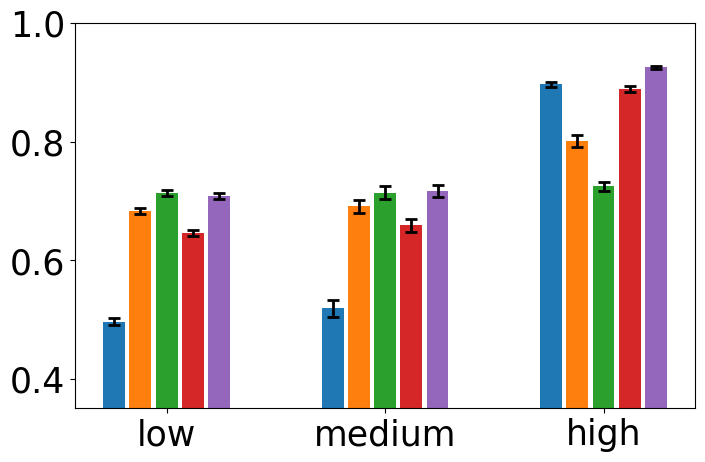} & \includegraphics[width=0.32\textwidth]{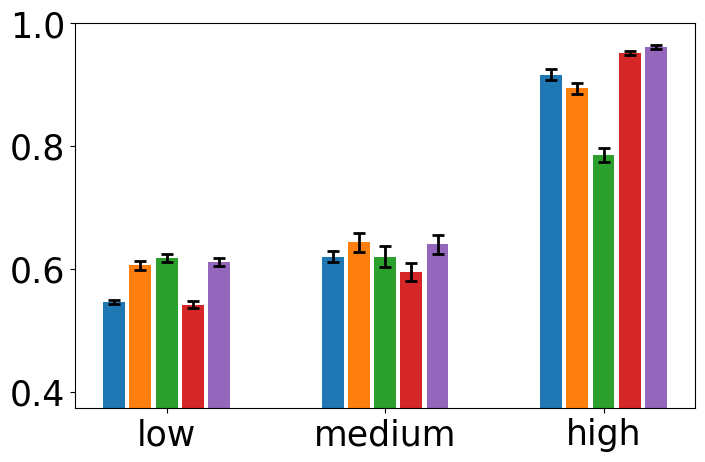} & \includegraphics[width=0.32\textwidth]{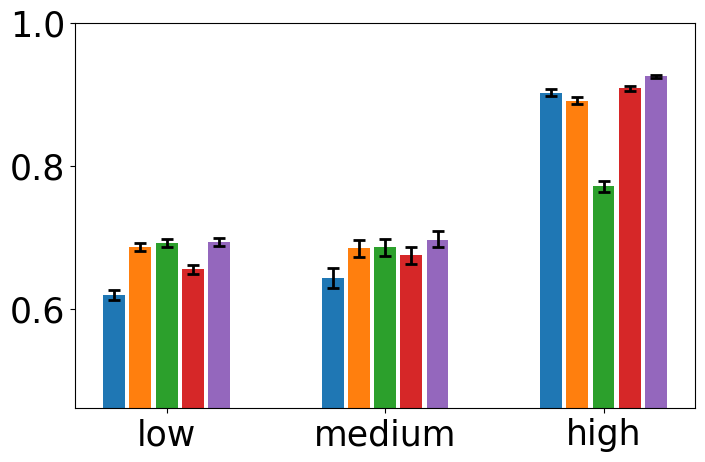} \\
    Mistral-En-Hi-ID & Qwen-En-Ta-ID & Qwen-En-Hi-ID \\
   \multicolumn{3}{c}{\includegraphics[width=1\textwidth]{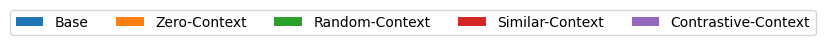}} \\
\end{tabular}
    \caption{\label{fig:accuracy}\small Effect of fine-tuning a {\color{RoyalBlue}base model} with different strategies ({\color{orange}\zs}, and \fsft\ under {\color{ForestGreen}\fsrand}, {\color{Red}\fsse}, and {\color{violet}\ours}) on accuracy over varying grades of similarity to in-context examples for nine of 32 different models, language-pairs, and test-sets (remaining plots in Appendix~\ref{sec:appendix:fig1}). \new{X-axis: Level of maximum similarity of target with in-context examples. The similarity ranges here are - Low: $0-0.33$, Medium: $0.33-0.67$, High: $0.67-1$.} Y-axis: Accuracy (COMET score). 
    }
    \end{small}
\end{figure*}

\vspace{-0.1in}
We fine-tune four open source models on four machine translation tasks, eleven Text-to-SQL task, three multilingual semantic parsing tasks, and in Appendix~\ref{sec:app:align} also include two synthetic alignment reasoning tasks.

\label{sec:setup}


\xhdr{Evaluation Setup}
To study the effect of different grades of relatedness of the input to the in-context examples, we evaluate under four different contexts:
\begin{enumerate*}[(1)]
\item Randomly selected $k$ examples from $D$,
\item Select $k$ examples most similar to $\vxt$ using BM25 
\item $k-1$ random examples and $1$ example most similar to $\vxt$ using BM25,
\item $k-1$ random examples and one closely related example obtained to be a paraphrase of $\vxt$ with $\vy_i=\vyt$. This mode checks the scenario of whether a model can harness closely related feedback from users after training.
\end{enumerate*}
Over the union of all (context,test) instances, we plot accuracy as a function of the maximum similarity of the input $\vxt$ to an example in its context. 

\xhdr{Models}
We use four open-source LLMs:
\begin{enumerate*}
    \item Llama 3.2 1B base, 
    \item Llama 3.1 8B base~\cite{llama3}
    \item Qwen 2.5 7B~\cite{qwen2.5}. 
    \item Mistral 7B v0.3~\cite{jiang2023mistral7b}
\end{enumerate*}

\subsection{Machine Translation (MT) Tasks}
We translate from English to four different languages: Lithuanian, Tamil, Hindi, and German.  We train on standard benchmarks, and evaluate under two test-sets, for a total of 32 (model,task,test-set) configurations.  For testing, we consider two types: a generic dataset from Flores (ID), and to evaluate domain adaptation, an OOD dataset from a specialized domain such as judiciary and religion.   
Appendix~\ref{sec:appendix:datasets} provides more information.  The prompt is given in Appendix~\ref{sec:appendix:prompt_icl}. 

\xhdr{Methods Compared}
We compare \fsour\ against the following baseline methods of fine-tuning LLMs for MT.
\begin{enumerate*}[(1)]
\item \zs~\cite{vieira-etal-2024-much}: Standard zero-shot fine-tuning frequently used for domain-adapting MT models.
    \item \fsrand: ~\citet{alves-etal-2023-steering} is one of the earliest works on harnessing \fsft\ for MT, and proposes to select examples randomly.
    \item \fssim: Recently, \citet{zebaze-etal-2025-context} evaluated \fsft\ with similarity based example selection where BM25 emerged as a competitive option, so we use this as a baseline.
\end{enumerate*}
As reference we also show accuracy for the base untuned models.
 For all training setups (\fsrand, \fssim, \fsour) we choose $k$ randomly from $[0\ldots 5]$ and for testing $k$ is set to 5.  We use 
COMET-22 to measure accuracy.

\begin{figure*}[!htbp]
\centering
\begin{tabular}{ccc}
    \includegraphics[width=0.3\textwidth]{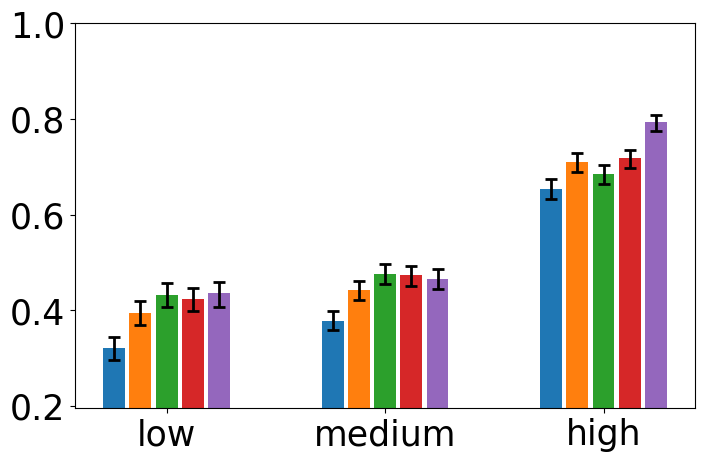} & \includegraphics[width=0.3\textwidth]{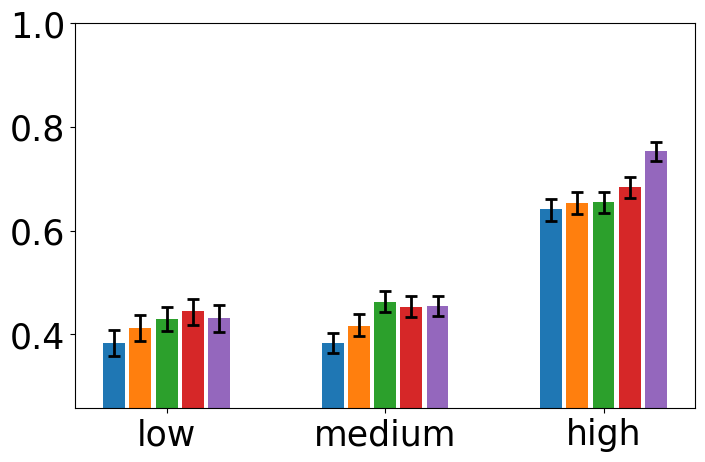} & \includegraphics[width=0.3\textwidth]{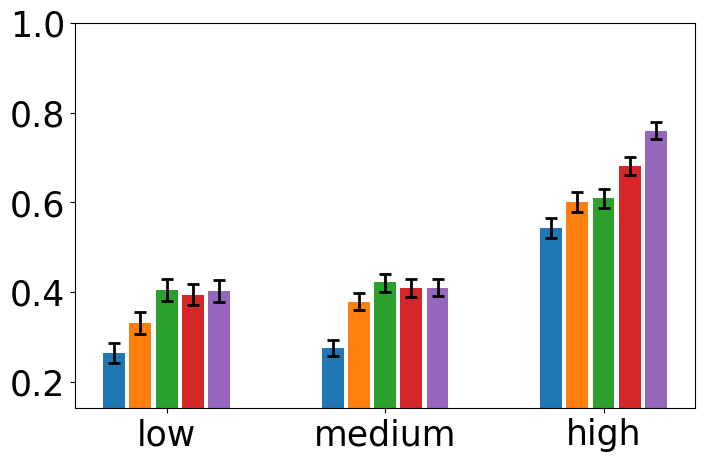}\\
    \multicolumn{1}{c}{Llama8B Instruct} & \multicolumn{1}{c}{Qwen Instruct} & \multicolumn{1}{c}{Mistral Instruct} \\[0.5em]
     \multicolumn{3}{c}{(a) Execution accuracy for the Text-to-SQL task. } 
\end{tabular}
\includegraphics[width=\textwidth]{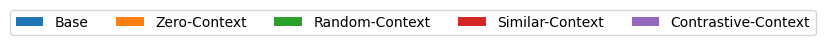}
\caption{\small \label{fig:text-to-sql} Accuracy (Y-axis) against three levels of similarity (X-axis). The similarity ranges here are - (a) Low: $0-33$, Medium: $33-67$, High: $67-100$, (b) Low: $0-0.33$, Medium: $0.33-0.67$, High: $0.67-1$. Error bars show 95\% confidence intervals.}
\end{figure*}
\todo{remove (a) and combine captions. change in main text (3(a) -> (3).}
\label{sec:expt:accuracy}
In Figure~\ref{fig:accuracy} we plot accuracy against the maximum target-context similarity
on combinations of four models, four tasks, two test settings, \todo{rewrite as 9 of 32
different models, language-pairs, and test-sets} and five training methods:
\zs, \fsft\ with \fsrand, \fssim, \fsour, and the original untuned base model.  \new{For clarity, we group similarity values into three bins of Low, Medium, and High similarity. }
\new{The error bars show the 95\% confidence interval with 1000 bootstrap resampling of the test data.}
\new{Raw plots and over all 32 task-model combinations appear in Appendix~\ref{sec:appendix:fig1}.}
Based on these plots, we make the following observations:

\xhdr{\fsrand\ worse than \zs\ in harnessing related in-context examples}
\fsft\ with \fsrand\ (\textbf{{\color{ForestGreen}green}} bar) provides almost a flat accuracy \todo{not flat for Qwen. As mentioned in rebuttal, see qwen2.5 technical report} even as we  increase the similarity of context to examples  in almost all 32 model-task-test settings.
    Its performance is worse than even \zs\ when presented with similar examples in context!  
    Already \zs\ (\textbf{{\color{Orange}orange}} bar) fine-tuned is worse than the base untuned model (\textbf{{\color{RoyalBlue}blue}} bar) for high target similarity, and \fsft\ was meant to preserve ICL capability, but we observe that with random context, context usage actually deteriorates.

    
\xhdr{\fssim\ weak in the low similarity region}
\fsft\ with \fssim\ (\textbf{{\color{Red}red}} bar)  suffers in the low similarity region, and is worse than all other forms of fine-tuning in that region.   However, it responds to increasing target-context similarity with increasing accuracy.  

\xhdr{\fsour\ robust on the full target-context similarity spectrum}
\fsour (\textbf{{\color{violet}violet}} bar), is among the highest performing methods.  Compared to \fssim, the second best performer, \fsour\ scores when highly related examples are present in context,  and is competitive in low similarity ranges. 


\subsection{Text-to-SQL and Semantic Parsing}
\textbf{Text to SQL Task}
\label{sec:sql_expts}
We experiment on Text-to-SQL as an instance of a Text-to-code generation task where the need for online adaptation to private databases is compelling.  We use the BIRD dataset~\cite{li2024can}, with its official train split for fine-tuning and dev split spanning eleven distinct databases. More details in Appendix~\ref{sec:appendix:sql_expts}.  The results shown in Figure~\ref{fig:text-to-sql}(a) further add evidence to the robustness of \fsour\ to handling contexts at varying similarity levels.  Observe how \fsft\ with \fsrand\ is worse \todo{worse or comparable} than even \zs\ in the high similarity range, and with \fsour\ we obtain the best \todo{best or comparable} accuracy across all levels.  In this task \fssim\ does not suffer in the low-similarity range because the labeled pool is small, and \fssim\ almost reduces to random for many instances.

\new{
\xhdr{Multilingual Semantic Parsing task}
\label{sec:semantic_parsing_expts}
We experiment with the cross-lingual semantic parsing task MTOP~\cite{zhang2023xsemplr} on three languages --- Spanish, German, and French. 
The results in Appendix Figure~\ref{fig:text-to-sql_semantic_parsing}(b) show 
that \fssim\ suffers in the low and medium similarity regions compared to \fsrand\ and \fsour. In the high similarity region, \fsrand\ suffers compared to \fssim\ and \fsour. In this task too \fsour\ performs competitively or better across all similarity levels.}

\begin{figure*}[!h]
\centering
\begin{tabular}{ccc}
    \includegraphics[width=0.32\textwidth]{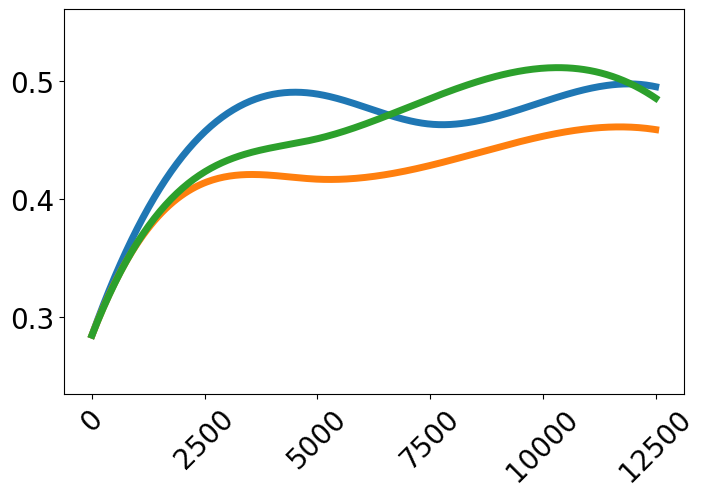} & \includegraphics[width=0.32\textwidth]{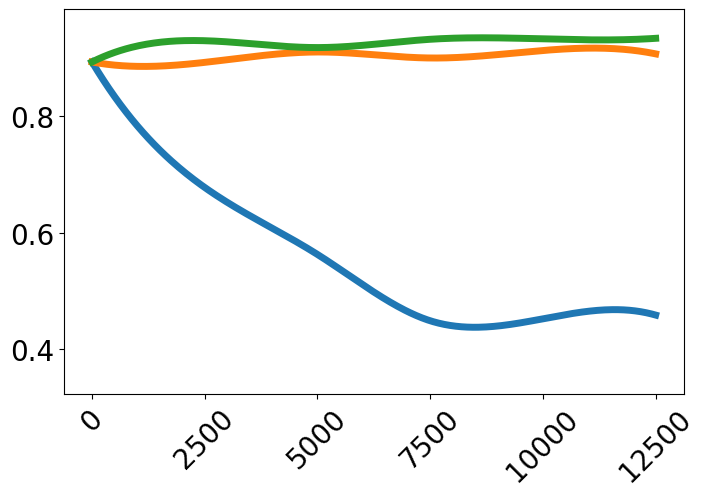} & \includegraphics[width=0.32\textwidth]{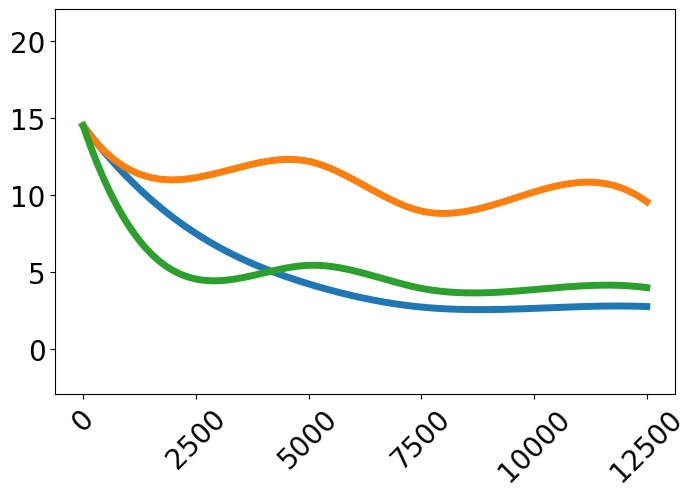} \\
    \iwlscore\ Llama1B-En-Ta-ID & \iclscore\ Llama1B-En-Ta-ID & \copyscore\ Llama1B-En-Ta-ID \\
    \includegraphics[width=0.32\textwidth]{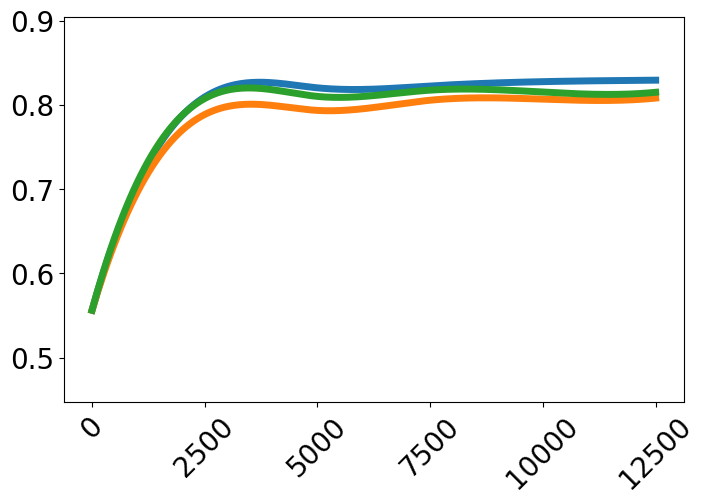} & \includegraphics[width=0.32\textwidth]{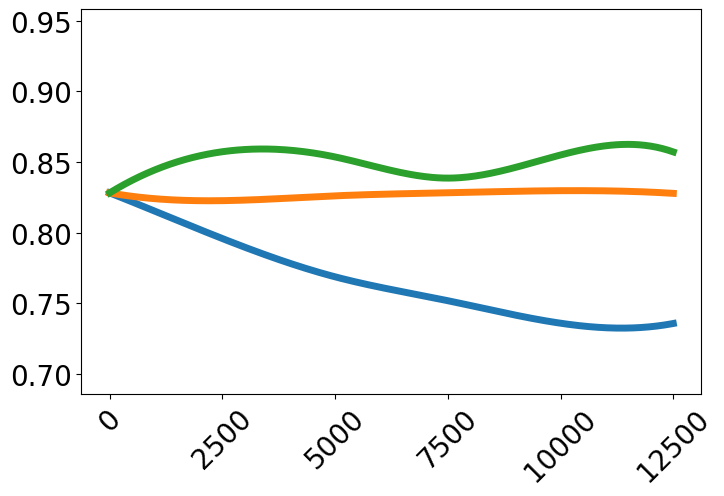} & \includegraphics[width=0.32\textwidth]{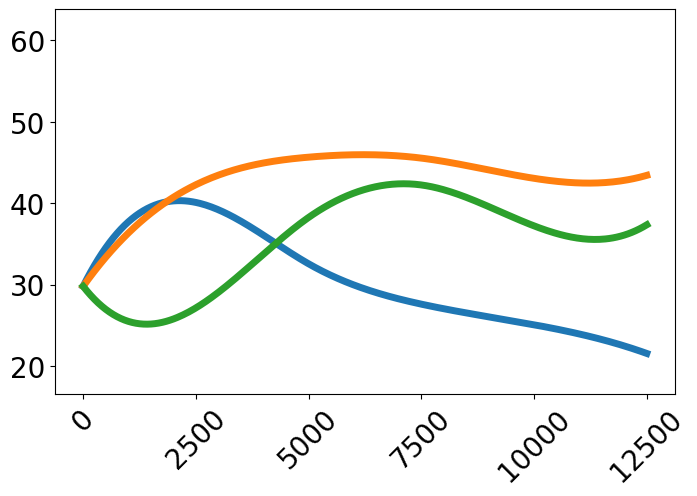} \\
    \iwlscore\ Llama1B-En-De-OOD & \iclscore\ Llama1B-En-De-OOD & \copyscore\ Llama1B-En-De-OOD \\
    \includegraphics[width=0.32\textwidth]{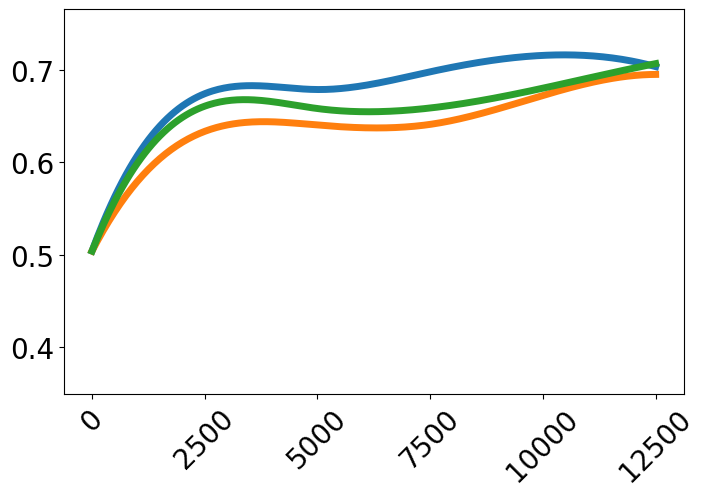} & \includegraphics[width=0.32\textwidth]{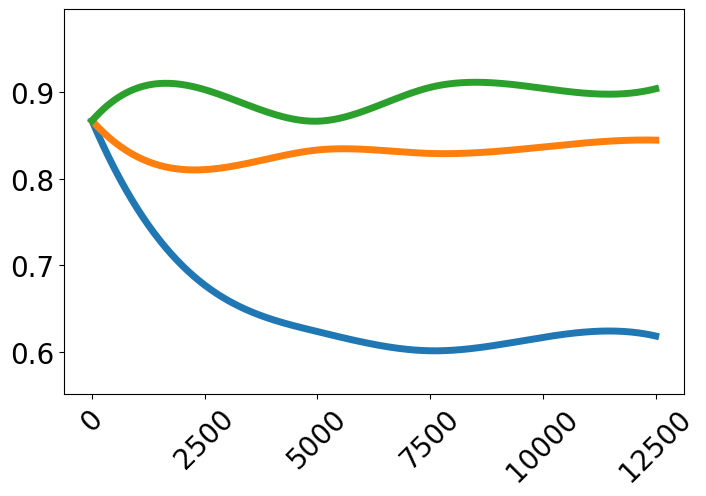} & \includegraphics[width=0.32\textwidth]{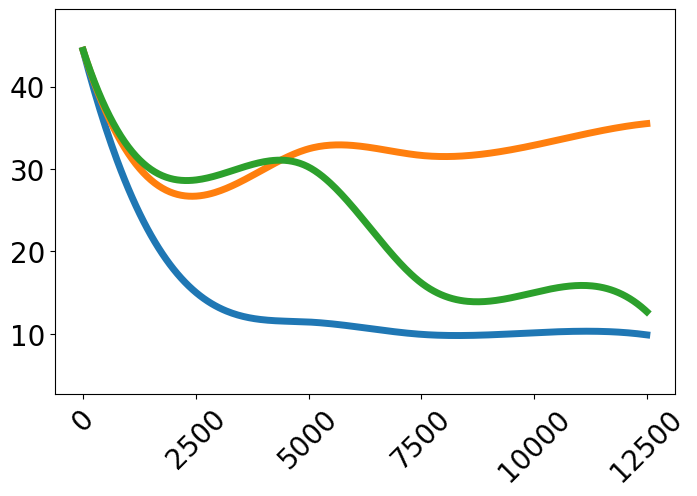} \\
    \iwlscore\ Mistral-En-Lt-OOD & \iclscore\ Mistral-En-Lt-OOD & \copyscore\ Mistral-En-Lt-OOD \\
   \multicolumn{3}{c}{\includegraphics[width=0.6\textwidth]{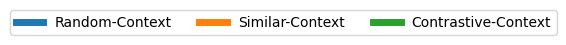}} \\
\end{tabular}
    \vspace*{-1em}
    \caption{\small \label{fig:emergence}Emergence of different forms of learning in three training methods: \fsrand, \fsse, and \ours. X-axis is training steps and Y-axis denotes scores of one of the three probes. Results of other model-task and datasets in Appendix Figure~\ref{sec:appendix:fig3}. IWL-score of \fssim\ is lowest, ICL-score of \fsrand\ diminishes fast with training, Copy-score (a stress test under shuffled contexts, see text) of \fssim\ is higher. \fsour\ provides best retention of ICL and IWL capabilities without resorting to copying. Curves are cubic-spline interpolations through the evaluated checkpoints, shown for readability. 
 %
 }
\end{figure*}

\subsection{Probing In-Weights, In-Context, and Copying Behaviour}
\label{sec:emergeReal}
To understand the effect of target-context similarity on the emergence of different forms of learning, we design two probes for in-weights and in-context learning, together with a third diagnostic for the propensity to copy context labels, and analyze these as training progresses.
\subsubsection*{Probes to detect different forms of learning}
Let $\vyh_C=\argmax_\vy P_\theta(\vy | C, \vxt)$ denote the predicted output under set $C=[\vx_1,\vy_1,\ldots \vx_{k},\vy_{k}]$ and  $\vyh_\phi$ denote zero-shot prediction.  Let $\cK(\vx,\vx')$ denote similarity between two inputs and $\simS(\vy,\vy')$ between outputs. 

\xhdr{In-Weights Learning Probe}
We quantify the in-weights learning as 
the similarity between predictions under random and empty context. 
\label{eq:iwlscore}
$
\iwlscore(\cM)\! =\! 
\E_{\vxt}\E_{C \sim \text{Random}(D,k)}
\simS(\vyh_C, \vyh_\phi)
$

\xhdr{In-Context Learning Probe} 
We quantify ICL capability by comparing prediction $\vyh_C$ under a one-similar context $C$ to the labels in context weighted by their similarity to $\vxt$. One-similar context is obtained by injecting a paraphrase for $\vxt$ with gold $\vyt$ among remaining $k-1$ random examples, so $C=[\vx_1,\vy_1,\ldots,\vx_{i^*}=\text{paraphrase}(\vxt), \vy_{i^*}=\vyt,\ldots \vx_{k},\vy_{k}]$. $\iclscore(\cM) = $
\label{eq:iclscore}
\[
\begin{aligned}
\E_{\vxt}\E_{C \sim \text{One-similar}(D,\vxt,k)} 
\sum_{j=1}^k  \frac{\cK(\vxt,\vx_j)\simS(\vyh_C, \vy_j)}{\sum_i \cK(\vxt,\vx_i)}
\end{aligned}
\]

\xhdr{Blind Copy Probe}
A model with propensity to copy blindly from the context would output a $\vyh$  similar to a label $\vy_i$ in-context irrespective of $\vxt$'s similarity to $\vx_i$. 
We quantify this by shuffling the one-similar context (above) so
that no $\vx_i$ and $\vy_i$ are correctly matched. An example for $k=3$ is: 
$C=[\vx_1,\vyt,\vx_{2}=\text{paraphrase}(\vxt), \vy_{3}, \vx_{3},\vy_{1}]$.
 Pure ICL would output $\vy_3$ based on similarity of $\vx_2$ with $\vxt$ whereas a model that copies independent of $\vx$ could output one of the other labels.  Thus, we define copy score as maximum similarity to a label other than the one attached to $\vx_{i^*}$ where $i^*$ is the position of $\vxt$'s paraphrase. 
\[
\begin{aligned}
&\copyscore(\cM) = \\
&\E_{\vxt}\E_{C \sim \text{Shuffle}(\text{One-similar}(D,\vxt,k))} \max_{i\in C, i \ne i^*}\simS(\vyh_C, \vy_i)
\end{aligned}
\]
We used cosine similarity of sentence embeddings from all-MiniLM-L6-v2 \cite{wang2020minilmdeepselfattentiondistillation} for $\cK$ and COMET for $\simS$ for the first two probes but BLEU for the copy score to measure the model's propensity to blindly copy the lexical tokens.
The shuffled context is deliberately off-distribution: no model encounters it during training or deployment. We therefore treat this probe as a stress test that isolates whether a model's use of context is conditioned on input similarity, rather than as a measurement of deployment behaviour. 

\xhdr{Main observations} 
In Figure~\ref{fig:emergence} we show the emergence of different forms of learning as measured by these three probes for various model, language pair, and dataset combinations. More combinations appear in the Appendix~\ref{sec:appendix:fig3}. We can make the following observations from these graphs:
\begin{enumerate*}[(1)]
    \item The IWL-score (first column) of \fsrand, and \fsour\ steadily increases with training steps with \fsour\ lagging only slightly behind.  However, for \fssim\ the IWL-score is distinctly lower, sometimes by a significant margin.  This, coupled with the observation that \fssim\ provides lower accuracy in Figure~\ref{fig:accuracy} when target-context similarity is low, shows that \fssim\ is less effective to learn the task in-weights, and/or is distracted by irrelevant examples.
    \item The ICL-score (second column) shows that for \fsrand\ ICL capability steadily drops with training, explaining why it provides almost the same accuracy across different grades of target-context similarity. For \fsour\ and \fssim, the ICL capability stays almost the same or increases. 
    \item The copy score (last column) decreases with training using \fsrand, and \fsour\ and gets substituted by either in-weights or in-context learning.  In contrast, for \fssim\ the copy score is higher than for the other models.  On real datasets, \fssim\ training may not consistently show increased copy tendencies because not every target will find all similar top-k examples, and the training data may naturally contain a mix of random and similar contexts.  But in spite of the natural mixing, we observe \fssim\ to result in reduced IWL and more blind copying compared to \fsour.  
    In Table~\ref{tab:blind-copy-prompt} we present an example of blind copying even under normal deployment context.
    
\end{enumerate*}

\begin{figure*}
\centering
\begin{tabular}{ccc}
    \includegraphics[width=0.32\textwidth]{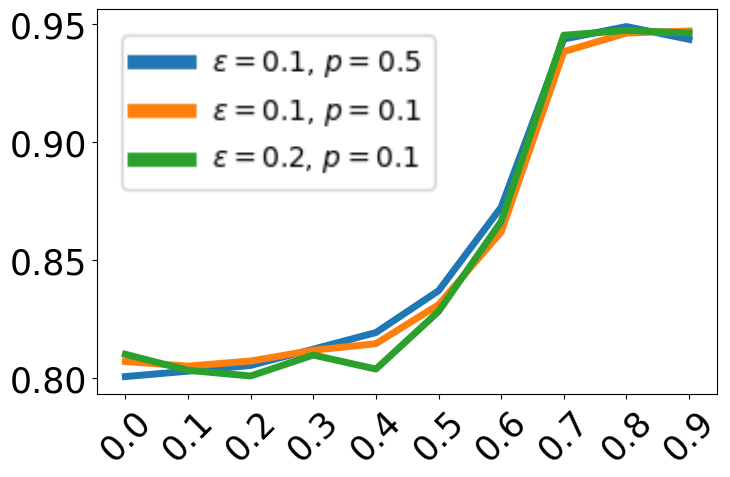} & \includegraphics[width=0.32\textwidth]{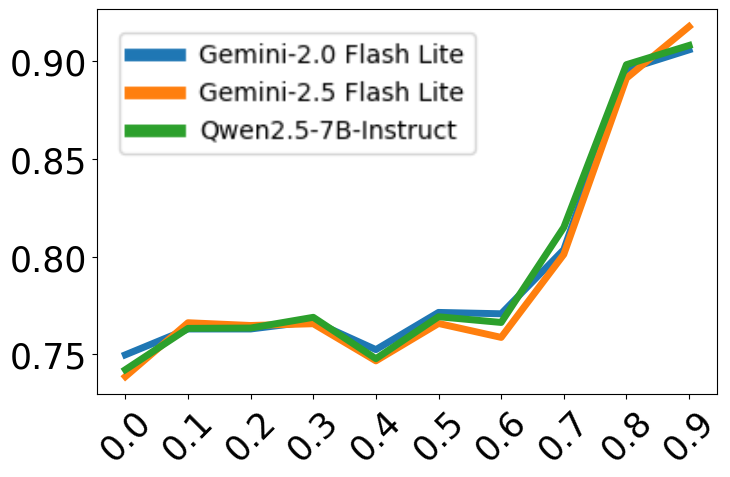} &  \includegraphics[width=0.32\textwidth]{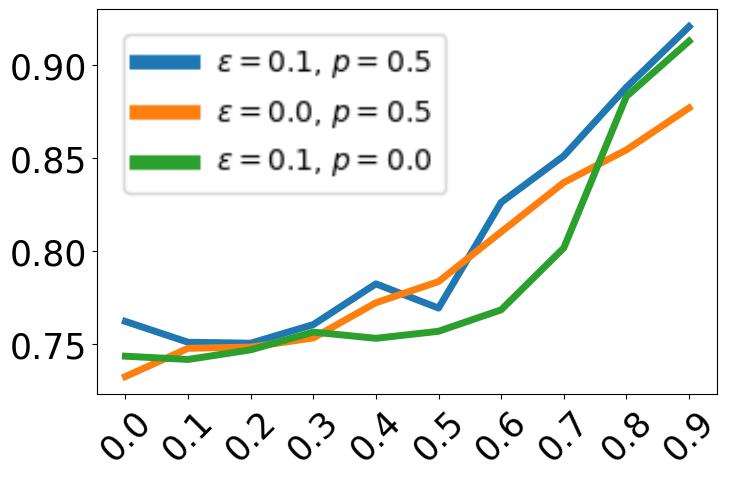} \\
   Llama1B-En-De-ID & Llama1B-En-De-OOD & Llama8B-En-Lt-OOD \\
    \includegraphics[width=0.32\textwidth]{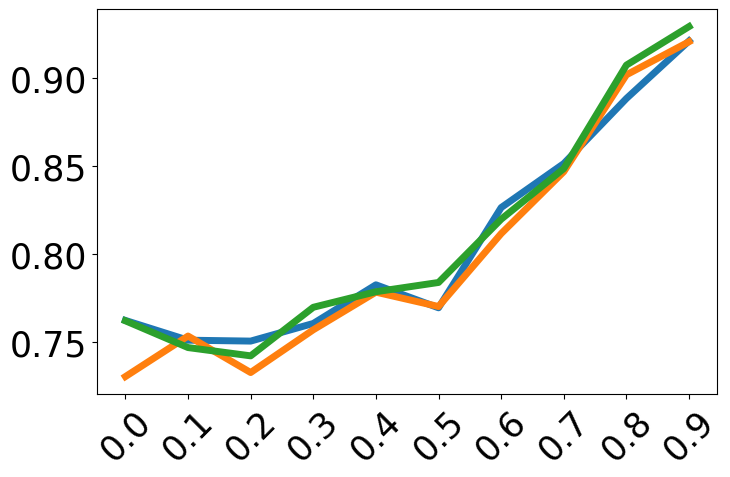} & \includegraphics[width=0.32\textwidth]{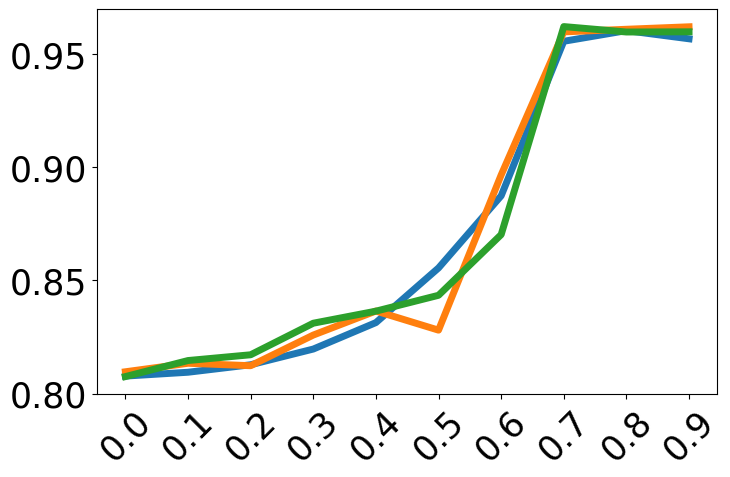} &  \includegraphics[width=0.32\textwidth]{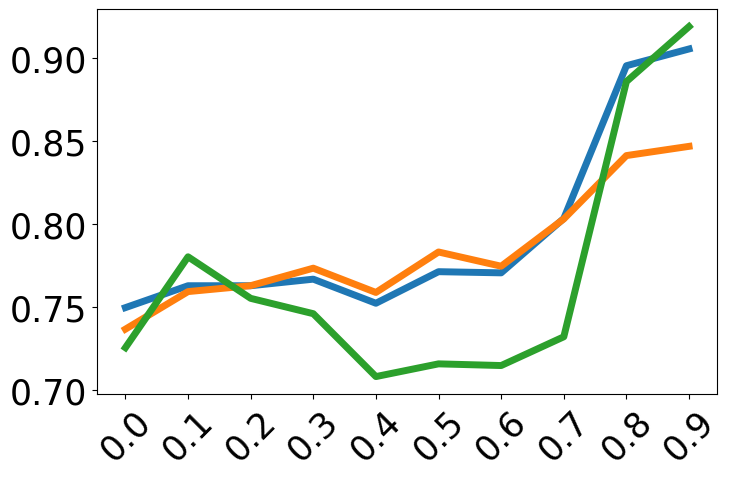} \\
   Llama8B-En-Lt-OOD & Llama8B-En-Lt-ID & Llama1B-En-De-OOD \\
    (a) Sensitivity to $\epsilon,p$ values & (b) Quality of paraphrasing & (c) Role of different types of contrasts \\
\end{tabular}
    \vspace*{-1em}
    \caption{\label{fig:ablation} Ablations to show (a) the robustness of \ours\ to variant non-zero $\epsilon$, $p$ values, (b) various paraphrasing models, (c) importance of various levels of similarity in the training data. X-axis is target-context similarity and Y-axis is accuracy.}
\end{figure*}
\todo{show $\epsilon$ and p used in the main paper}

\subsection{Ablations}
We present brief ablations on our method here. Appendix~\ref{sec:appendix:ablations} has more details.
\xhdr{Sensitivity to $\epsilon,p$ values} \fsour\ creates contrast using parameter $p$ to control the split of random Vs similar contexts, and control the fraction of highly similar synthetic examples ($\epsilon$).  In Figure~\ref{fig:ablation}(a) we show the method is robust to alternative values $\epsilon,p$ between 0.1 and 0.5.  
\xhdr{Quality of paraphrasing}
To study impact paraphrase quality, we evaluated with two alternative models \texttt{gemini-2.5-flash-lite} a strong model, and \texttt{Qwen2.5-7B-Instruct} as an open source, possibly weaker model compared to \texttt{gemini-2.0-flash-lite}. Figure~\ref{fig:ablation}(b) shows that \fsour\ remains stable across these three paraphrasing models. 
%
\xhdr{Role of different types of contrasts}
We evaluate the role of different types of contrasts that \fsour\ creates in Figure~\ref{fig:ablation}(c). We set $\epsilon=0$ to suppress paraphrase-augmentation and observe that test instances with high target-
context similarity suffers. We set $p=0$ to suppress the generation of similar real examples, which causes medium-scale accuracy to drop. 

\section{Related Work}
\xhdr{Fine-tuning with in-context examples}
Zero-shot fine-tuning is known to erode the ICL ability of pre-trained
LLMs~\cite{brownicl}, an effect documented across instruction
tuning~\cite{duan2024exploringICLandIT,goyal2025contextparametric,wang2024losscontext}
and domain-adapted machine translation~\cite{alves-etal-2023-steering}.
Fine-tuning in ICL mode (\fsft) is the standard remedy:
\citet{alves-etal-2023-steering} show for MT that it preserves ICL ability that
zero-shot fine-tuning loses, \citet{Zheng2024FinetuningLL} report similar gains
on English--German and English--Chinese, \citet{bornschein2025finetuned} study
\fsft\ as an efficient adaptation mechanism, and
\citet{min-etal-2022-metaicl} apply it in multi-task
meta-training. All of this work selects in-context examples
\emph{at random}, and we showed that this preserves ICL far less reliably, and in the high-similarity regime can be \emph{worse} than zero-shot fine-tuning.

\xhdr{Similarity-based example selection}
More recent methods select in-context examples by similarity to the input.
At inference time this is well established to improve
ICL~\cite{LevyBB23,kothyari-etal-2025-diverse,zebaze-etal-2025-context};
\citet{Zheng2024FinetuningLL} carry such selection into training.
\citet{gao2024customizinglanguagemodelresponses} also use contrastive examples,
but define contrast by the \emph{correctness} of the context rather than
relatedness to the target, and at prompting rather than fine-tuning time.

\xhdr{Effect of context structure on model learning}
Closest in spirit, \citet{Chan2022} show that training-data distribution
governs whether ICL or in-weights learning (IWL) emerges: bursty distributions,
where related examples recur in context, elicit ICL, non-bursty ones elicit IWL,
and Zipfian data yields both. \citet{singh2023transient} observe that ICL can be
transient, decaying into IWL with continued training, and in-context repetition
has been shown to accelerate ICL~\cite{zucchet2025the,bratulić2025unlockingincontextlearningnatural}.
We recast this in terms of graded target--context \emph{similarity}, which
generalizes burstiness and repetition to real NLP tasks where relatedness is
continuous rather than binary. Our distinctive finding that contrast \emph{within} a context
matters does not follow from burstiness alone, and is what separates
in-context learning from indiscriminate copying of context labels.

\xhdr{ICL--IWL tradeoffs under task mixtures}
A substantial literature studies ICL emergence when training over a mixture of
$T$ tasks from a family, where the $k$ in-context examples share the target's
task $f_\tau$
~\cite{reddy2024the,singh2024what,akyurek2024context,edelman2024evolution,park2025competition,nguyen2025differential,wurgaft2025,ku2025predictability,kim2025taskdiversityshortensicl,singh2025,fu2024breaking}.
These studies use synthetic regression, classification, or sequence-completion
tasks~\cite{what_icl_garg,Zhang2023TrainedTL,Olsson2022IncontextLA,rajaraman2024transformers,shi2024why,agarwal2025}
with randomly drawn contexts. We instead fine-tune on a \emph{single}, real
sequence-to-sequence task, where the axis of variation is not which task an
example belongs to but how related it is to the target.

\section{Conclusion}
We studied how the similarity between a target and its in-context examples during IC-Train shapes the model that results. Evaluated across the full spectrum of target-context relatedness, the two prevailing strategies prove accurate over complementary parts of that spectrum: random contexts yield a model that gains little from a related example even when one is placed in its context, while similarity-retrieved contexts weaken accuracy on targets without close neighbors. Probes for in-weights learning, in-context learning, and copying trace these profiles to distinct training dynamics. Both point to the same missing ingredient, and we showed that introducing contrast in similarity both across training batches and within a single context, yields models that remain accurate across the whole spectrum on machine translation, Text-to-SQL, and semantic parsing. For practitioners deploying a model alongside a growing labeled pool, the similarity structure of IC-Train data is a design choice worth making deliberately.

\section*{Limitations}

\paragraph{Dependence on an external paraphraser.}
\ours\ assumes access to a model that can produce meaning-preserving
near-duplicates of a target input. For MT we paraphrase the English source, i.e.\ the
high-resource side of the pair; whether comparable paraphrase quality is attainable
when the input side is itself low-resource is untested. 
We ablate the choice of
paraphraser (Figure~\ref{fig:ablation}), but not the regime in which
paraphrases are semantically drifted or degenerate.

\paragraph{Task and language coverage.}
All our tasks are fixed, single-task sequence-to-sequence problems with relatively short
inputs: translation into four languages (En$\rightarrow$X only), Text-to-SQL, semantic
parsing, and synthetic alignment. We do not evaluate classification, long-form
generation, multi-turn dialogue, or reasoning-heavy tasks, where the notion of a
``similar'' example---and hence of a useful context---is less well defined. We also
restrict contexts to $k \in [0,5]$; the behaviour of the three context regimes under
many-shot contexts, where retrieved examples naturally span a wider similarity spectrum,
remains open.

\paragraph{Hyperparameters and compute overhead.}
We do not tune $p$ and $\epsilon$ per task; we only show robustness over the range
$0.1$--$0.5$ on a subset of settings, and we give no principled procedure for choosing
them as a function of pool size or task.
We report wall-clock figures for these steps but do not
present a compute-matched comparison against the baselines.


\bibliography{custom,ICL,regression_ICL,pubs}

\appendix
\onecolumn
\let\addcontentsline\SAVEDaddcontentsline
\etocsettocstyle{\section*{Appendix Table of Contents}}{}
\etocsetnexttocdepth{subsection}
\localtableofcontents

\section{Prompts for various tasks}
\subsection{Prompts for obtaining translations with in-context examples}
\label{sec:appendix:prompt_icl}
\begin{itemize}
    \item 
\begin{wrapverb}
Translate the source text from English to German.
Source: Redistribution is only possible if there is actually something produced to be redistributed.
Target:
\end{wrapverb}
    \item 
\begin{wrapverb}
Consider the following 4 translations from English to German.
Example 1:
Source: This is now to be given concrete shape in the proposal to give EU citizens the right of free movement and residence.
Target: Jetzt soll sie in dem Vorschlag über das Recht der Unionsbürger auf Freizügigkeit und Aufenthalt konkret ausgestaltet werden.

Example 2:
Source: These data can only be seen as striking, alerting us to the urgent need for humanitarian aid to these countries to be properly directed to the area of health and the provision of basic medical care to highly deprived communities.
Target: Diese Zahlen sind erschreckend und müssen für uns ein Alarmsignal sein, dass es dringend notwendig ist, die humanitäre Hilfe für diese Länder in angemessener Weise in den Bereich des Gesundheitswesens und der Bereitstellung von medizinischer Grundversorgung für die am meisten benachteiligten Gemeinschaften zu lenken.

Example 3:
Source: There are a number of actions we need to take in future in the area of budget control within the agencies.
Target: Es gibt eine Reihe von Maßnahmen, die wir in Zukunft im Bereich der Haushaltskontrolle in den Agenturen ergreifen müssen.

Example 4:
Source: It is precisely in the area of the environment that people's dissatisfaction has been most deeply felt.
Target: Gerade im Umweltbereich spürt man die Unzufriedenheit der Bevölkerung am deutlichsten.

Translate the source text from English to German.
Source: It is precisely in the area of budget policy that these three basic principles need to be given concrete expression.
Target:
\end{wrapverb}
\end{itemize}

\subsection{Prompt to generate paraphrases}
\label{sec:appendix:prompt_paraphrase}
\begin{wrapverb}
Generate 5 diverse paraphrases for the given input English sentence.
Before paraphrasing, carefully analyze the original sentences and keywords. 
Ensure significant variation in phrasing, structure, and wording while preserving meaning.
The paraphrases should be such that their translation into another language could still be the same.
Only modify the grammar and sentence structure such that all the paraphrases should be translatable to a common sentence in another language.

### Inputs:
Original Sentences:
{sentences}

### Example:
1. 
#### Input:
Original Sentences: 
[
    "Sentence 1": "In New York City, the streets were alive with the sound of honking taxis and chatter from the crowded sidewalks.",
    "Sentence 2": "\"I can't believe it's already time for the annual reunion,\" Emily said, her voice filled with excitement.",
    "Sentence 3": "The Amazon rainforest is home to countless species of plants and animals, many of which have yet to be discovered.",
]

#### Example Paraphrases:
{{
    "Sentence 1": [ 
        "The streets of New York City buzzed with the noise of honking taxis and voices from the bustling sidewalks.", 
        "In the heart of New York City, honking taxis and lively chatter filled the streets.",
        "New York City's streets were vibrant with the sounds of honking taxis and the chatter of crowded sidewalks.",
        "In the bustling streets of New York City, honking taxis and lively conversations created a lively atmosphere.",
        "The sound of honking taxis and the chatter from crowded sidewalks made the streets of New York City come alive."
    ],
    "Sentence 2": [ 
        "Emily exclaimed, her excitement evident, \"I can't believe it's time for the annual reunion already.\"", 
        "With excitement in her voice, Emily said, \"I can't believe the annual reunion is here already.\"",
        "Emily's voice was filled with excitement as she said, \"I can't believe it's already time for the annual reunion.\"",
        "With a voice full of excitement, Emily remarked, \"I can't believe the annual reunion is already upon us.\"",
        "Filled with excitement, Emily exclaimed, \"I can't believe it's already time for the annual reunion!\""
    ],
    "Sentence 3": [ 
        "The Amazon rainforest shelters an untold number of plant and animal species, many still waiting to be discovered.", 
        "Countless species of plants and animals call the Amazon rainforest home, many of which remain undiscovered.",
        "Home to countless species of plants and animals, the Amazon rainforest holds many that are yet to be discovered.",
        "The Amazon rainforest is a habitat for innumerable species of plants and animals, many of which are still unknown.",
        "Many species of plants and animals, yet to be discovered, thrive in the Amazon rainforest."
    ],
}}

### Output Instructions:
- Generate exactly 5 paraphrases.
- Ensure correct JSON output syntax like the example
- Always escape quotes within a paraphrase (like in example 2)
\end{wrapverb}

\section{Datasets and Additional Tasks}
\subsection{Additional Setup Details}
For training, we adopted LoRA with a rank of $16$, a scaling factor of $\alpha = 32$, and a dropout rate of $0.05$. $\epsilon$ is set to 0.1 and $p$ is set to 0.5. Only the attention projection matrices (except the output projection matrix) were trained. We also trained all the attention projection matrices along the MLP layer matrices as an ablation. The results for this ablation can be found in Appendix~\ref{sec:appendix:ablations:lora_ablation}. Training was performed with a batch size of 2 using Adam with a learning rate of $2 \times 10^{-4}$ and a linear decay learning rate scheduler.   To obtain paraphrases for \fsour, we employed the \texttt{gemini-2.0-flash-lite} model. 

For the experiments in the main text, the training data size is 50000, and for \fsour, $\epsilon$ is set to 0.1. So, only 5000 paraphrases have to be generated. These are prompted to the paraphrasing model in groups of 10; hence, 500 API calls are required. This takes around 1.2 hours. This can be sped up greatly with multithreading.

We use 50000 training instances and 5000 validation instances for all models. Generating these instances by retrieving similar examples from a BM25 index takes approximately 25 minutes using 32 parallel processes.

\subsection{Low Resource Translation}
\label{sec:appendix:datasets}

\begin{table}[H]
\centering
\begin{minipage}{\linewidth}
\begin{center}
\begin{small}
\begin{tabular}{|c|cc|cc|cc|}
\hline
 & \multicolumn{2}{c|}{Train} & \multicolumn{2}{c|}{ID Test} & \multicolumn{2}{c|}{OOD Test} \\
\hline
 LP & Source & Size & Source & Size & Source & Size \\
\hline
En-De & Europarl    & 50000 & Flores+ & 1012 & EMEA     & 500 \\
En-Hi & Samanantar  & 50000 & Flores+ & 1012 & Judicial & 500 \\
En-Ta & Samanantar  & 50000 & Flores+ & 1012 & Tanzil   & 500 \\
En-Lt & Europarl    & 50000 & Flores+ & 1012 & EMEA     & 500 \\
\hline
\end{tabular}
\end{small}
\end{center}
\end{minipage}
\caption{\label{tab:dataset-mt}Datasets Used for MT. Flores+ refers to Flores-200. \textbf{Sources:} Europarl, Tanzil, and EMEA from \citet{opus}; Samanantar from \citet{samanantar}; Flores+ from \citet{flores}; Judicial from \citet{kunchukuttan-etal-2018-iit}.}
\end{table}

\begin{table}[H]
\centering
\begin{tabular}{|c|c|}
\hline
\textbf{Dataset} & \textbf{Link} \\
\hline
Europarl & \href{https://huggingface.co/datasets/Helsinki-NLP/europarl}{\nolinkurl{Helsinki-NLP/europarl}} \\
Tanzil    & \href{https://huggingface.co/datasets/Helsinki-NLP/tanzil}{\nolinkurl{Helsinki-NLP/tanzil}} \\
EMEA   & \href{http://huggingface.co/datasets/Helsinki-NLP/emea}{\nolinkurl{Helsinki-NLP/emea}} \\
Samanantar   & \href{https://huggingface.co/datasets/ai4bharat/samanantar}{\nolinkurl{ai4bharat/samanantar}} \\
Flores+   & \href{https://huggingface.co/datasets/openlanguagedata/flores_plus}{\nolinkurl{openlanguagedata/flores_plus}} \\
Judicial   & \href{https://huggingface.co/datasets/cfilt/iitb-english-hindi}{\nolinkurl{cfilt/iitb-english-hindi}} \\
\hline
\end{tabular}
\caption{Links to datasets used}
\end{table}

\subsection{Text-to-SQL}
\label{sec:appendix:sql_expts}
The paraphrases for this task are obtained using OpenAI-O3-mini.  Given an NL query, SQL pair $(\vx,\vy)$, we prompted the LLM to generate first a paraphrased SQL $\Tilde{\vy}$ that differs from $\vy$ only in the mention of literal constants, and then generating the corresponding NL question $\Tilde{\vx}$ describing the SQL $\Tilde{\vy}$.  An example paraphrased pair appears below:
\begin{quote}
$\vx$: Is molecule TR183 known to be a carcinogen?\\
$\vy$: SELECT T.label FROM molecule AS T WHERE T.molecule\_id = 'TR183’\\
.\\
$\tilde{\vx}$: Molecule KAC16 is carcinogenic.  Yes or No? \\
$\tilde{\vy}$: SELECT T.label FROM molecule AS T WHERE T.molecule\_id = 'KAC16’
\end{quote}
The prompt to the LLM comprises of generic natural language instructions, followed by a description of
the schema and metadata of the database queried, followed by the in-context examples, and then
the current test question $\vxt$. Following standard practice, instead of providing the entire database metadata, we filter a subset of schema information (relevant table and column names) using an open source schema filtering tool \footnote{https://github.com/RUCKBReasoning/text2sql-schema-filter} \cite{li2022resdsql}, \cite{li2024codes}.

\begin{figure*}[!htbp]
\centering
\begin{tabular}{ccc}
\includegraphics[width=0.3\textwidth]{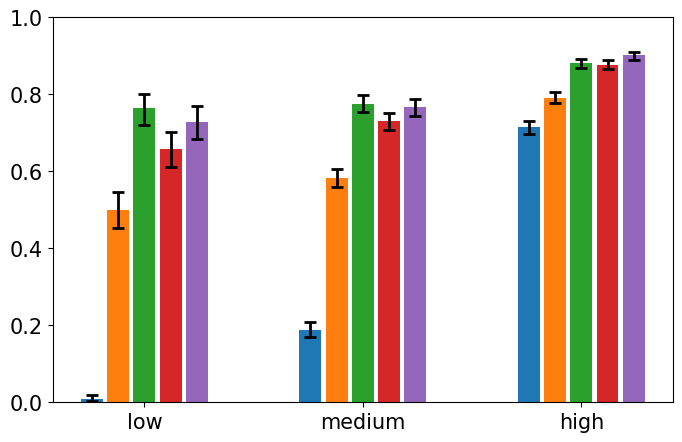} & \includegraphics[width=0.3\textwidth]{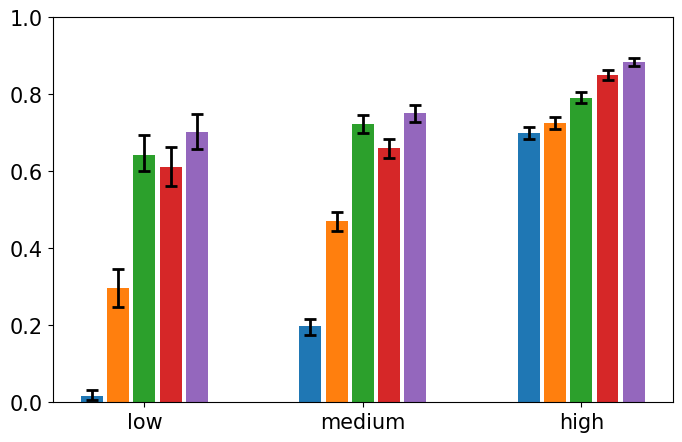} &
    \includegraphics[width=0.3\textwidth]{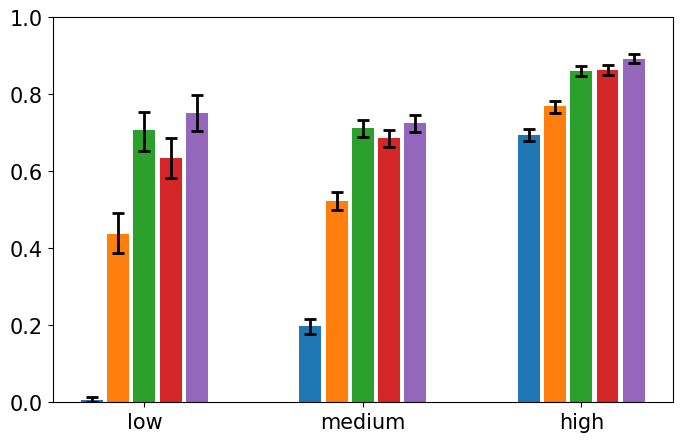}\\
     \multicolumn{1}{c}{Llama1B Spanish} & \multicolumn{1}{c}{Llama1B German} &
     \multicolumn{1}{c}{Llama1B French} \\[0.5em]
     \multicolumn{3}{c}{(b) Exact-Match accuracy for semantic parsing on MTOP}
\end{tabular}
\includegraphics[width=\textwidth]{figures_bar/exec_sim_legend.png}
\caption{\small \label{fig:text-to-sql_semantic_parsing} Accuracy (Y-axis) against three levels of similarity (X-axis). The similarity ranges here are - (a) Low: $0-33$, Medium: $33-67$, High: $67-100$, (b) Low: $0-0.33$, Medium: $0.33-0.67$, High: $0.67-1$. Error bars show 95\% confidence intervals.}
\end{figure*}
\todo{remove (b) and combine captions. change in main text (6(b) -> (6).}

\new{
\subsection{Synthetic Alignment Reasoning Task}
\label{sec:app:align}
We use the dataset as discussed by~\cite{agarwal2025} for reasoning about the alignment between two synthetic sequences. This task first defines a vocabulary $\cV$ of symbols, each symbol $\sigma \in \cV$ is associated with a probabilistic finite state automata (PFA) to generate sequences of length $c$ over elements over its own vocabulary. The input sequence $\vx$ contains $m$ symbols chosen from $\cV$, the output sequence $\vy$ is obtained by a fixed (unknown) permutation of $\vx$, and then sampling a length $c$ sequence from the FSA for each element in the permuted $\vx$.  More details can be found in \cite{agarwal2025}.  In order to correctly generate the output for a sequence, the model needs to reason about the input-output alignment and then the PFA of each token in $\cV$.  Examples appear below:}

\begin{table*}
\begin{tabular}{|l|l|} 
\hline
\textbf{Prompt} & Prompt string: $\vx^1: \vy^1 ~~\vx^2: \vy^2~~ \vx^3: \vy^3 ~~\vxt$ \\ \hline
Example $m=3,c=2$  & 
\texttt{ACB: rijjpr CAB: jjriwp ABC: rtprjh BCA:} \\ \hline
Example $m=2,c=4$ &
\texttt{AC: ririjhjh CA: jjhhriir AB: rttrprrp BA:} \\ \hline
\end{tabular}
\end{table*}

\begin{figure*}[!h]
\centering
\begin{tabular}{cc}
    \includegraphics[width=0.4\textwidth]{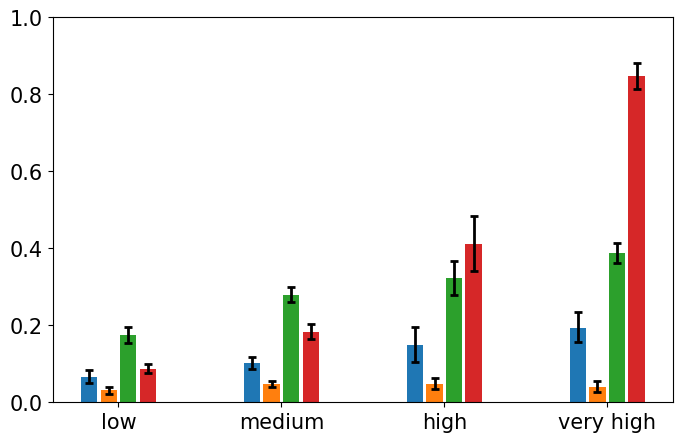} & \includegraphics[width=0.4\textwidth]{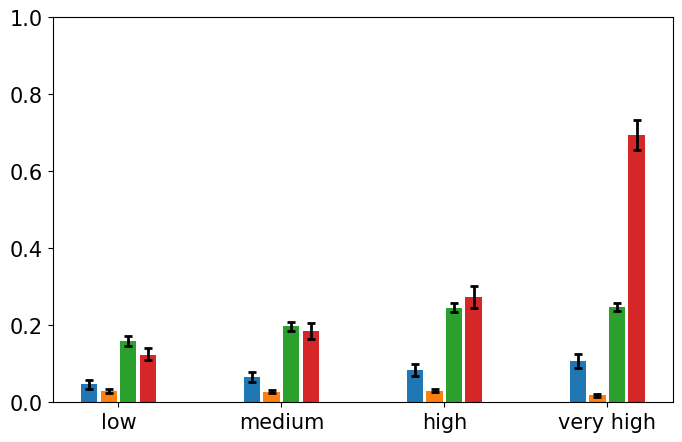}\\
    Llama1B ($m=4,\ c=4$) & Llama1B ($m=6,\ c=6$)\\
   \multicolumn{2}{c}{\includegraphics[width=0.8\textwidth]{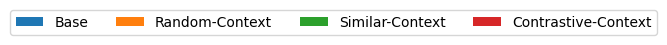}} \\
\end{tabular}
\caption{\label{fig:icatune_results}X-axis: Level of maximum Jaccard similarity between the sets of letters used in the target and the context examples. Y-axis: Average maximum accepted length of respective subsequence in y for the PFAs of each letter in $\vxt$. \fsrand\ performs worse than the baseline irrespective of the similarity level, probably because it has learnt the PFAs before learning the alignment, and thus it chooses a PFA at random for every letter in $\vxt$. On the other hand, the baseline model performs better because it blindly copies an ICL target example sequence corresponding to an ICL context example having some symbols in the same position as $\vxt$. Although \fsour\ performs second only to \fssim\ in the low and medium similarity regions, \fsour\ outperforms all the models in the high and very high similarity regions.}
\end{figure*}

\new{
For this task we choose $|\cV|=25$ and ran under two settings of sequence lengths ($m=4,\ c=4$) and ($m=6,\ c=6$) with the alignment as a permutation. The training is done on 200 instances for 5 epochs. In the original paper, each instance sampled its own PFA to simulate an infinite mixture.  In this paper, since our goal is to develop ICL-IWL mixture for a fixed task, the PFAs per symbol were fixed throughout the entire process. 
We sampled in-context examples at different similarity levels as follows: the~\fsrand\ data is sampled using a vocabulary of size 25, and the~\fssim\ data is created by using a vocabulary of size 12. For the~\ours\ data, the paraphrases are generated by taking the vocabulary as the set of letters used in the target example, $\vxt$. Hence, the paraphrase is simply a shuffling of the letters in $\vxt$. The results are presented in Figure~\ref{fig:icatune_results}.
}

\newpage

\begin{figure*}[!htbp]
\centering
\begin{tabular}{cccc}
    \includegraphics[width=0.40\textwidth]{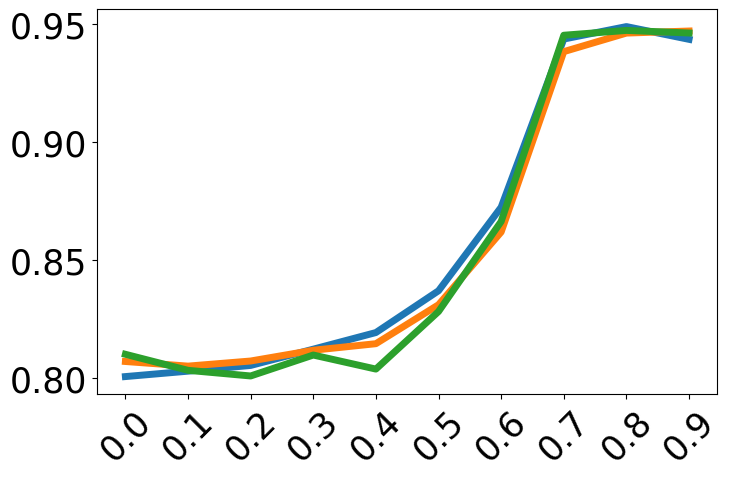} & \includegraphics[width=0.40\textwidth]{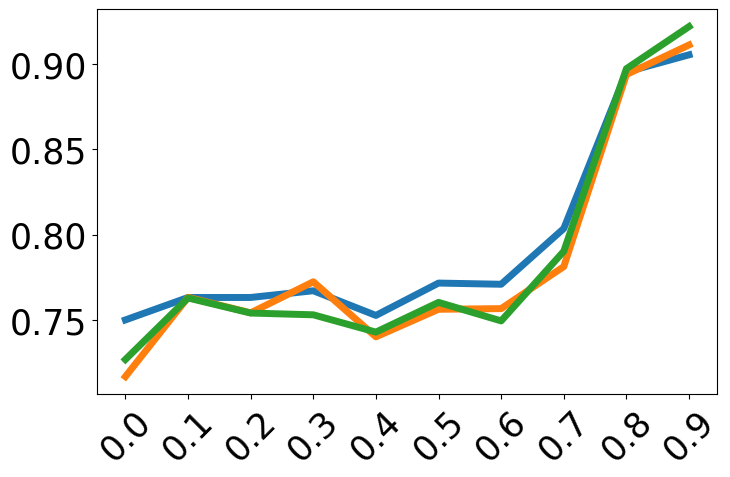} \\
    \multicolumn{2}{c}{Llama1B-En-De ID/OOD} \\
    \includegraphics[width=0.40\textwidth]{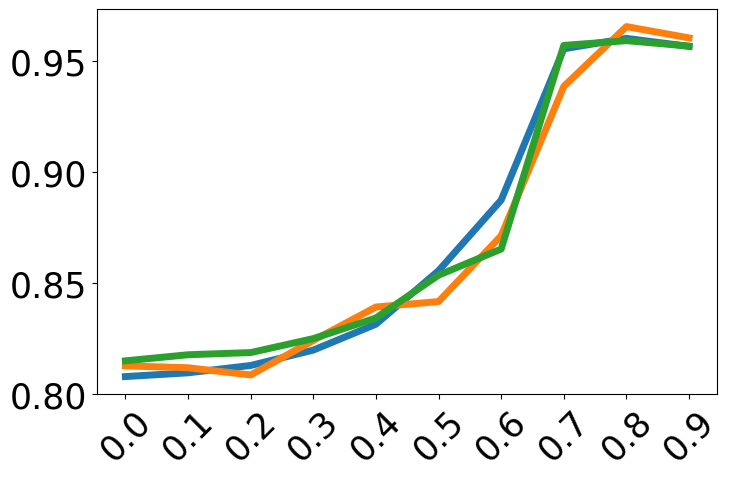} & \includegraphics[width=0.40\textwidth]{figures/contrast_comp_comet_sim_en-lt_8B_OOD.png} \\
    \multicolumn{2}{c}{Llama8B-En-Lt ID/OOD} \\[0.5em]
\end{tabular}
 \includegraphics[width=\textwidth]{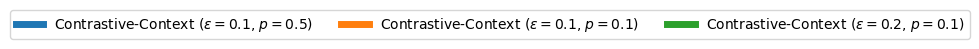}
 \caption{\label{fig:otherEpsP} \fsour\ is robust to specific choices of $\epsilon$ and $p$ as long as sufficient contrast is maintained. X-axis is target-example similarity and Y-axis is accuracy.}
\end{figure*}


\section{Anecdotes showing \ours\ outperform \fssim}
\label{sec:appendix:blind-copying}
Here we evaluate the \fssim{} and \ours{} models on \fsrand{} data, where no
demonstration is guaranteed to resemble the query. Table~\ref{tab:blind-copy-prompt}
shows two representative prompts. In both, \fssim{} reproduces one of the
in-context targets verbatim instead of translating the query, while \ours{}
produces a translation under the same prompt. In (b) the copied demonstration
shares only the token \emph{Iraq} with the query, suggesting that a shallow
surface cue is enough to trigger the copy.

\begin{table*}[t]
\centering
\footnotesize
\setlength{\extrarowheight}{1pt}
\begin{tabularx}{\textwidth}{@{}l@{\hspace{8pt}}X@{}}
\toprule
 
\multicolumn{2}{@{}l@{}}{\textbf{(a)}~Qwen~$\cdot$~OOD~$\cdot$~En--Lt} \\
\addlinespace[2pt]
$x_1$ & \copiedcolor{Each tablet contains 121.3 mg lactose.} \\
$y_1$ & \copiedcolor{Kiekvienoje tabletė yra 121, 3 mg laktozė} \\
\addlinespace[2pt]
$x_2$ & (hallucinations), mania (mental condition characterised by episodes of overactivity, elation or irritability), paranoia, suicidal thoughts \\
$y_2$ & Galite netekti riebalų Jūsų kojose, rankose ir veide, padaugėti riebalų pilve ir kituose vidaus \\
\addlinespace[2pt]
$x_3$ & Nitisinone is a competitive inhibitor of 4-hydroxyphenylpyruvate dioxygenase, an enzyme which precedes fumarylacetoacetate hydrolase in the tyrosine catabolic pathway. \\
$y_3$ & Nitizinonas yra konkuruojantis 4 - hidroksifenilpiruvato dioksigenazės (fermento, esančio prieš fumarilacetoacetato hidrolazę tirozino kataboliniame kelyje) inhibitorius. \\
\addlinespace[2pt]
$x_4$ & Humalog 100 U/ ml, solution for injection in cartridge \\
$y_4$ & Humalog 100 V/ ml, injekcinis tirpalas užtaise \\
\addlinespace[2pt]
$x_5$ & Nespo 10 micrograms solution for injection in a pre-filled pen (SureClick) Nespo 15 micrograms solution for injection in a pre-filled pen (SureClick) \emph{\ldots\ } \\
$y_5$ & Nespo 10 mikrogramų injekcinis tirpalas užpildytomis švirkštimo priemonėmis (SureClick) Nespo 15 mikrogramų injekcinis tirpalas užpildytomis švirkštimo priemonėmis (SureClick) \emph{\ldots\ } \\
\addlinespace[3pt]
Query & The list below contains adverse reactions seen in clinical trials. \\
\addlinespace[2pt]
\fssim & \copiedcolor{Kiekvienoje tabletė yra 121, 3 mg laktozė.} \\
\ours & \ourscolor{Žemiau pateikta neigiamų reakcijų, matomų klinikiniuose bandymuose, sąrašas.} \\
Reference & Toliau pateikiamas nepageidaujamų reakcijų, pastebėtų klinikinių tyrimų metu, sąrašas. \\
 
\midrule
 
\multicolumn{2}{@{}l@{}}{\textbf{(b)}~Llama-1B~$\cdot$~ID~$\cdot$~En--Ta} \\
\addlinespace[2pt]
$x_1$ & Tours are cheaper for larger groups, so if you're by yourself or with just one friend, try to meet other people and form a group of four to six for a better per-person rate. \\
$y_1$ & \ta{பெரிய குழுக்களுக்கு சுற்றுப்பயணங்கள் மலிவானதாகும், எனவே, உங்களுக்கோ அல்லது ஒரு நண்பருடனோ இருந்தால், மற்றவர்களைச் சந்திக்க முயற்சி செய்யுங்கள், மேலும் ஒரு நபருக்கு ஒரு சிறந்த விகிதத்திற்காக நான்கு முதல் ஆறு பேர் கொண்ட குழுவை உருவாக்கவும்.} \\
\addlinespace[2pt]
$x_2$ & The valves opening allowed a pressure release for the system and oil flowed on a pad to a tank that can hold 55,000 barrels (2.3 million gallons). \\
$y_2$ & \ta{வால்வுகளைத் திறப்பதென்பது, அமைப்பு அழுத்தத்தில் இருந்து விடுபடுவதை அனுமதித்தது மற்றும் 55,000 பேரல்களை (2.3 மில்லியன் கேலன்கள்) வைத்திருக்கக்கூடிய ஒரு டேங்குக்கான திண்டில் எண்ணெய் பாய்ந்தது.} \\
\addlinespace[2pt]
$x_3$ & \copiedcolor{Iraq's Abu Ghraib prison has been set alight during a riot.} \\
$y_3$ & \textcolor{copyred}{\ta{ஈராக்கின் அபு கிரைப் சிறை ஒரு கலவரத்தின் போது தீ வைக்கப்பட்டது.}} \\
\addlinespace[2pt]
$x_4$ & Fifteen of these rocks are attributed to the meteorite shower last July. \\
$y_4$ & \ta{கடந்த ஜூலை மாதம் விண்கற்கள் பொழிந்ததற்கு இவற்றின் பதினைந்து பாறைகள் காரணமாக இருந்தது.} \\
\addlinespace[2pt]
$x_5$ & The ACMA also found that despite the video being streamed on the Internet, Big Brother had not breached online content censorship laws as the media had not been stored on Big Brother's website. \\
$y_5$ & \ta{இணையத்தில் வீடியோ ஒளிபரப்பப்பட்டபோதிலும், பிக் பிரதர் இணையதளத்தில் மீடியா சேமிக்கப்படாததால், பிக் பிரதர் ஆன்லைன் உள்ளடக்க தணிக்கைச் சட்டங்களை மீறவில்லை என்பதையும்} ACMA \ta{கண்டறிந்தது.} \\
\addlinespace[3pt]
Query & The Iraq Study Group presented its report at 12.00 GMT today. \\
\addlinespace[2pt]
\fssim & \textcolor{copyred}{\ta{ஈராக்கின் அபு கிரைப் சிறை ஒரு கலவரத்தின் போது தீ வைக்கப்பட்டது.}} \\
\ours & \textcolor{oursblue}{\ta{ஈராக் செய்தியாளர்கள் இன்று காலை 12.00 மணிக்கு அறிக்கையை வெளியிட்டனர்.}} \\
Reference & \ta{ஈராக் ஆய்வுக் குழு, தனது அறிக்கையை இன்று 12.00} GMT\ta{-க்கு வழங்கியது.} \\
 
\bottomrule
\end{tabularx}
\caption{Two \fsrand\ prompts shown in full, with the outputs they
elicit. Each prompt contains five demonstrations $(x_i, y_i)$. \copiedcolor{Red italics} mark the demonstration whose target \fssim\ reproduces verbatim as its translation; \ourscolor{blue} marks the output of \ours, which translates the query in both cases. The very long fifth demonstration in (a) is truncated here for space; the model saw it in full.}
\label{tab:blind-copy-prompt}
\end{table*}

\section{Ablations}
\label{sec:appendix:ablations}
\subsection{Experiments with non-zero $\epsilon$ and $p$}
We present robustness of \fsour\ to alternative choice of $\epsilon$ and $p$ parameters in Figure~\ref{fig:otherEpsP}.

\subsection{Importance of training with different similarity levels}
\fsour\ includes in-context examples at three levels of similarity --- randomly selected examples which are of low similarity, naturally occurring top-k most similar examples from the training pool which are typically of medium similarity levels, and highly similar examples obtained by paraphrasing the target. 
We establish the necessity of employing these multiple similarity levels during training by creating two ablations: first,  we remove the highly similar paraphrased examples by setting $\epsilon=0$, and second, we remove the natural similar examples by setting  $p=0$.
\begin{figure*}[!htbp]
\centering
\begin{tabular}{cc}
    \includegraphics[width=0.40\textwidth]{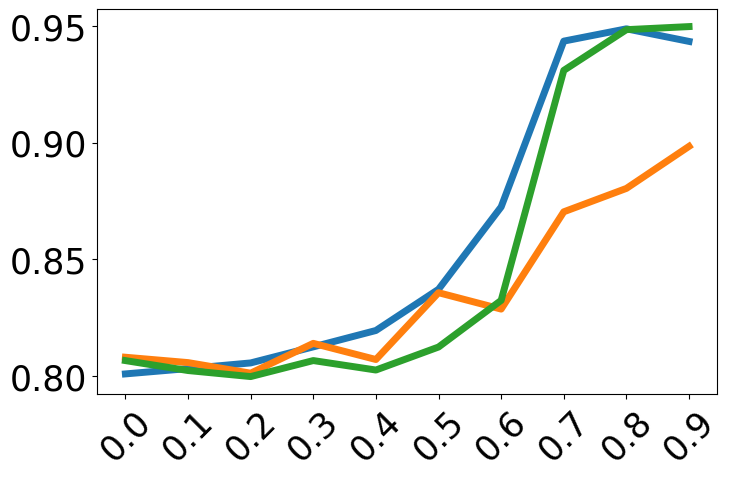} & \includegraphics[width=0.40\textwidth]{figures/contrast_comp_comet_sim_en-de_1B_OOD_p_0_eps_0.png} \\
     \multicolumn{2}{c}{Llama1B-En-De ID/OOD}\\
    \includegraphics[width=0.40\textwidth]{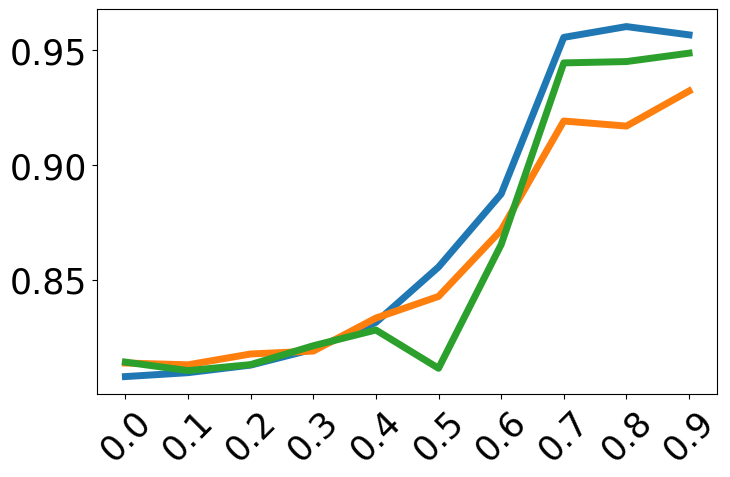} & \includegraphics[width=0.40\textwidth]{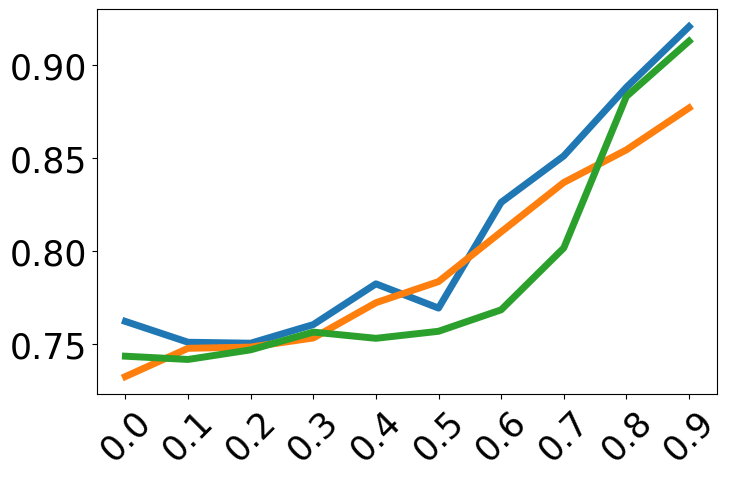} \\
    \multicolumn{2}{c}{Llama8B-En-Lt ID/OOD} \\[0.5em]
\end{tabular}
 \includegraphics[width=\textwidth]{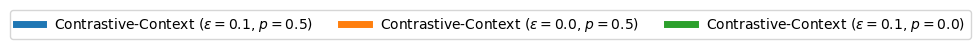}
 \caption{\label{fig:contrast_sim_levels} Ablations on \fsour\ to establish the importance of training with different similarity levels: X-axis is target-example similarity and Y-axis is accuracy.  Removing highly similar examples obtained via paraphrasing ($\epsilon=0$) causes test accuracy in the high similarity range to suffer.  Removing natural similar examples ($p=0$) causes accuracy in the medium range to suffer.}
\end{figure*}

Figure~\ref{fig:contrast_sim_levels} shows evaluation on test instances created with mixed similarity levels as described in Section~\ref{sec:expt:accuracy} (Evaluation Setup). We observe that test instances with high target-context similarity suffer on models trained with $\epsilon=0$.  This is evidenced by the dip in accuracy of the \textbf{{\color{orange}orange}} line compared to the original \fsour\ (blue line) in the right quarter of the X-axis).   Likewise, the model trained without naturally occurring Top-K pairs ($p=0$), performs poorly on test instances where IC examples are at medium levels of similarity with the target.  This is seen by the dip in accuracy of the  \textbf{{\color{ForestGreen}green}} line in the middle part of the $X$ axis. 

\new{
Overall, we observe the trend that it is important for \fsft\ to be trained at multiple similarity levels in order to provide the best accuracy under all levels of relatedness of the test example with the context. 
}
\label{sec:appendix:contrast_comp}
\new{

\subsection{Intra-Context Vs Inter-Context Contrast}
\fsour\ creates contrasts in similarity levels both amongst examples within a context, and across contexts.  Here we present ablations to understand the importance of contrasts within a context by comparing with a baseline that mixes Random-Context and Similar-Context to preserve only inter context contrast.  We present results in Figure~\ref{fig:contrast_intra_inter}.  On these tasks, the Random-Context and Similar-Context mixture (\textbf{{\color{Green}green}} line) are as good as \fsour\ (\textbf{{\color{RoyalBlue}blue}} line) in the low and medium similarity regime.  In the high similarity regime, because of the absence of paraphrases, this mixture loses out. When we remove the paraphrases from \fsour\ (\textbf{{\color{Orange}orange}} line) the two methods are equivalent on this task.
\begin{figure*}[!htbp]
\centering
\begin{tabular}{cc}
    \includegraphics[width=0.44\textwidth]{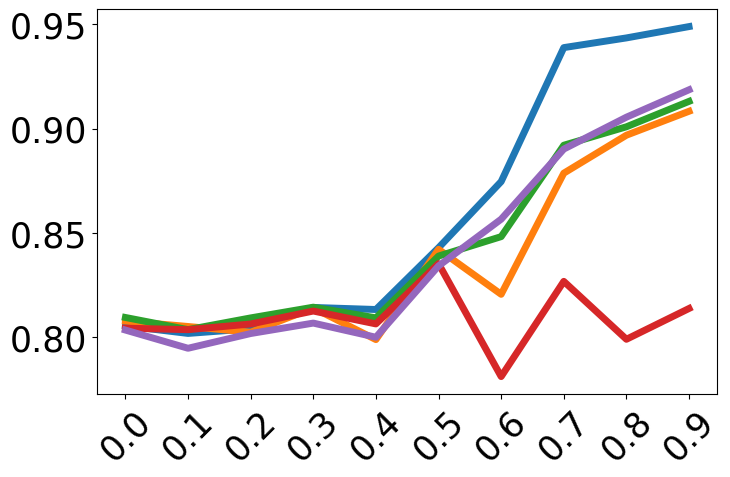} & \includegraphics[width=0.44\textwidth]{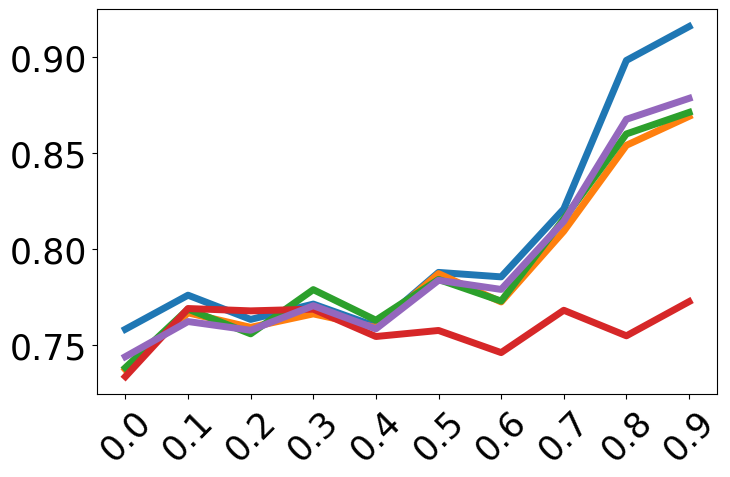} \\
      \multicolumn{2}{c}{Llama1B-En-De ID/OOD} \\
     \includegraphics[width=0.44\textwidth]{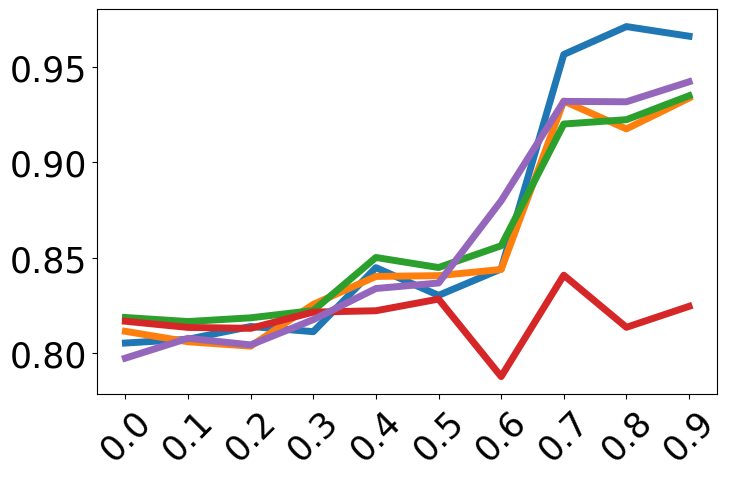} & \includegraphics[width=0.44\textwidth]{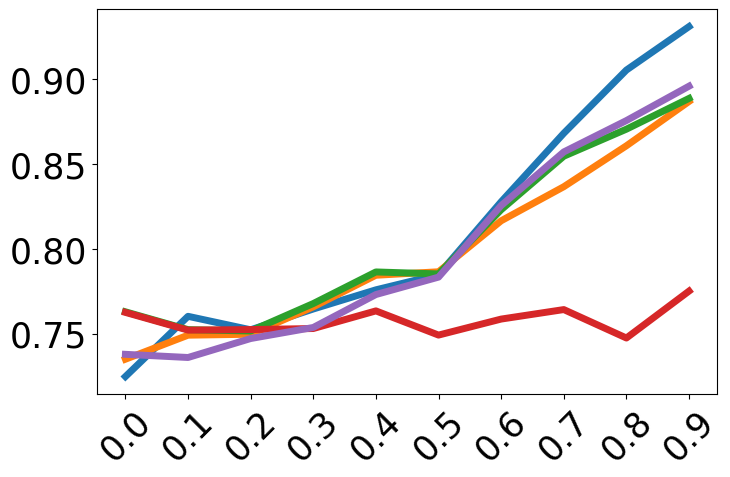} \\
   \multicolumn{2}{c}{Llama8B-En-Lt ID/OOD} \\[0.5em]
\end{tabular}
\includegraphics[width=\textwidth]{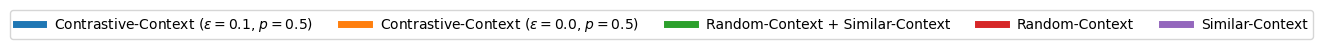}
\caption{\label{fig:contrast_intra_inter} Comparing a \fsrand+\fssim\ mixture with \fsour\ that additionally also creates contrasts within a context, and other baselines \fsrand\ and \fssim. X-axis is target-example similarity and Y-axis is accuracy.}
\end{figure*}
However, conceptually, a model that mixes Random-Context and Similar-Context could learn to swing between ignoring the context (IWL) and blindly copying from the context based on aggregated similarity with the context.  This forms a IWL+Copy mixture, which might not perform well when tested with only a subset of context examples as relevant.  Contrast within a context promotes true in-context learning where similarity between the x-s determines which y-s are copied.  In real-life limited data regimes, the top-K similar examples may differ in their similarity to the target, and thus Random-Context+Similar-context may behave like Contrastive-Context like we observed above. 

\subsection{Synthetic Paraphrases for Data Augmentation Vs Contrast}
An interesting question is whether the observed accuracy boost with synthetic paraphrased examples is due to increased contrast, or vanilla data augmentation.  To answer this question, we added the synthetic paraphrases for all target training examples to the training pool as augmented data.  We then invoked the \fsse\ method on this augmented dataset.  A comparison appears in Figure~\ref{fig:dataAugment}. We observe that with such data augmentation \fsse\ does perform well on test data in the high similarity range but shows a huge drop in accuracy in the low similarity range.  This is because \fsse\ goes for the Top-K most similar examples, and all training instances contain a paraphrase in the context causing no incentive for IWL to develop.  \fsour's use of paraphrases in a contrastive setting is important to develop IWL+ICL mixtures.

\begin{figure*}[!htbp]
\centering
\begin{tabular}{cc}
    \includegraphics[width=0.40\textwidth]{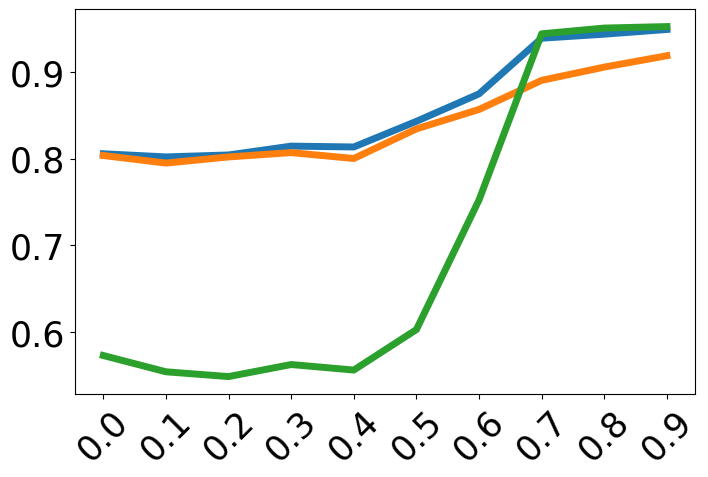} & \includegraphics[width=0.40\textwidth]{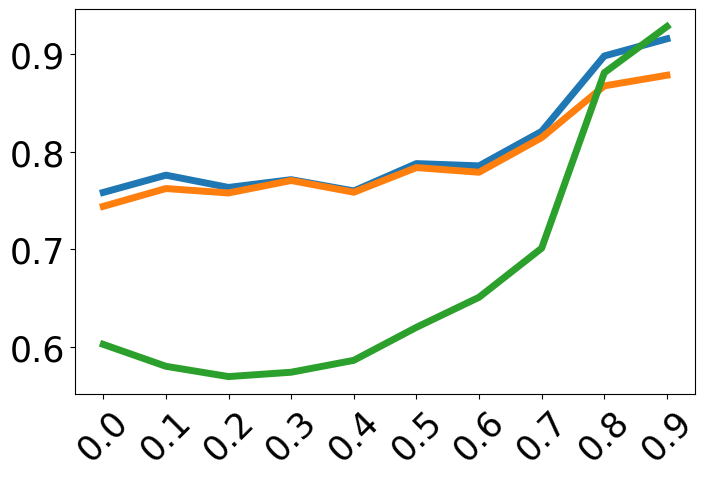} \\ 
     \multicolumn{2}{c}{Llama1B-En-De ID/OOD} \\
     \includegraphics[width=0.40\textwidth]{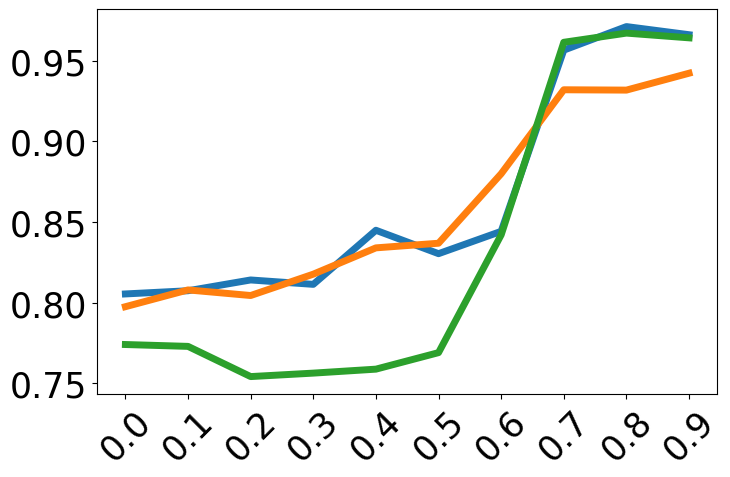} & \includegraphics[width=0.40\textwidth]{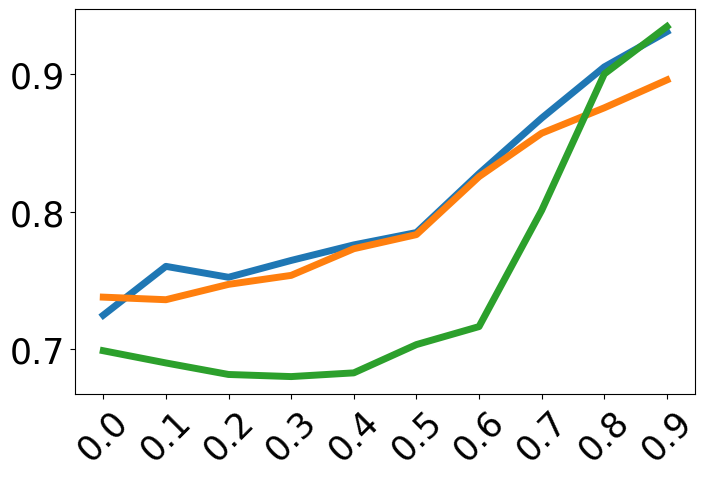} \\
    \multicolumn{2}{c}{Llama8B-En-Lt ID/OOD} \\[0.5em]
\end{tabular}
\includegraphics[width=\textwidth]{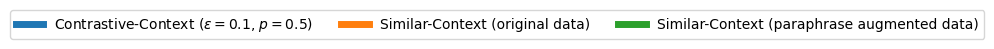}
\caption{\label{fig:dataAugment} Comparing \fsse\ on original data and on data augmented with paraphrases. X-axis is similarity of test target instance with in-context examples, and Y axis is accuracy.  \fsour\ uses paraphrases selectively to create contrast, whereas \fsse\ greedily selects Top-K most similar examples. While performance of \fsse\ improves in the high similarity range, it suffers in the low similarity range because IWL does not develop with highly similar context. }
\end{figure*}

\subsection{Experiments with various paraphrasing models}
Although our proposed method relies on an external paraphrasing model (specifically, \texttt{gemini-2.0-flash-lite}), we demonstrate that the choice of the paraphraser has negligible impact on overall performance. To validate this, we conduct additional experiments using two alternative models \texttt{gemini-2.5-flash-lite} as an example of a strong model, and \texttt{Qwen2.5-7B-Instruct} as an open source, possibly weaker model. The results in Figure~\ref{fig:varyParaModel} shows that \fsour\ remains stable regardless of whether the paraphrasing model is open-source or proprietary. 
\begin{figure*}[!htbp]
\centering
\begin{tabular}{cccc}
    \includegraphics[width=0.40\textwidth]{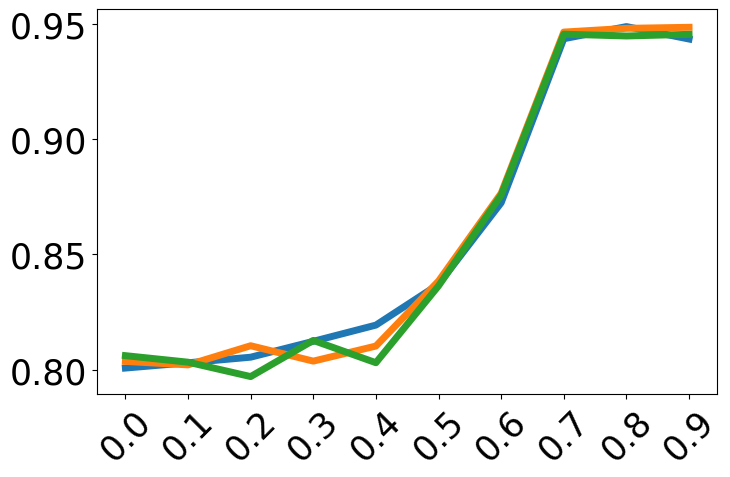} & \includegraphics[width=0.40\textwidth]{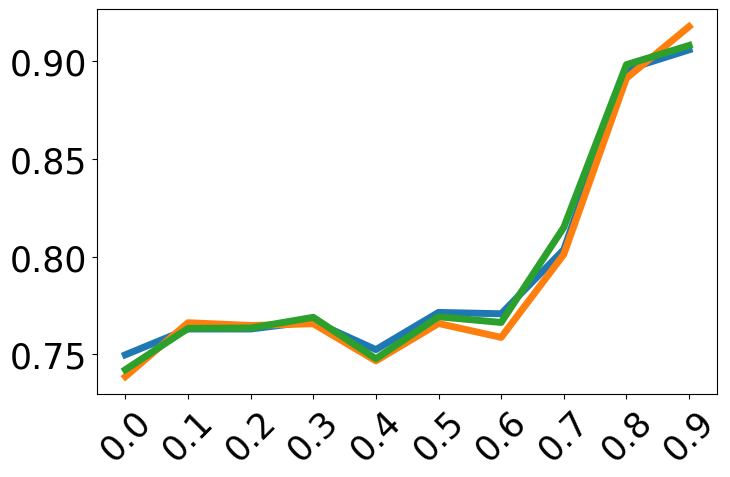} \\
    \multicolumn{2}{c}{Llama1B-En-De ID/OOD} \\
    \includegraphics[width=0.40\textwidth]{figures/contrast_comp_comet_sim_en-lt_8B_ID_para_model_ablation.png} & \includegraphics[width=0.40\textwidth]{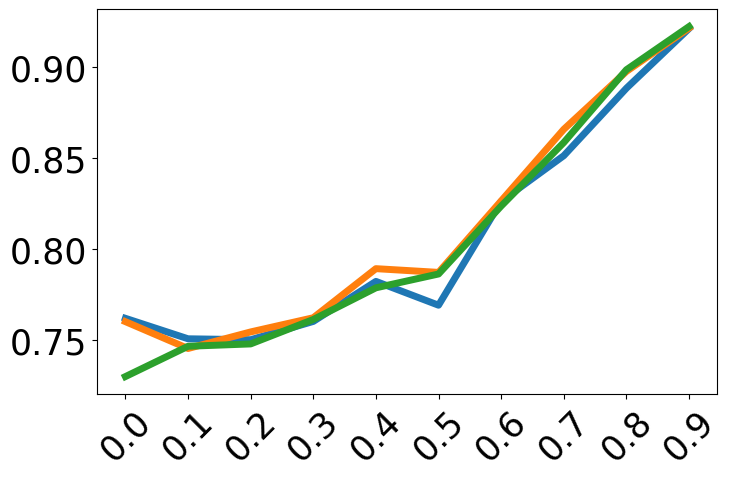} \\
    \multicolumn{2}{c}{Llama8B-En-Lt ID/OOD} \\[0.5em]
\end{tabular}
 \includegraphics[width=\textwidth]{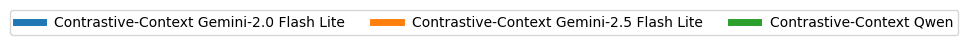}
 \caption{\label{fig:varyParaModel} Comparing \ours\ with highly similar examples created with various paraphrasing models. X-axis is target-example similarity and Y-axis is accuracy on the test data. \ours\ is robust to the choice of the paraphrasing model.}
\end{figure*}

\subsection{Effect of LoRA Target Modules}
\label{sec:appendix:ablations:lora_ablation}
We investigate the effect of applying LoRA to different subsets of transformer projections. In our main experiments, we keep the target modules as only the attention projection matrices (except the output projection matrix), but as an ablation, we also experiment by keeping all the attention projection matrices and the MLP layer matrices trainable. The comparison is shown in Figure~\ref{fig:ablationLoRA}. We observe that with additional trainable parameters, the overall performance increases, but the general trend of results across training regimes remains unchanged.

\begin{figure*}[!htbp]
\centering
\begin{tabular}{cccc}
    \includegraphics[width=0.40\textwidth]{figures_bar/comet_sim_en-lt_1B_ID.png} & \includegraphics[width=0.40\textwidth]{figures_bar/comet_sim_en-lt_1B_OOD.png} \\
    \multicolumn{2}{c}{Llama1B-En-Lt ID/OOD for training only attention projection weights} \\
    \includegraphics[width=0.40\textwidth]{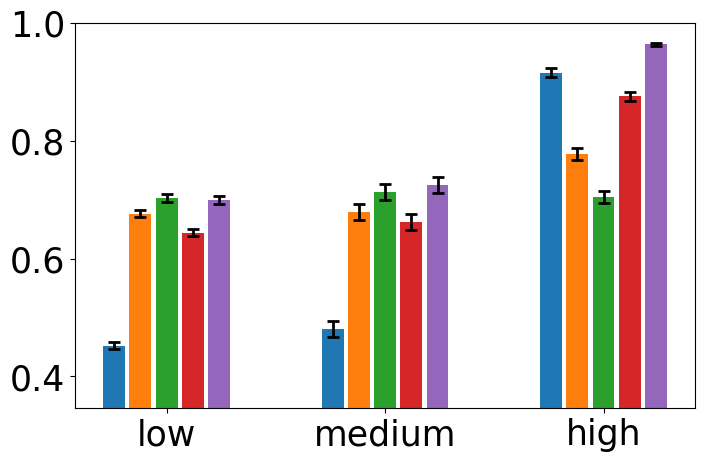} & \includegraphics[width=0.40\textwidth]{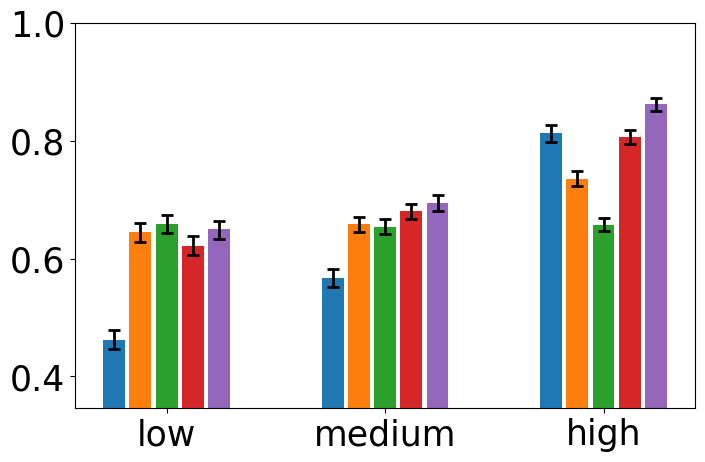} \\
    \multicolumn{2}{c}{Llama1B-En-Lt ID/OOD for training attention projection weights + MLP layer weights} \\[0.5em]
\end{tabular}
 \includegraphics[width=\textwidth]{figures_bar/comet_sim_legend.png}
 \caption{\label{fig:ablationLoRA} Comparing training only attention projection weights (except output projection) with training attention projection weights + MLP layer weights. X-axis: Level of maximum similarity of target with in-context examples. The similarity ranges here are - Low: $0-0.33$, Medium: $0.33-0.67$, High: $0.67-1$. Y-axis: Accuracy (COMET score).}
\end{figure*}
}

\section{Additional Experiments Comparing Different Training Strategies}
We present results of more model-language pair combinations that could not be fit in Figure~\ref{fig:accuracy} of the main paper in Figure~\ref{fig:additional_training_strategies}.
\label{sec:appendix:fig1}
\begin{figure*}[!htbp]
\centering
\begin{tabular}{cccc}
    \includegraphics[width=0.24\textwidth]{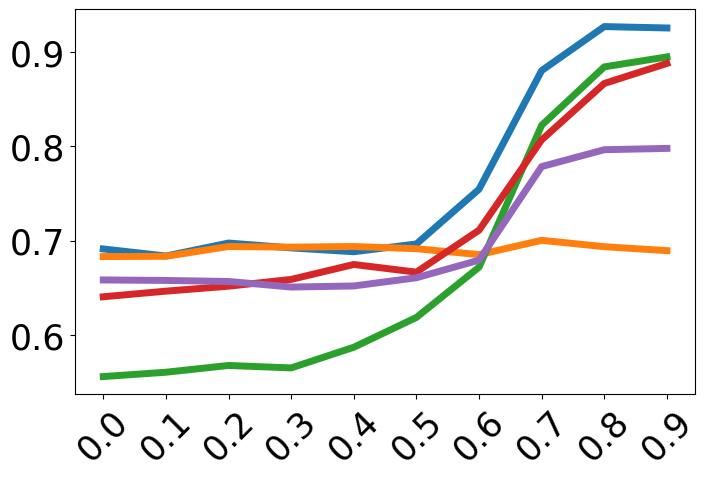} & \includegraphics[width=0.24\textwidth]{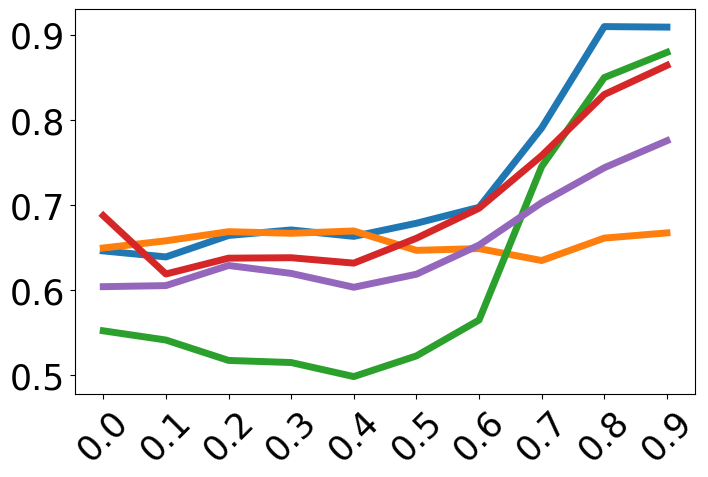} & \includegraphics[width=0.24\textwidth]{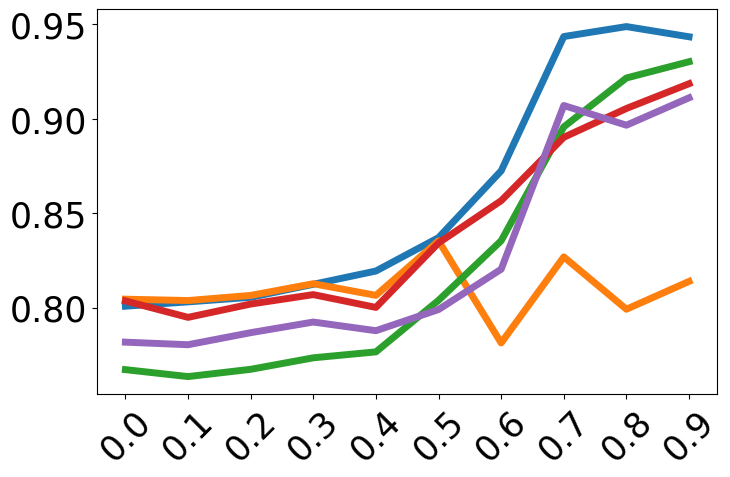} & \includegraphics[width=0.24\textwidth]{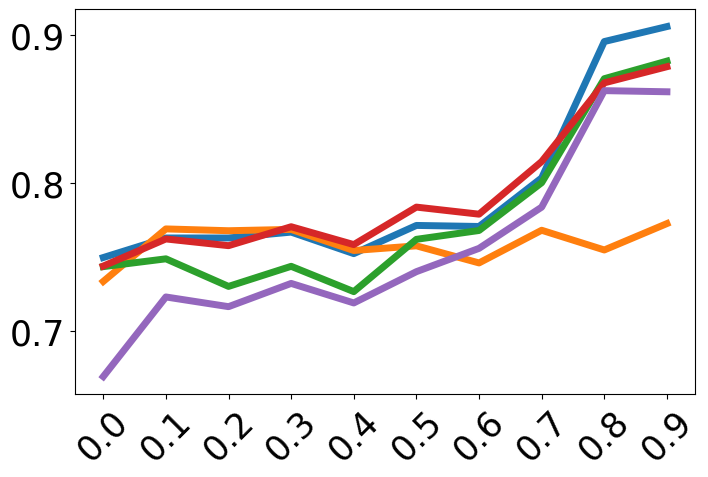} \\
    \multicolumn{2}{c}{Llama1B-En-Hi ID/OOD} & \multicolumn{2}{c}{Llama1B-En-De ID/OOD} \\[0.5em]
\end{tabular}
\end{figure*}

\begin{figure*}[!htbp]
\centering
\begin{tabular}{cccc}
    \includegraphics[width=0.24\textwidth]{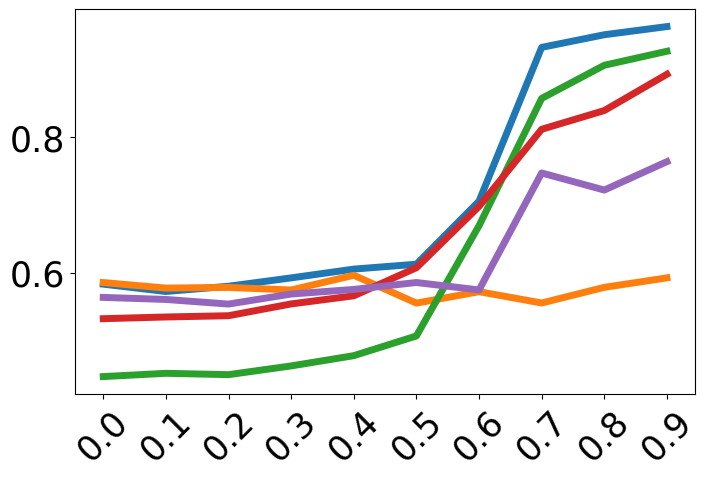} & \includegraphics[width=0.24\textwidth]{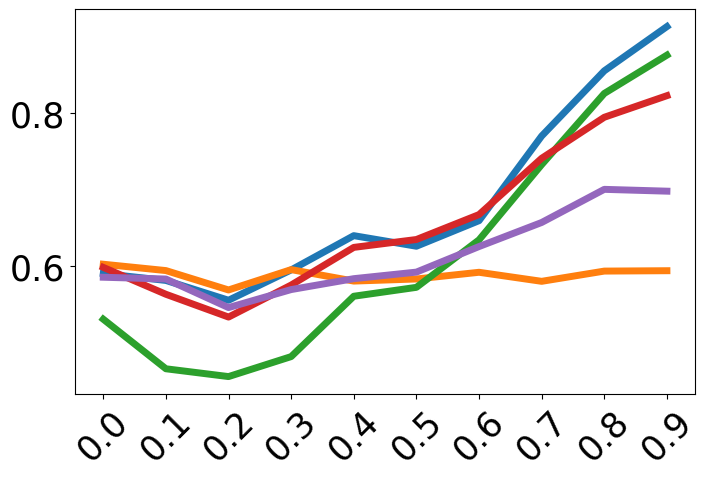} & \includegraphics[width=0.24\textwidth]{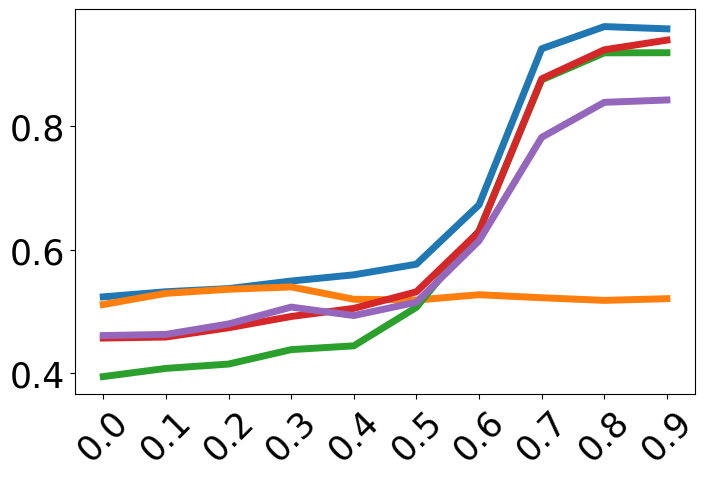} & \includegraphics[width=0.24\textwidth]{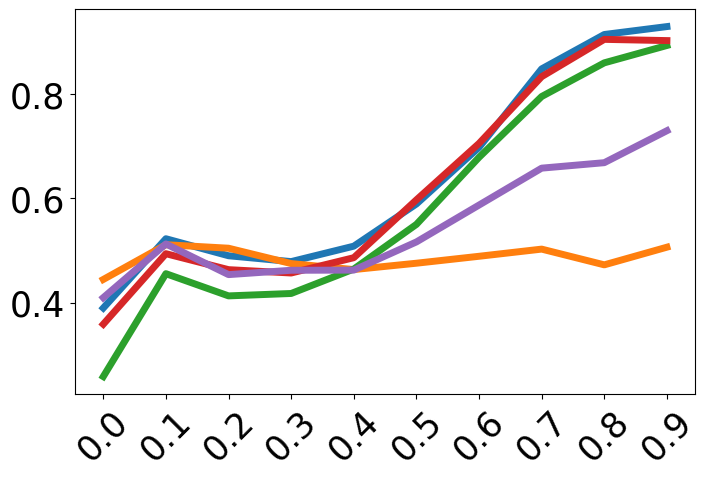} \\
    \multicolumn{2}{c}{Llama1B-En-Lt ID/OOD} & \multicolumn{2}{c}{Llama1B-En-Ta ID/OOD} \\[0.5em]
\end{tabular}
\end{figure*}

\begin{figure*}[!htbp]
\centering
\begin{tabular}{cccc}
    \includegraphics[width=0.24\textwidth]{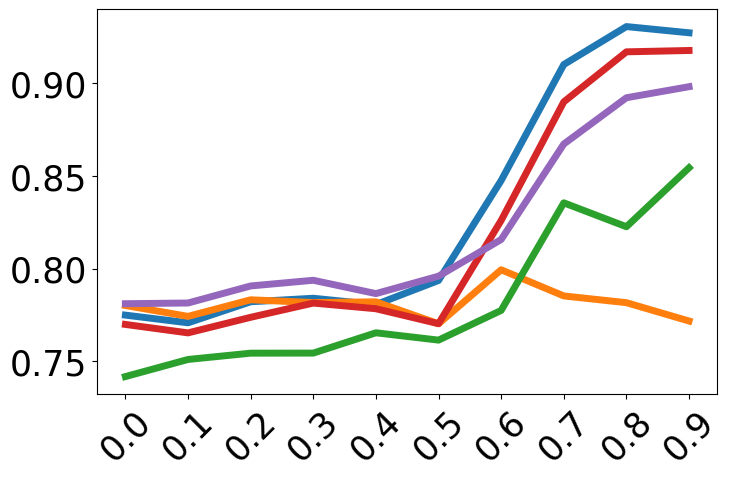} & \includegraphics[width=0.24\textwidth]{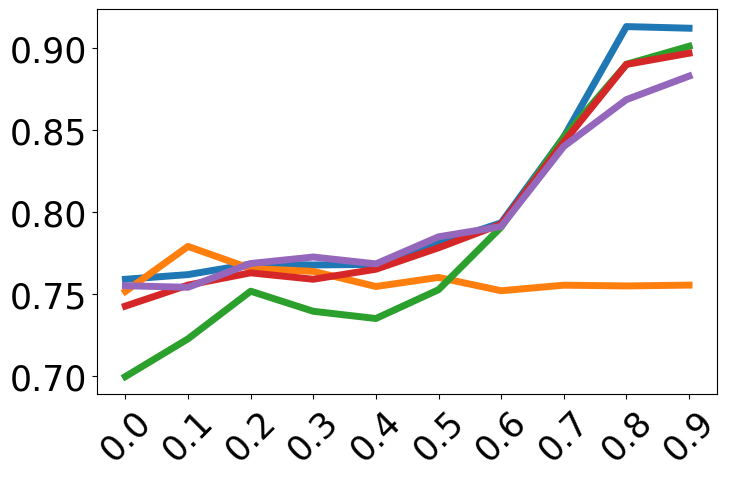} & \includegraphics[width=0.24\textwidth]{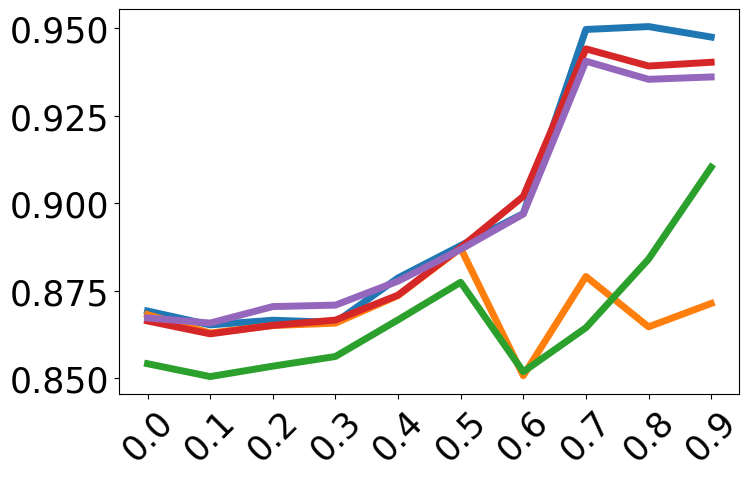} & \includegraphics[width=0.24\textwidth]{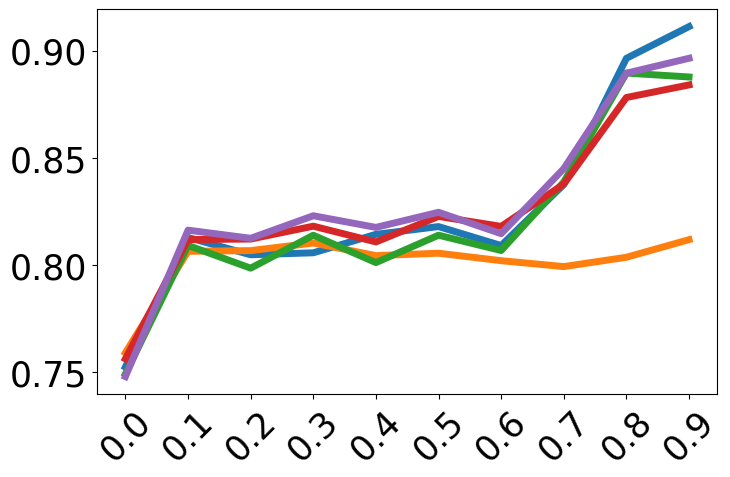} \\
    \multicolumn{2}{c}{Llama8B-En-Hi ID/OOD} & \multicolumn{2}{c}{Llama8B-En-De ID/OOD} \\[0.5em]
\end{tabular}
\end{figure*}

\begin{figure*}[!htbp]
\centering
\begin{tabular}{cccc}
    \includegraphics[width=0.24\textwidth]{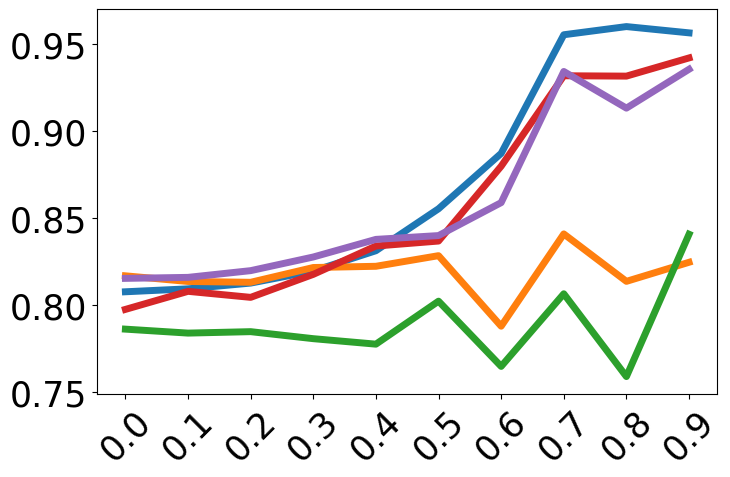} & \includegraphics[width=0.24\textwidth]{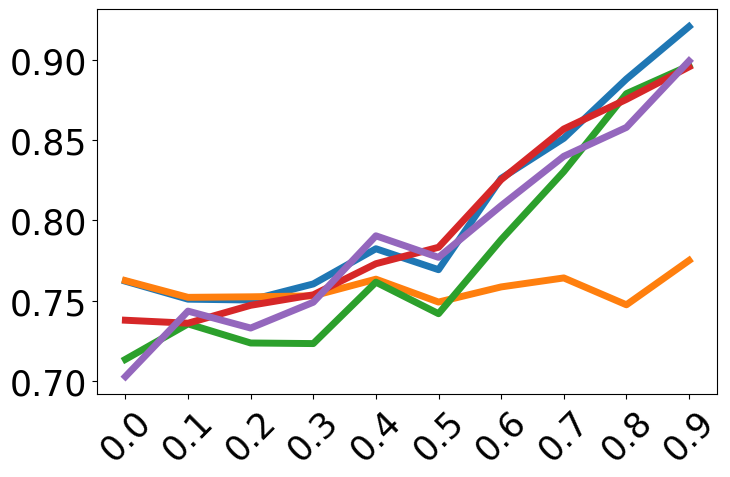} & \includegraphics[width=0.24\textwidth]{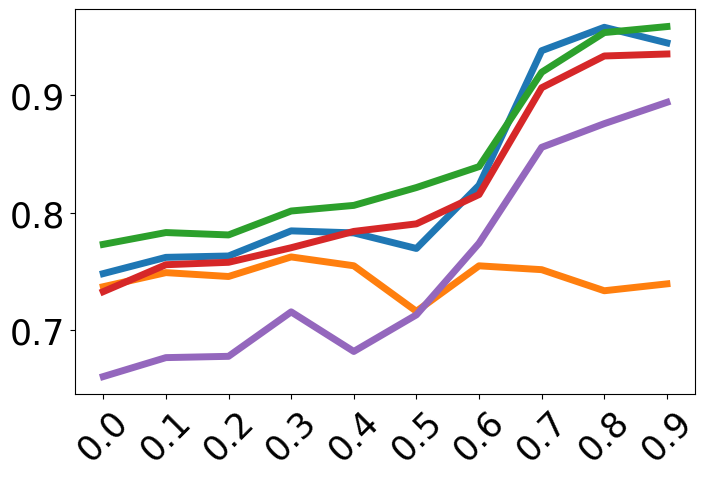} & \includegraphics[width=0.24\textwidth]{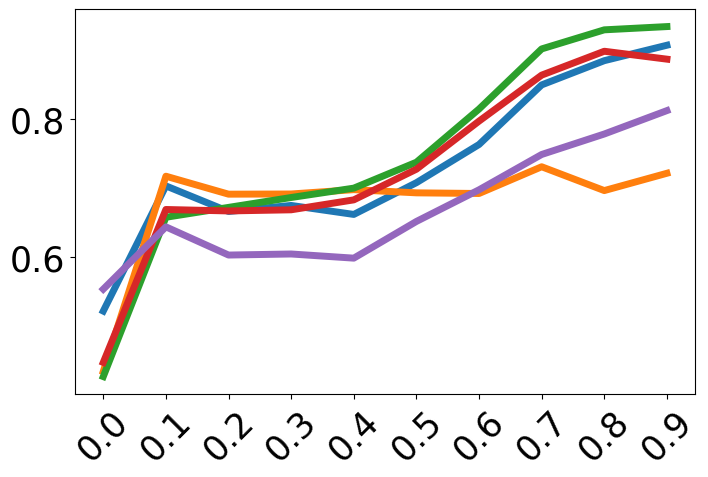} \\
    \multicolumn{2}{c}{Llama8B-En-Lt ID/OOD} & \multicolumn{2}{c}{Llama8B-En-Ta ID/OOD} \\[0.5em]
\end{tabular}
\end{figure*}

\begin{figure*}[!htbp]
\centering
\begin{tabular}{cccc}
    \includegraphics[width=0.24\textwidth]{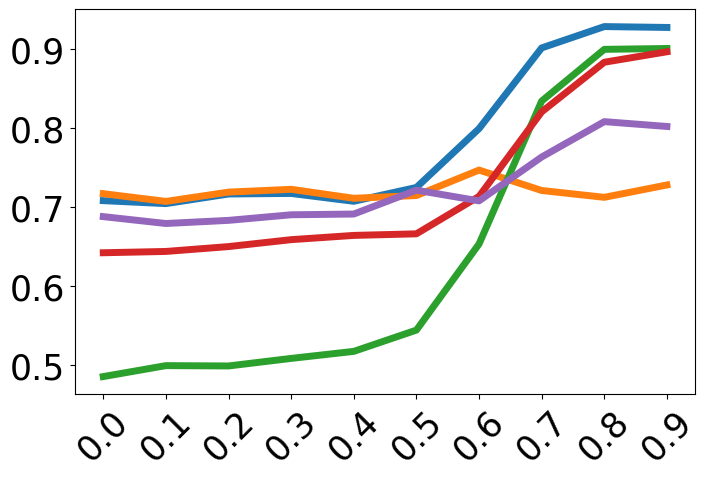} & \includegraphics[width=0.24\textwidth]{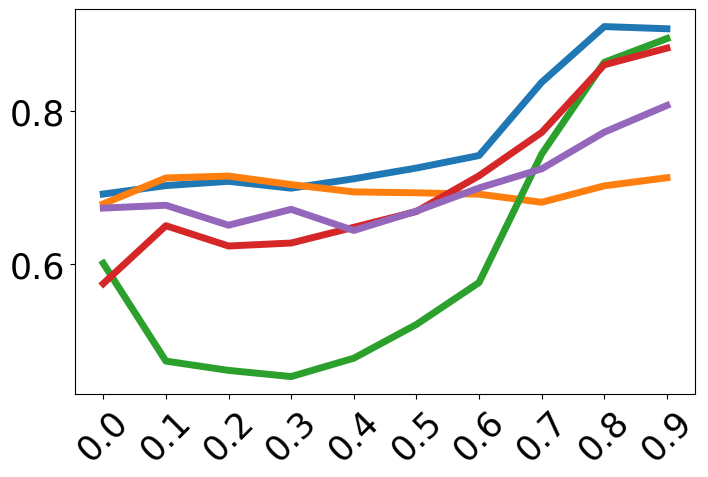} & \includegraphics[width=0.24\textwidth]{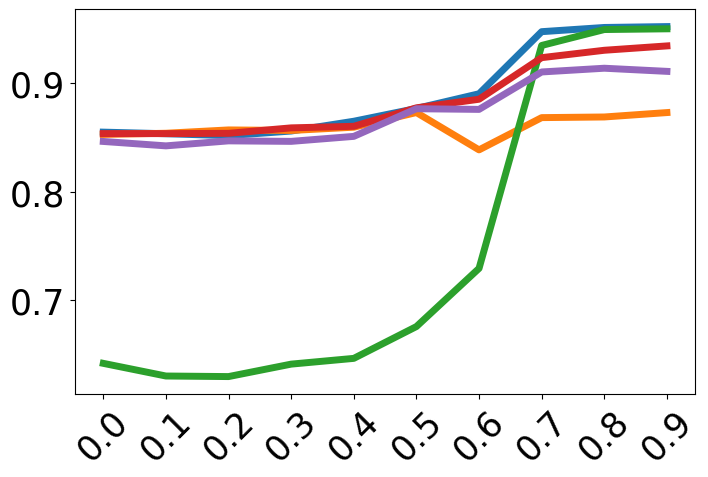} & \includegraphics[width=0.24\textwidth]{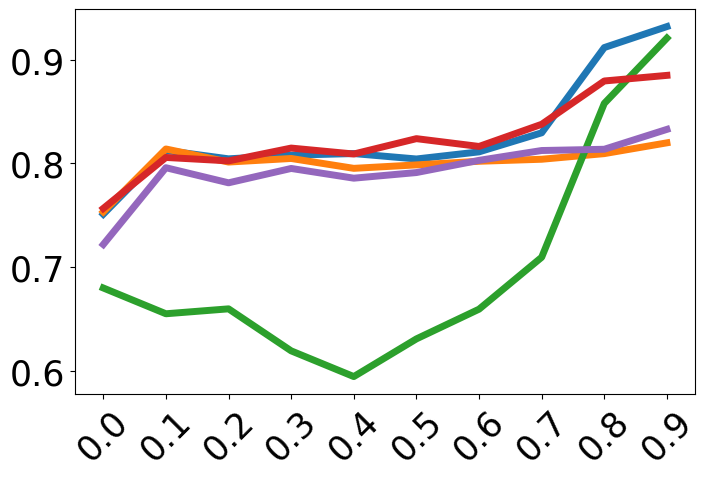} \\
    \multicolumn{2}{c}{Mistral-En-Hi ID/OOD} & \multicolumn{2}{c}{Mistral-En-De ID/OOD} \\[0.5em]
\end{tabular}
\end{figure*}

\begin{figure*}[!htbp]
\centering
\begin{tabular}{cccc}
    \includegraphics[width=0.24\textwidth]{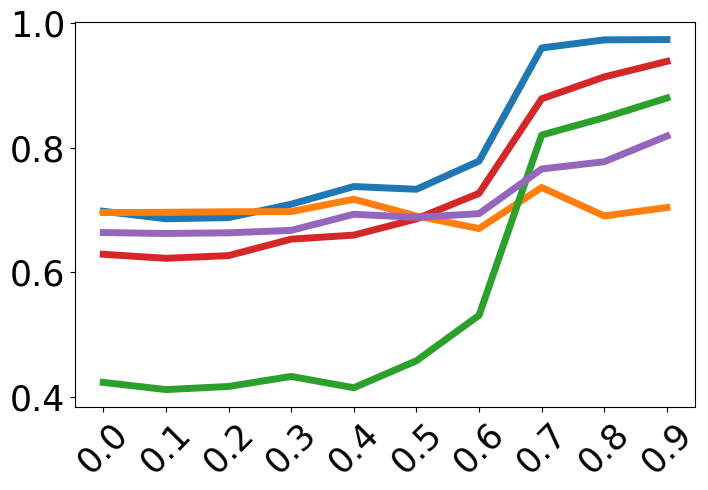} & \includegraphics[width=0.24\textwidth]{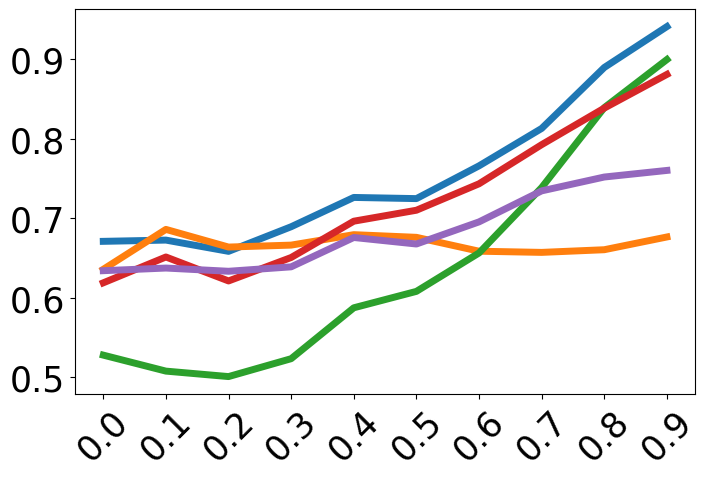} & \includegraphics[width=0.24\textwidth]{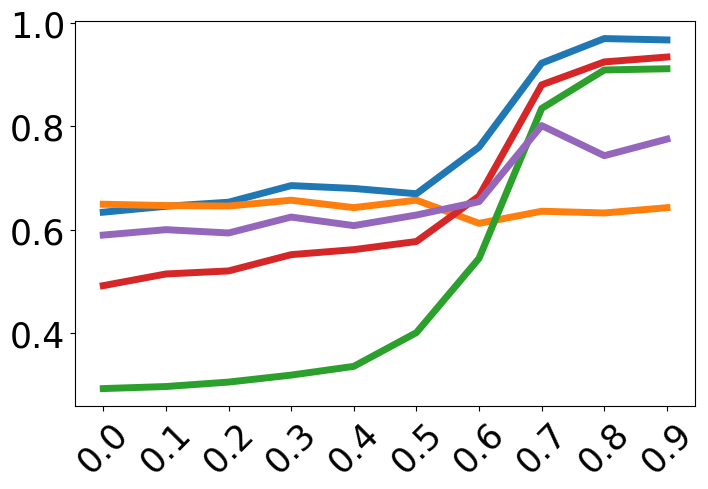} & \includegraphics[width=0.24\textwidth]{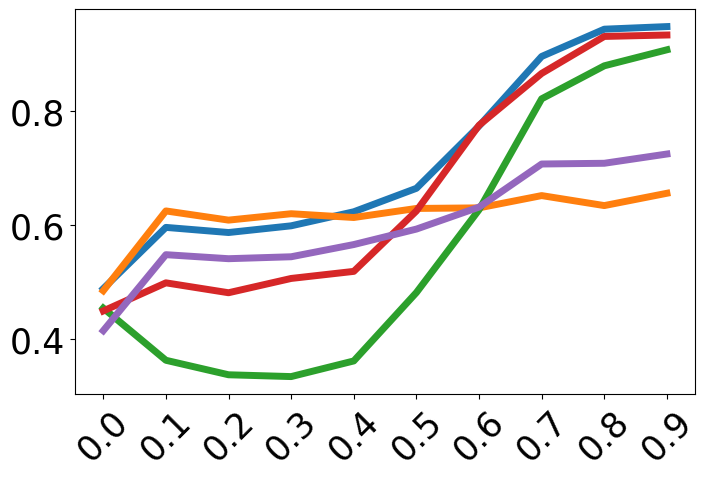} \\
    \multicolumn{2}{c}{Mistral-En-Lt ID/OOD} & \multicolumn{2}{c}{Mistral-En-Ta ID/OOD} \\[0.5em]
\end{tabular}
\end{figure*}

\begin{figure*}[!htbp]
\centering
\begin{tabular}{cccc}
    \includegraphics[width=0.24\textwidth]{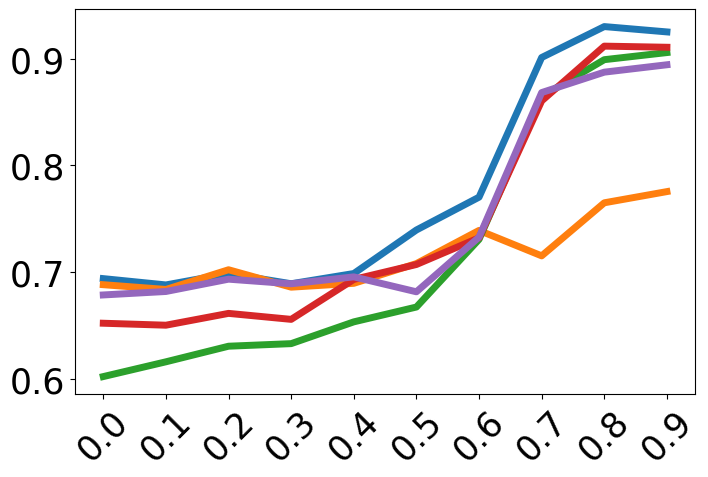} & \includegraphics[width=0.24\textwidth]{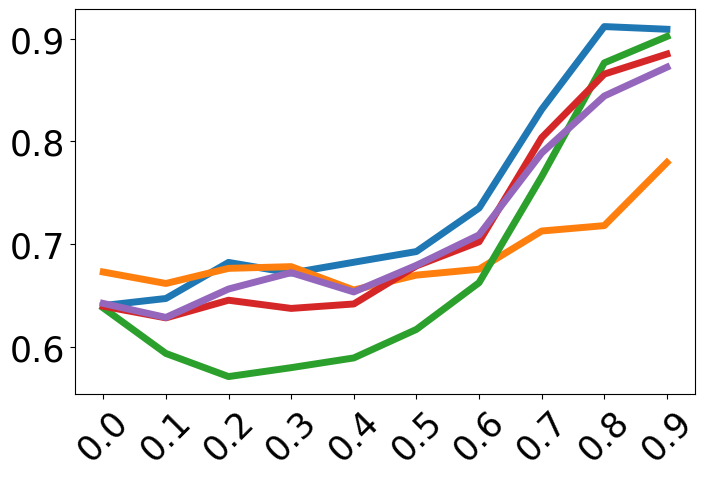} & \includegraphics[width=0.24\textwidth]{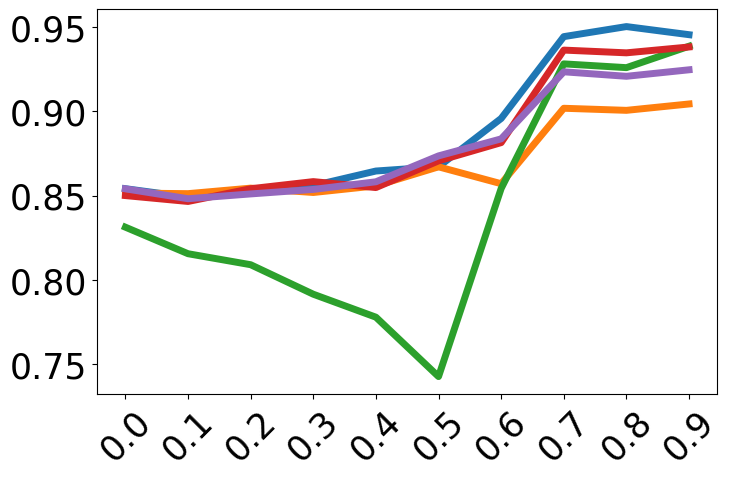} & \includegraphics[width=0.24\textwidth]{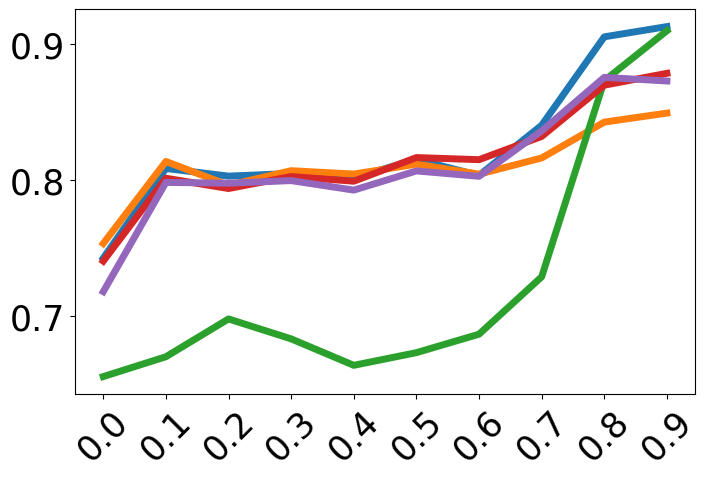} \\
    \multicolumn{2}{c}{Qwen-En-Hi ID/OOD} & \multicolumn{2}{c}{Qwen-En-De ID/OOD} \\[0.5em]
\end{tabular}
\end{figure*}

\begin{figure*}[!htbp]
\centering
\begin{tabular}{cccc}
    \includegraphics[width=0.24\textwidth]{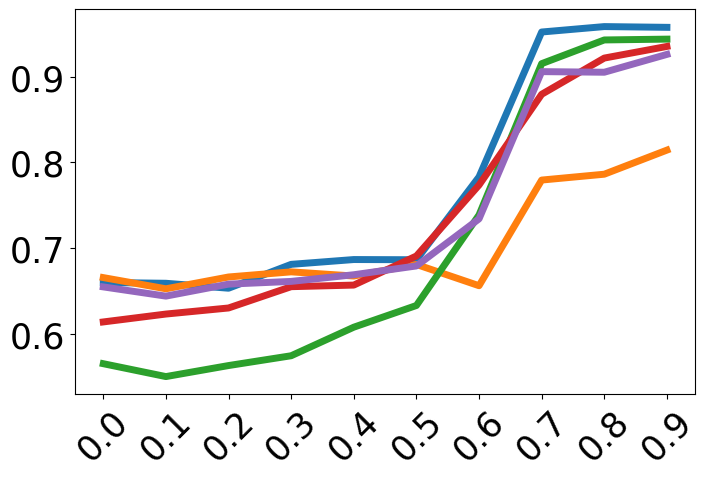} & \includegraphics[width=0.24\textwidth]{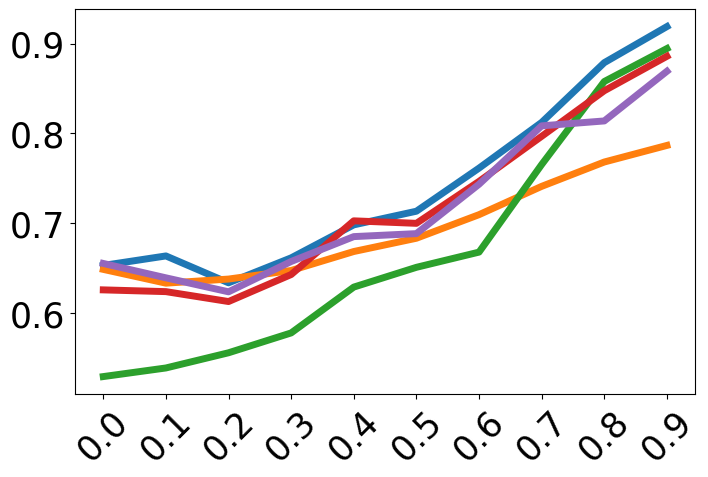} & \includegraphics[width=0.24\textwidth]{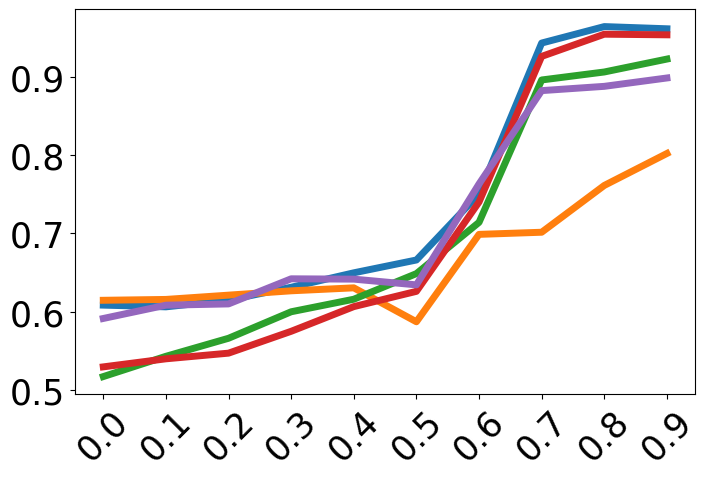} & \includegraphics[width=0.24\textwidth]{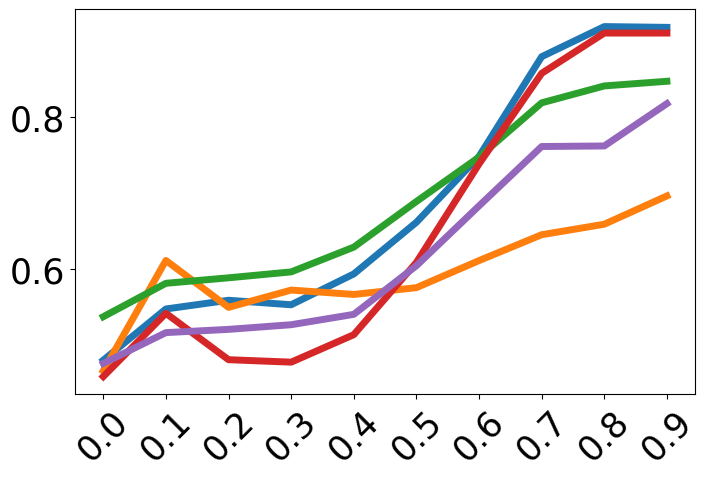} \\
    \multicolumn{2}{c}{Qwen-En-Lt ID/OOD} & \multicolumn{2}{c}{Qwen-En-Ta ID/OOD} \\[0.5em]
\end{tabular}
\end{figure*}

\begin{figure*}[!htbp]
\centering
    \includegraphics[width=\textwidth]{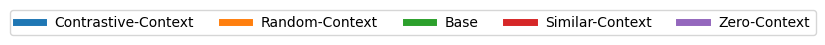}
    \caption{\label{fig:additional_training_strategies} All 32 plots for four different models tested on four different language pairs with ID and OOD distribution. X-axis is target-example similarity and Y-axis is accuracy.}
\end{figure*}

\newpage
\section{Additional Experiments on Learning Dynamics}
We present results of more model-language pair combinations that could not be fit in Figure~\ref{fig:emergence} of the main paper in Figure~\ref{fig:additonal_learning_dynamics}.
\label{sec:appendix:fig3}
\begin{figure*}[!htbp]
\centering
\begin{tabular}{ccc}
    \includegraphics[width=0.3\textwidth]{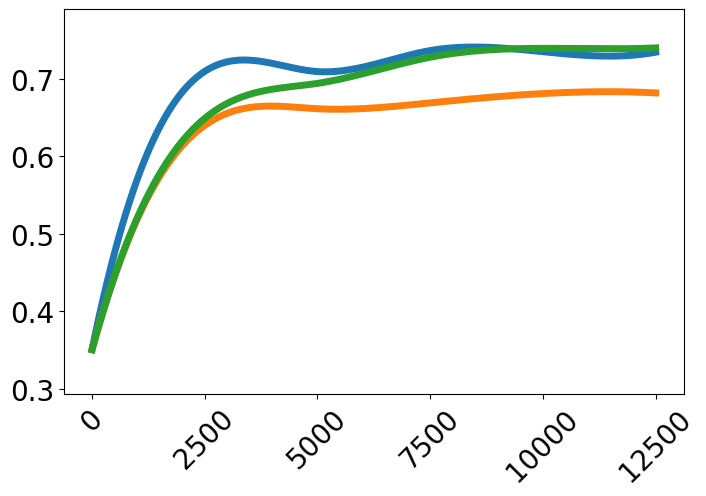} & \includegraphics[width=0.3\textwidth]{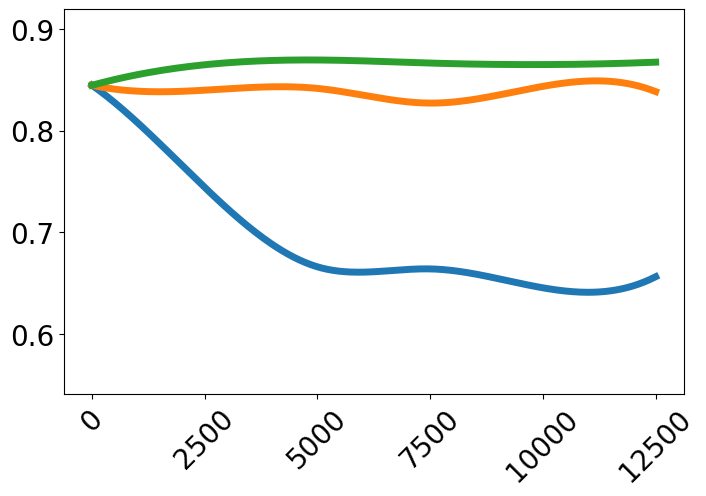} & \includegraphics[width=0.3\textwidth]{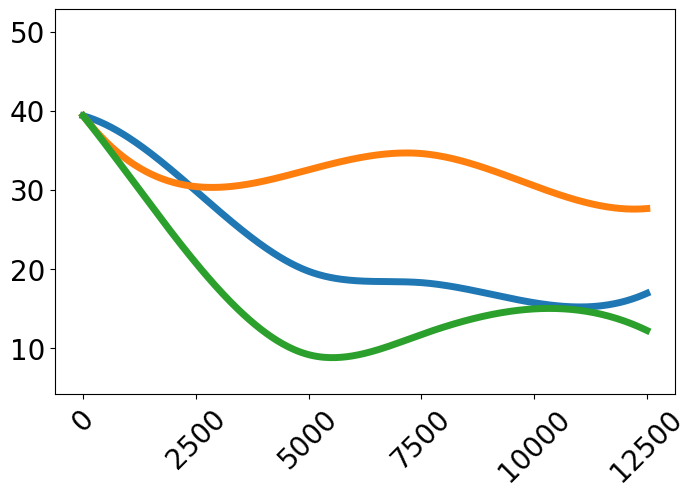} \\
    IWL score: 1B-En-Hi-ID & ICL score: 1B-En-Hi-ID & Copy score: 1B-En-Hi-ID \\[0.5em]
\end{tabular}
\end{figure*}

\begin{figure*}[!htbp]
\centering
\begin{tabular}{ccc}
    \includegraphics[width=0.3\textwidth]{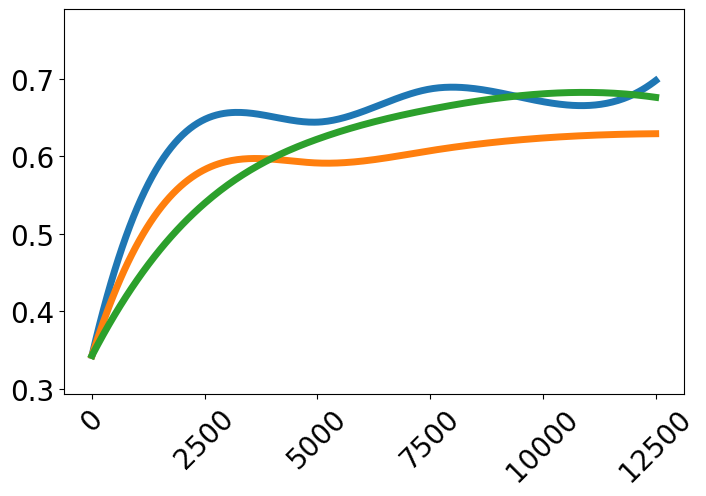} & \includegraphics[width=0.3\textwidth]{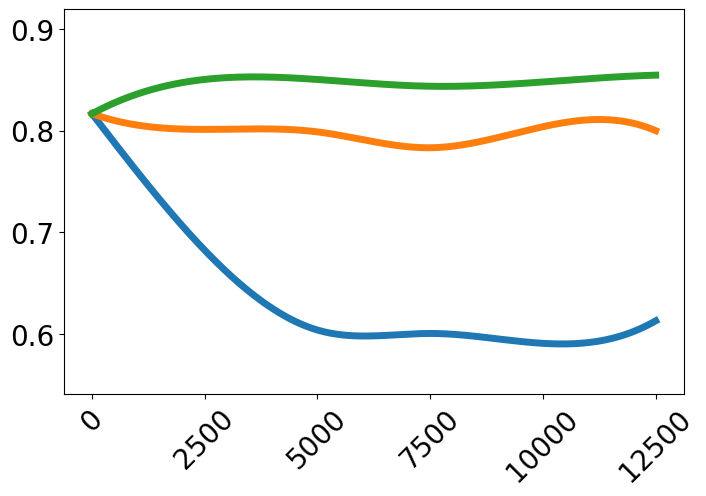} & \includegraphics[width=0.3\textwidth]{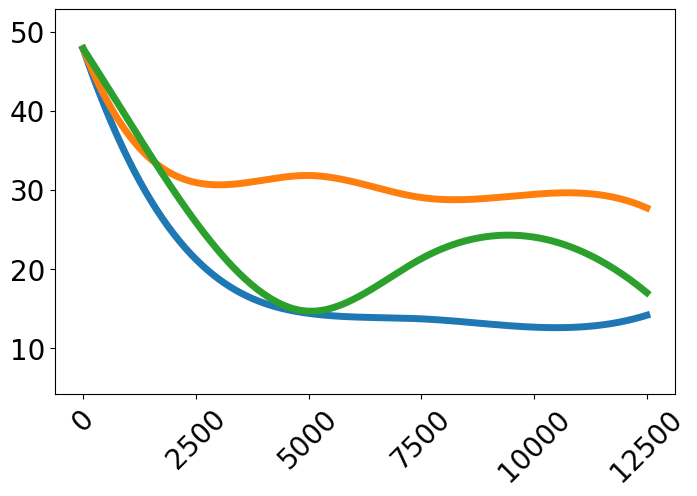} \\
    IWL score: 1B-En-Hi-OOD & ICL score: 1B-En-Hi-OOD & Copy score: 1B-En-Hi-OOD \\[0.5em]
\end{tabular}
\end{figure*}

\begin{figure*}[!htbp]
\centering
\begin{tabular}{ccc}
    \includegraphics[width=0.3\textwidth]{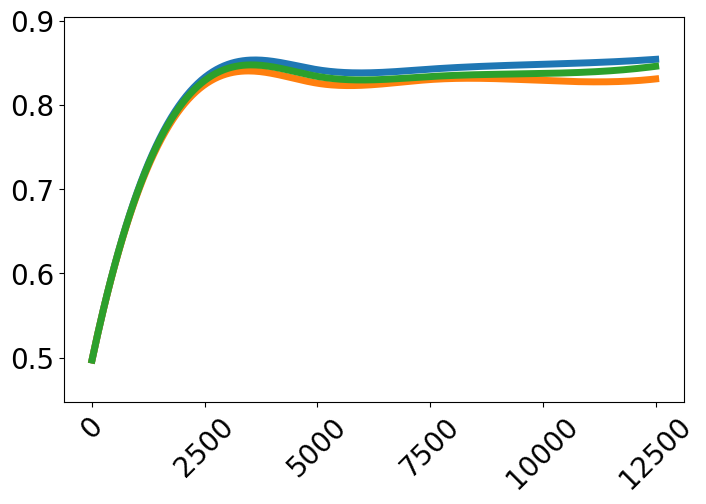} & \includegraphics[width=0.3\textwidth]{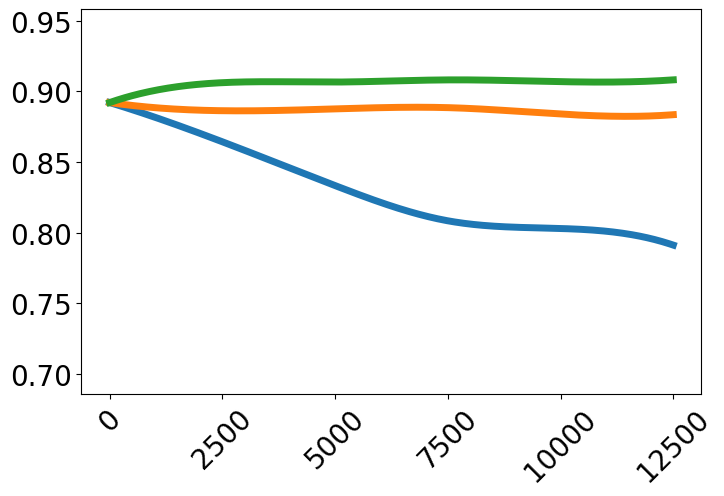} & \includegraphics[width=0.3\textwidth]{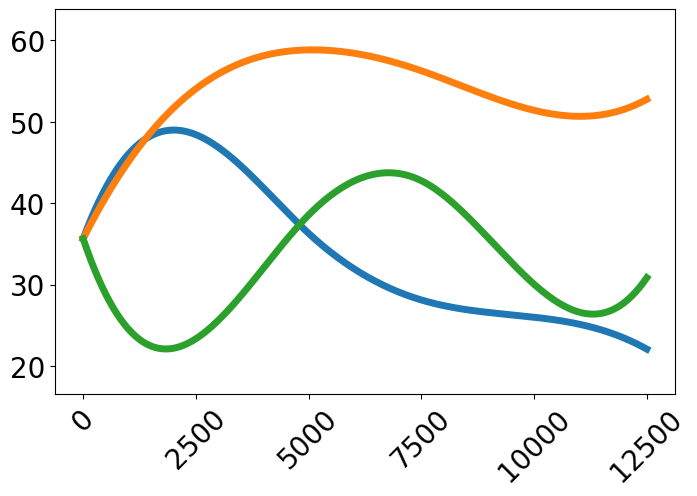} \\
    IWL score: 1B-En-De-ID & ICL score: 1B-En-De-ID & Copy score: 1B-En-De-ID \\[0.5em]
\end{tabular}
\end{figure*}

\begin{figure*}[!htbp]
\centering
\begin{tabular}{ccc}
    \includegraphics[width=0.3\textwidth]{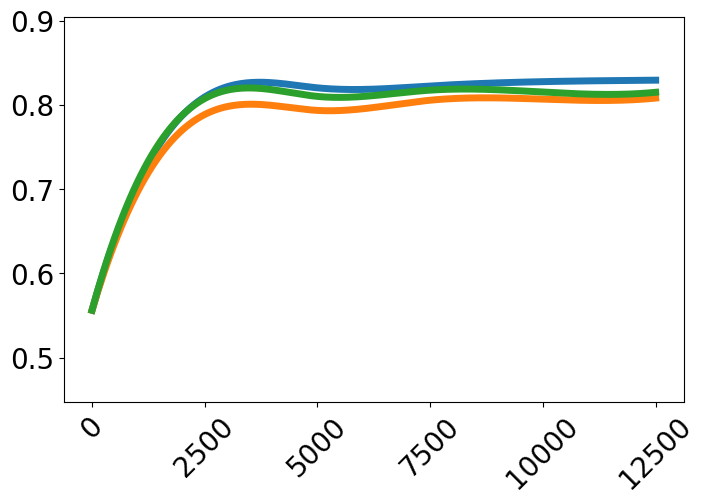} & \includegraphics[width=0.3\textwidth]{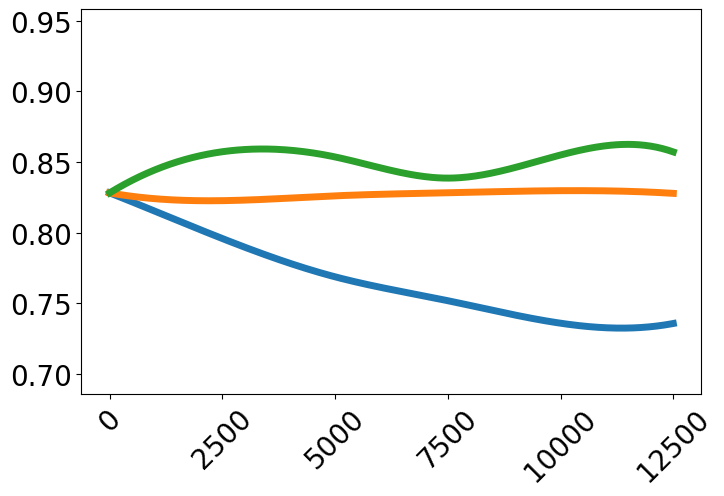} & \includegraphics[width=0.3\textwidth]{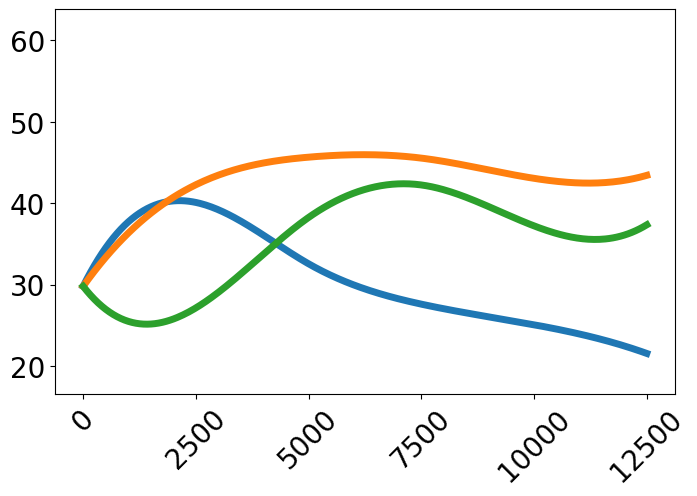} \\
    IWL score: 1B-En-De-OOD & ICL score: 1B-En-De-OOD & Copy score: 1B-En-De-OOD \\[0.5em]
\end{tabular}
\end{figure*}

\begin{figure*}[!htbp]
\centering
\begin{tabular}{ccc}
    \includegraphics[width=0.3\textwidth]{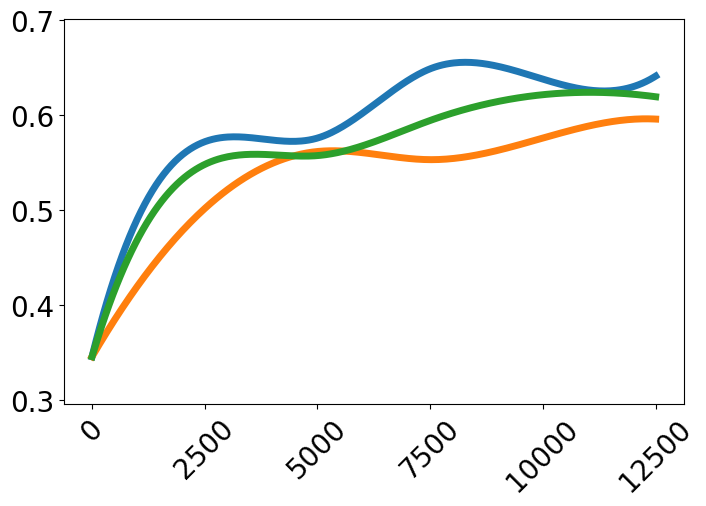} & \includegraphics[width=0.3\textwidth]{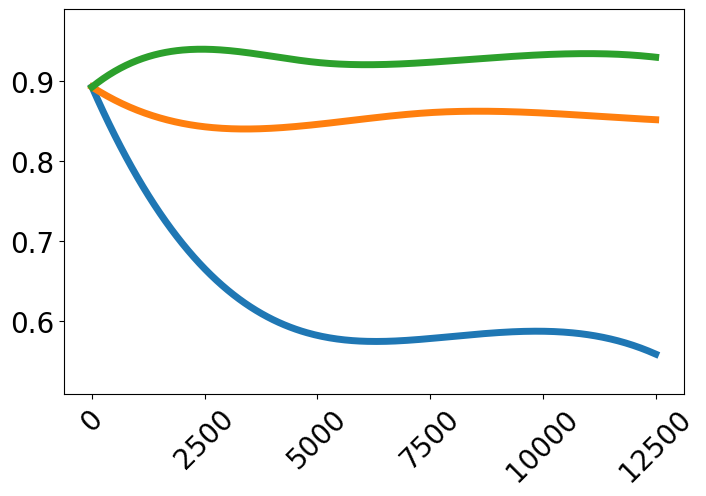} & \includegraphics[width=0.3\textwidth]{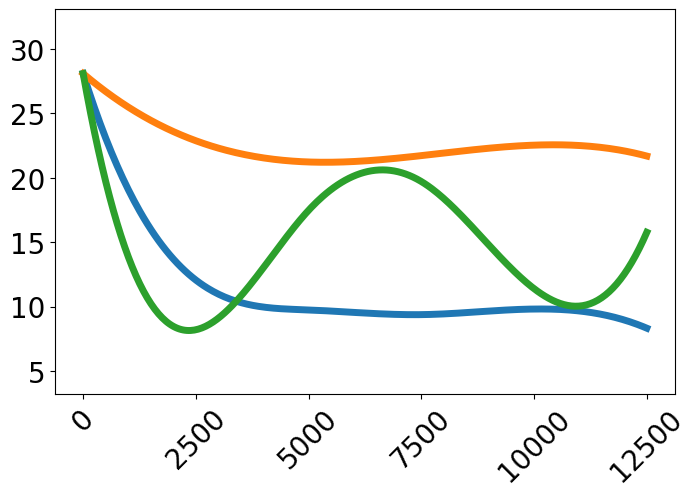} \\
    IWL score: 1B-En-Lt-ID & ICL score: 1B-En-Lt-ID & Copy score: 1B-En-Lt-ID \\[0.5em]
\end{tabular}
\end{figure*}

\begin{figure*}[!htbp]
\centering
\begin{tabular}{ccc}
    \includegraphics[width=0.3\textwidth]{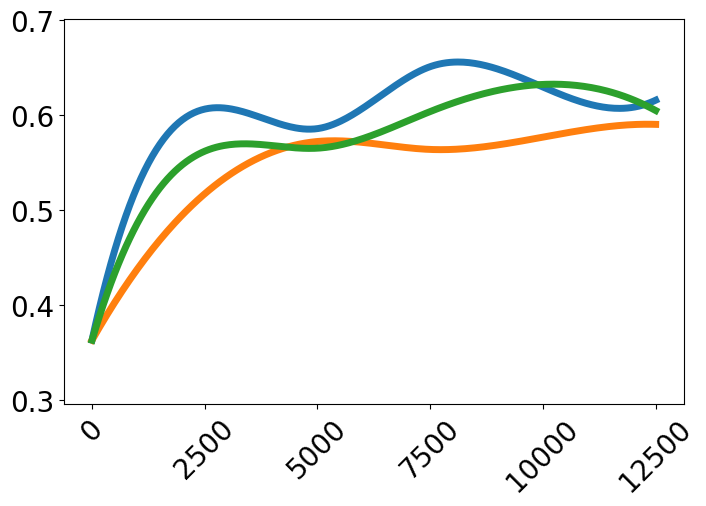} & \includegraphics[width=0.3\textwidth]{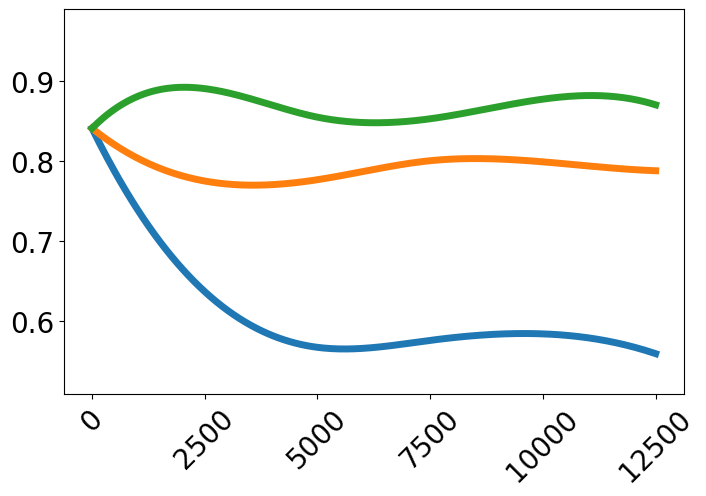} & \includegraphics[width=0.3\textwidth]{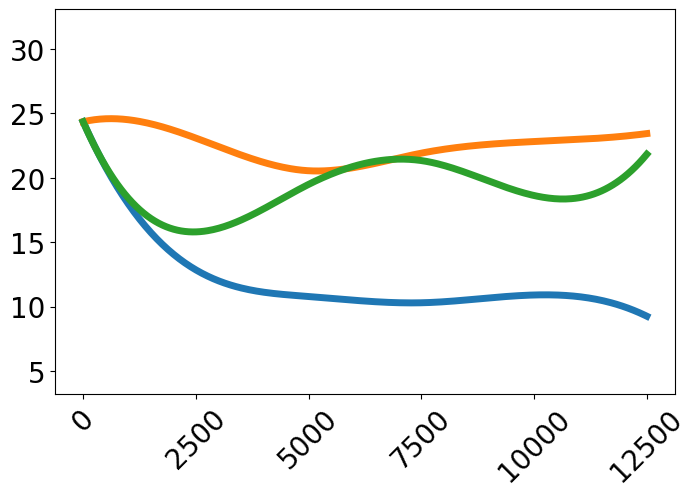} \\
    IWL score: 1B-En-Lt-OOD & ICL score: 1B-En-Lt-OOD & Copy score: 1B-En-Lt-OOD \\[0.5em]
\end{tabular}
\end{figure*}

\begin{figure*}[!htbp]
\centering
\begin{tabular}{ccc}
    \includegraphics[width=0.3\textwidth]{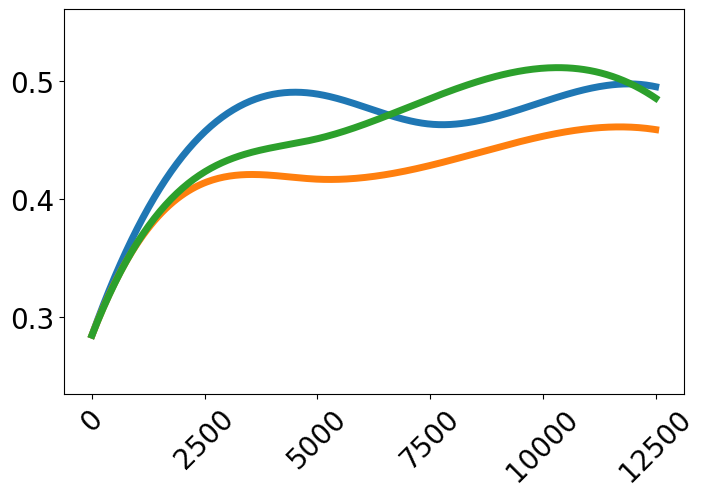} & \includegraphics[width=0.3\textwidth]{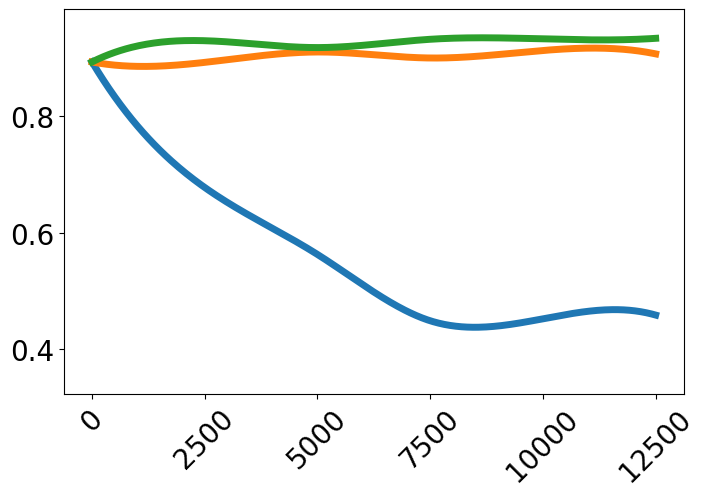} & \includegraphics[width=0.3\textwidth]{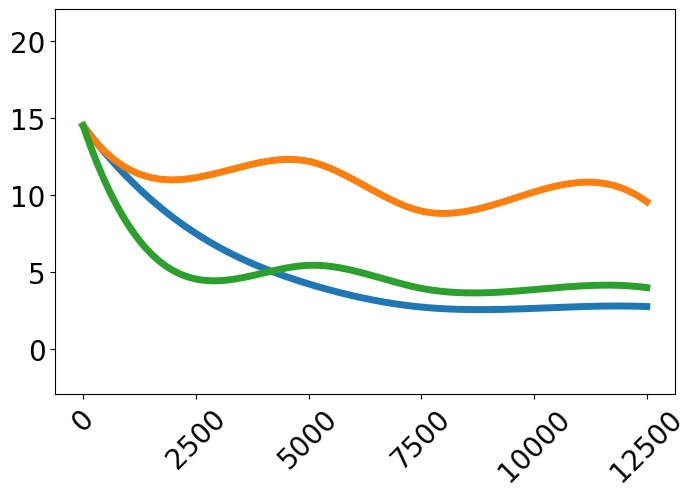} \\
    IWL score: 1B-En-Ta-ID & ICL score: 1B-En-Ta-ID & Copy score: 1B-En-Ta-ID \\[0.5em]
\end{tabular}
\end{figure*}

\begin{figure*}[!htbp]
\centering
\begin{tabular}{ccc}
    \includegraphics[width=0.3\textwidth]{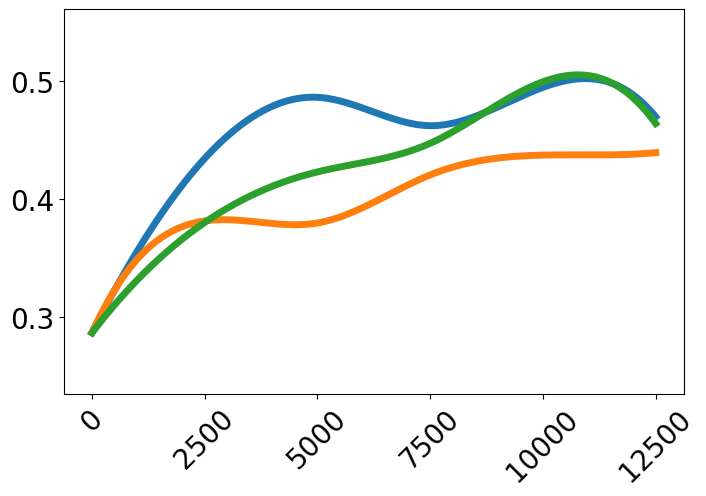} & \includegraphics[width=0.3\textwidth]{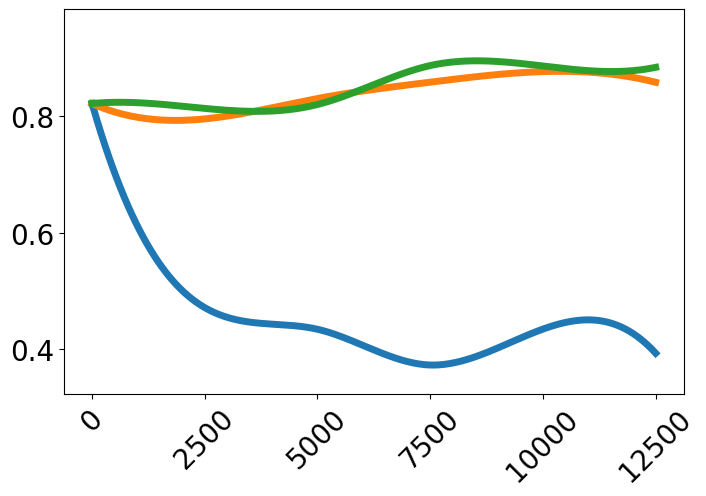} & \includegraphics[width=0.3\textwidth]{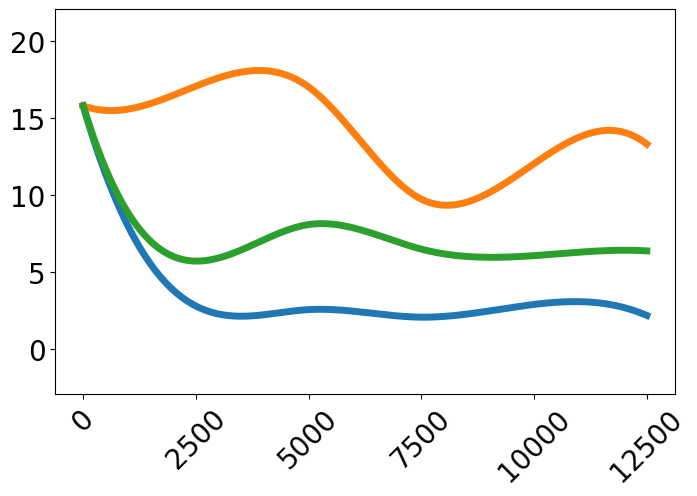} \\
    IWL score: 1B-En-Ta-OOD & ICL score: 1B-En-Ta-OOD & Copy score: 1B-En-Ta-OOD \\[0.5em]
\end{tabular}
\end{figure*}

\begin{figure*}[!htbp]
\centering
\begin{tabular}{ccc}
    \includegraphics[width=0.3\textwidth]{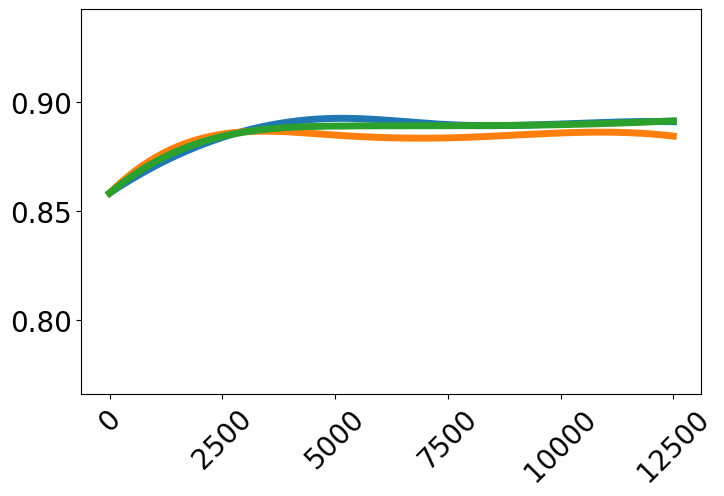} & \includegraphics[width=0.3\textwidth]{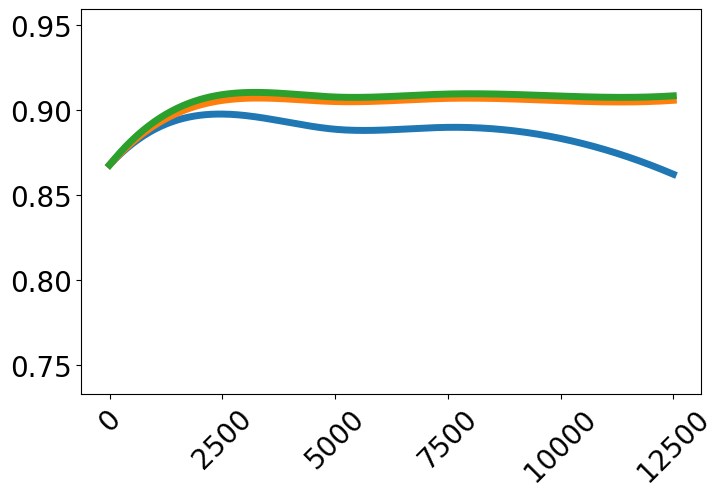} & \includegraphics[width=0.3\textwidth]{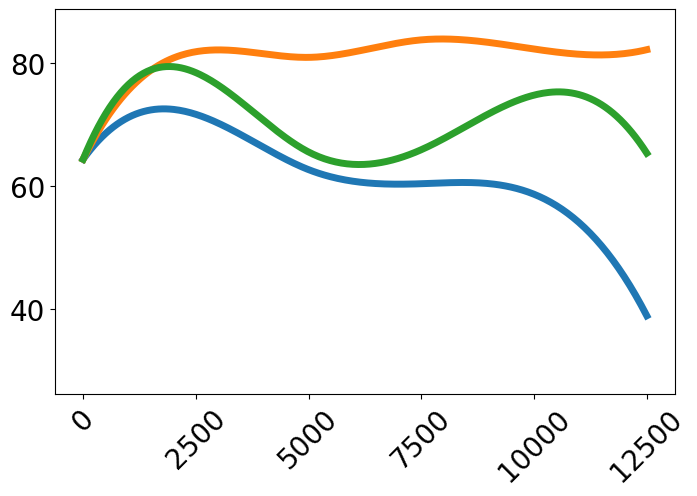} \\
    IWL score: 8B-En-De-ID & ICL score: 8B-En-De-ID & Copy score: 8B-En-De-ID \\[0.5em]
\end{tabular}
\end{figure*}

\begin{figure*}[!htbp]
\centering
\begin{tabular}{ccc}
    \includegraphics[width=0.3\textwidth]{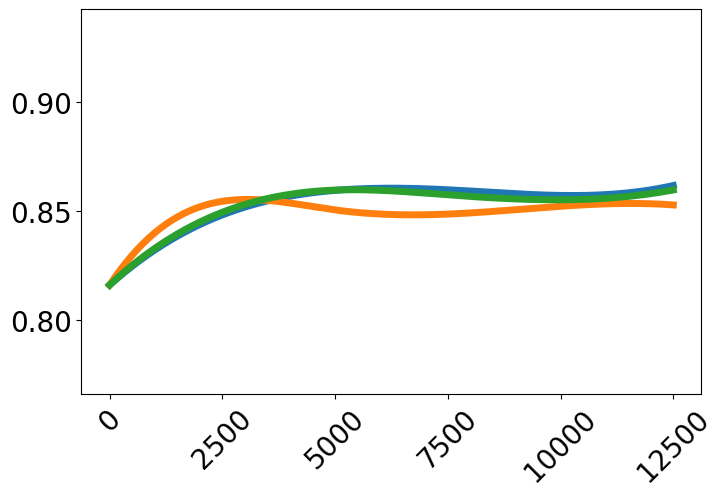} & \includegraphics[width=0.3\textwidth]{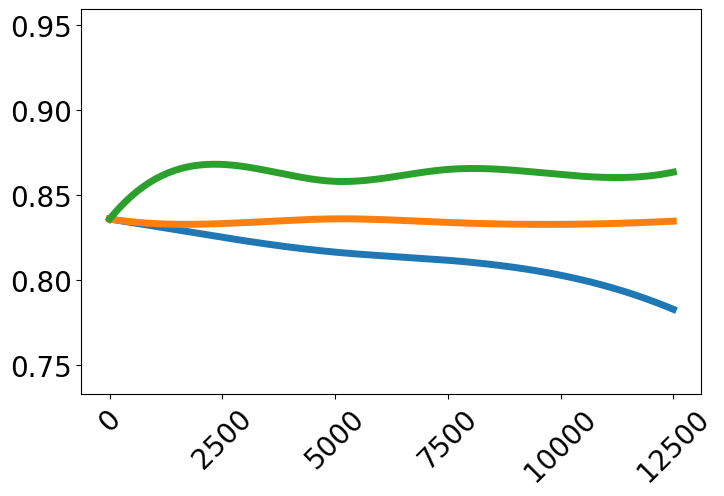} & \includegraphics[width=0.3\textwidth]{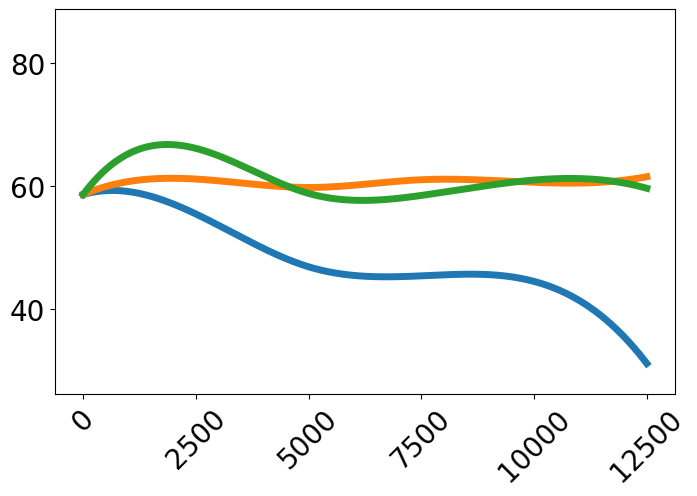} \\
    IWL score: 8B-En-De-OOD & ICL score: 8B-En-De-OOD & Copy score: 8B-En-De-OOD \\[0.5em]
\end{tabular}
\end{figure*}

\begin{figure*}[!htbp]
\centering
\begin{tabular}{ccc}
    \includegraphics[width=0.3\textwidth]{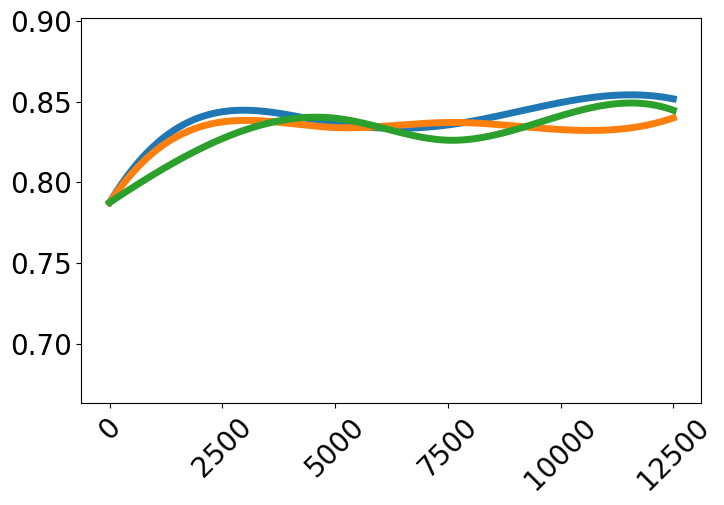} & \includegraphics[width=0.3\textwidth]{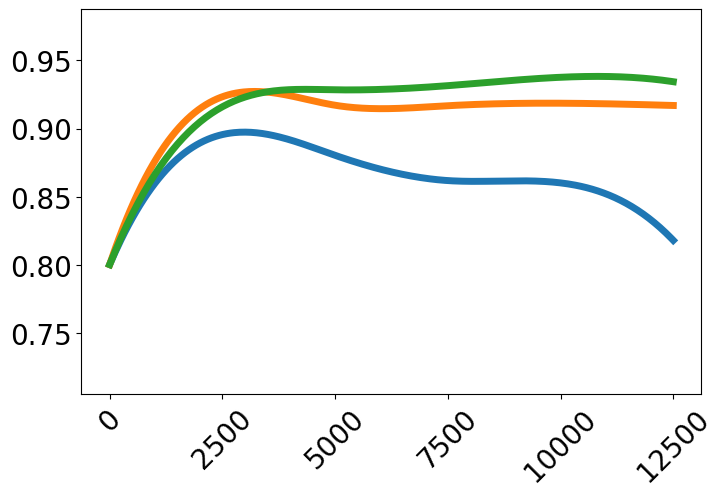} & \includegraphics[width=0.3\textwidth]{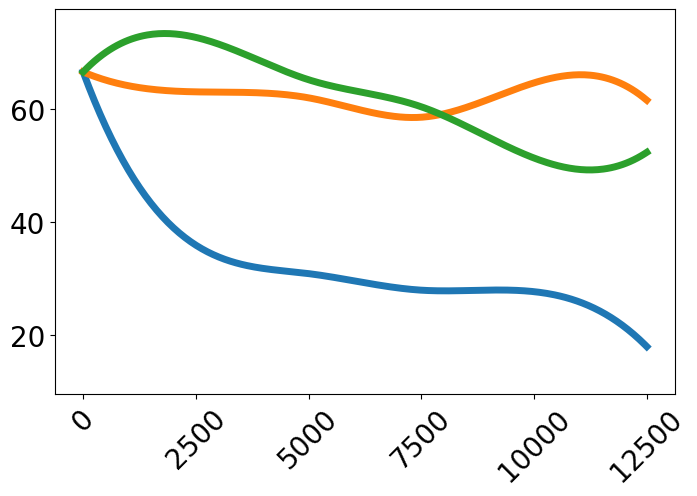} \\
    IWL score: 8B-En-Lt-ID & ICL score: 8B-En-Lt-ID & Copy score: 8B-En-Lt-ID \\[0.5em]
\end{tabular}
\end{figure*}

\begin{figure*}[!htbp]
\centering
\begin{tabular}{ccc}
    \includegraphics[width=0.3\textwidth]{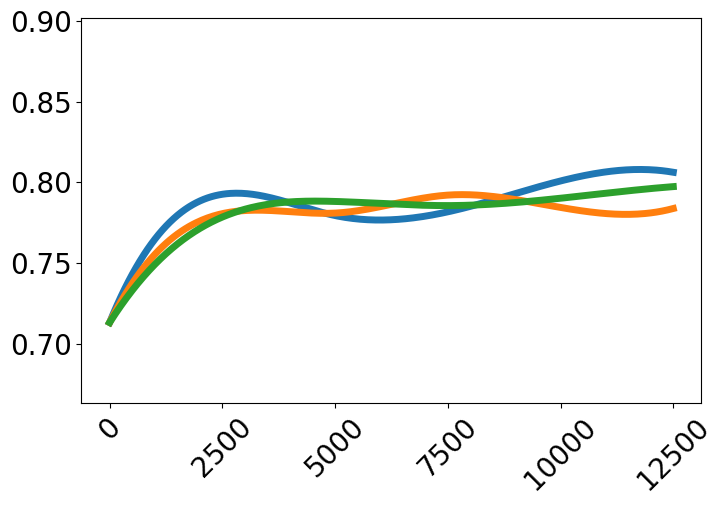} & \includegraphics[width=0.3\textwidth]{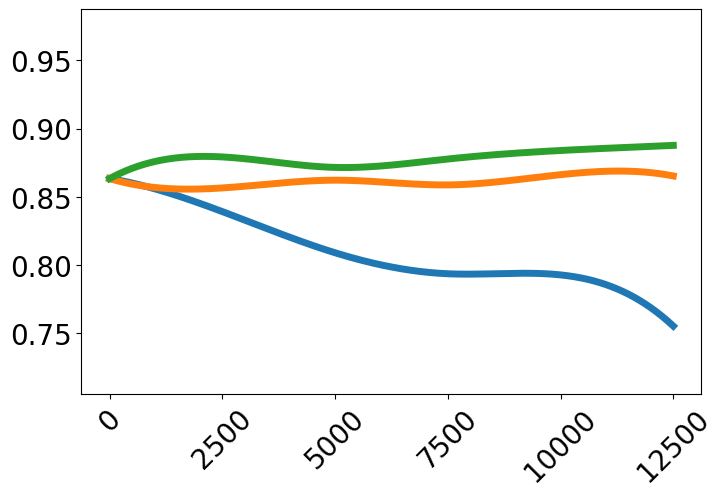} & \includegraphics[width=0.3\textwidth]{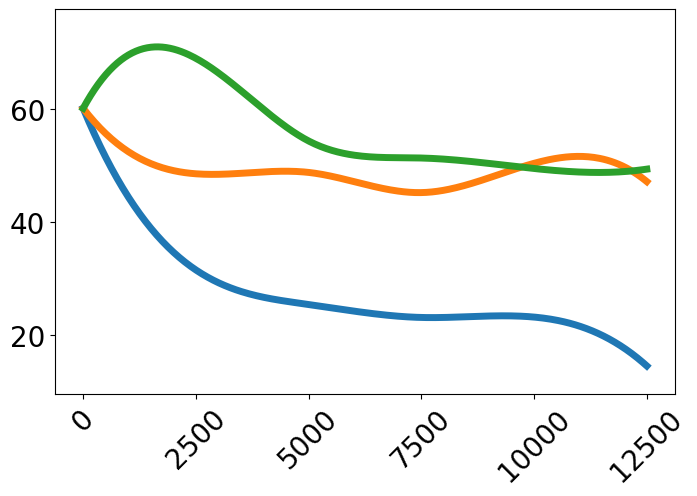} \\
    IWL score: 8B-En-Lt-OOD & ICL score: 8B-En-Lt-OOD & Copy score: 8B-En-Lt-OOD \\[0.5em]
\end{tabular}
\end{figure*}

\begin{figure*}[!htbp]
\centering
\begin{tabular}{ccc}
    \includegraphics[width=0.3\textwidth]{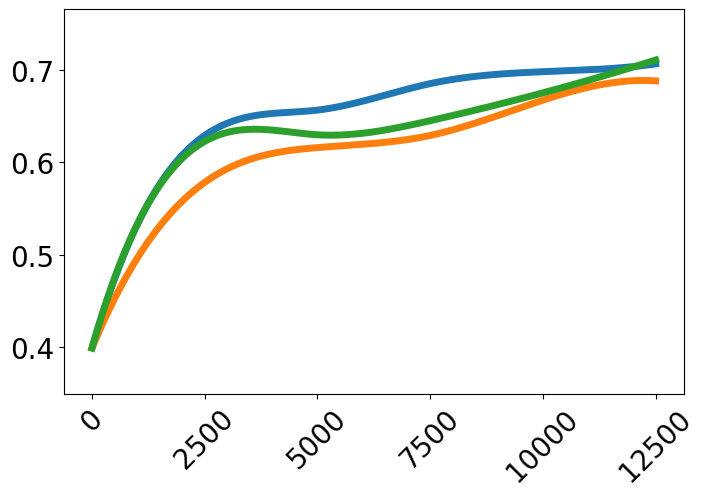} & \includegraphics[width=0.3\textwidth]{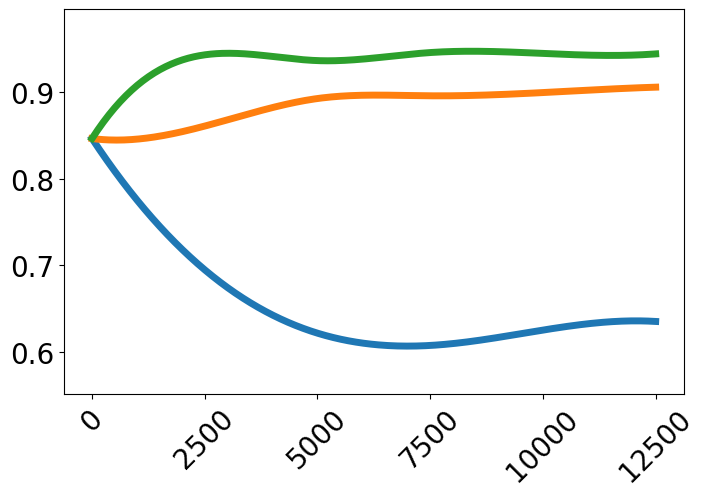} & \includegraphics[width=0.3\textwidth]{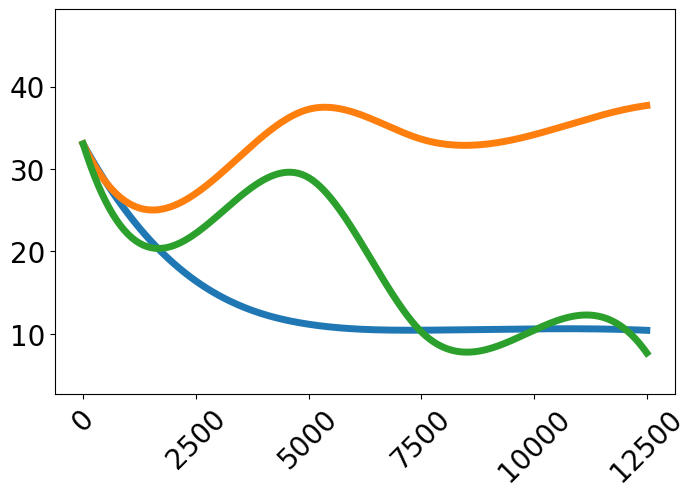} \\
    IWL score: Mistral-En-Lt-ID & ICL score: Mistral-En-Lt-ID & Copy score: Mistral-En-Lt-ID \\[0.5em]
\end{tabular}
\end{figure*}

\begin{figure*}[!htbp]
\centering
\begin{tabular}{ccc}
    \includegraphics[width=0.3\textwidth]{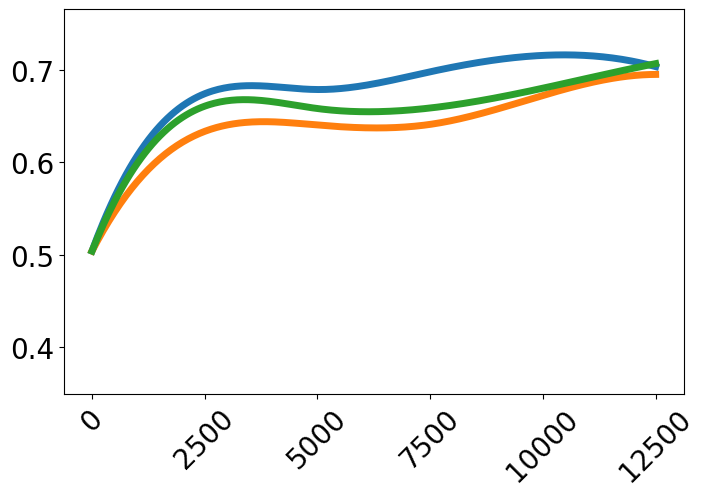} & \includegraphics[width=0.3\textwidth]{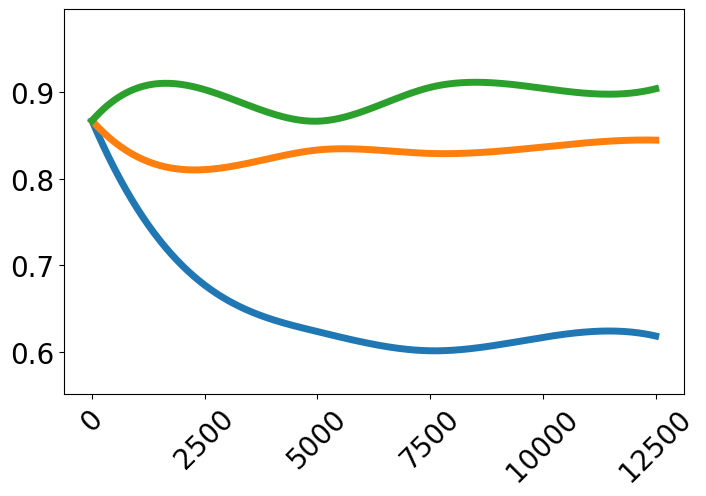} & \includegraphics[width=0.3\textwidth]{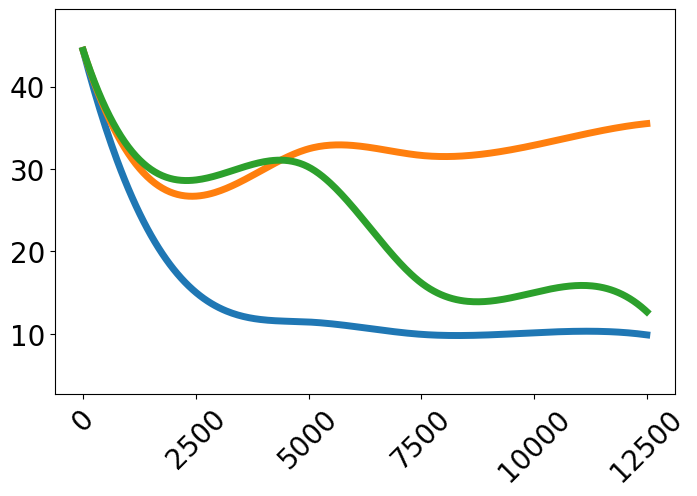} \\
    IWL score: Mistral-En-Lt-OOD & ICL score: Mistral-En-Lt-OOD & Copy score: Mistral-En-Lt-OOD \\[0.5em]
\end{tabular}
\end{figure*}

\begin{figure*}[!htbp]
\centering
\begin{tabular}{ccc}
    \includegraphics[width=0.3\textwidth]{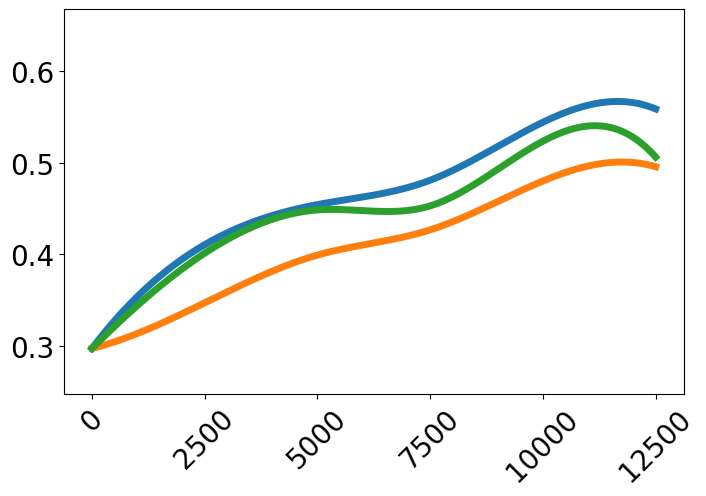} & \includegraphics[width=0.3\textwidth]{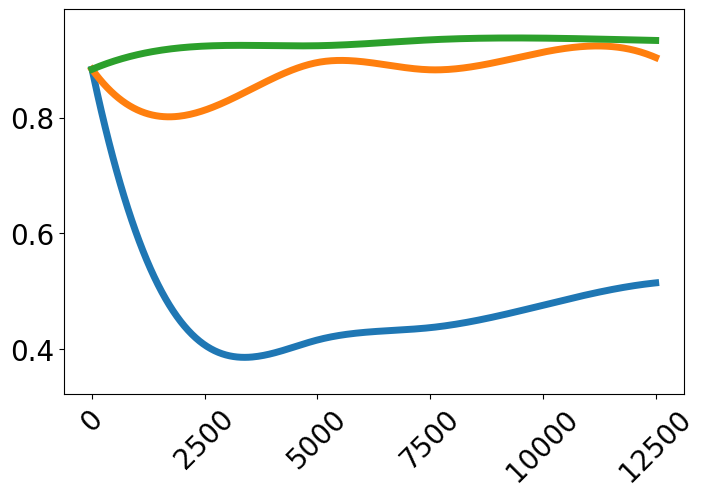} & \includegraphics[width=0.3\textwidth]{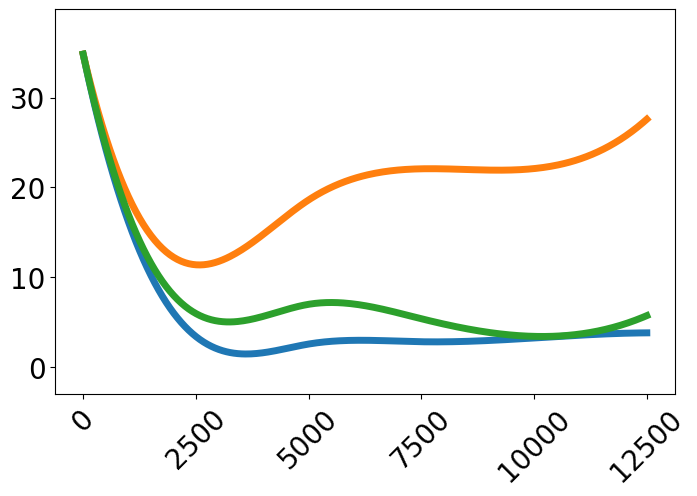} \\
    IWL score: Mistral-En-Ta-ID & ICL score: Mistral-En-Ta-ID & Copy score: Mistral-En-Ta-ID \\[0.5em]
\end{tabular}
\end{figure*}

\begin{figure*}[!htbp]
\centering
\begin{tabular}{ccc}
    \includegraphics[width=0.3\textwidth]{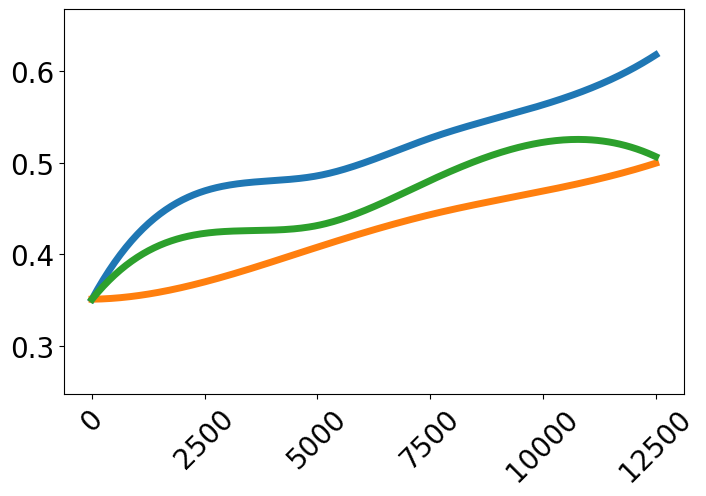} & \includegraphics[width=0.3\textwidth]{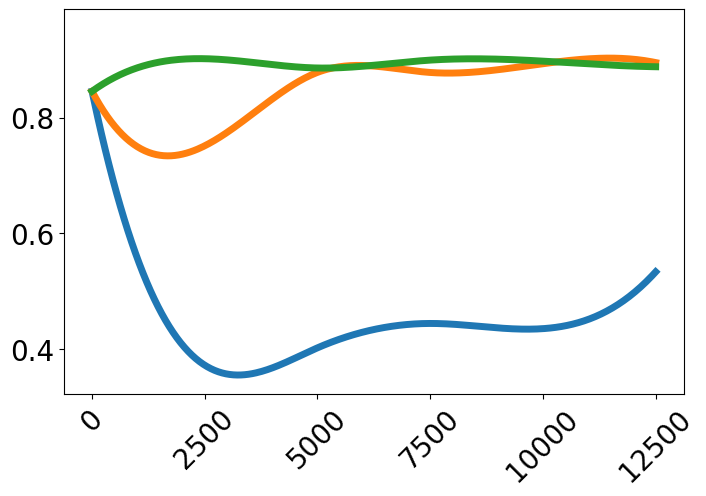} & \includegraphics[width=0.3\textwidth]{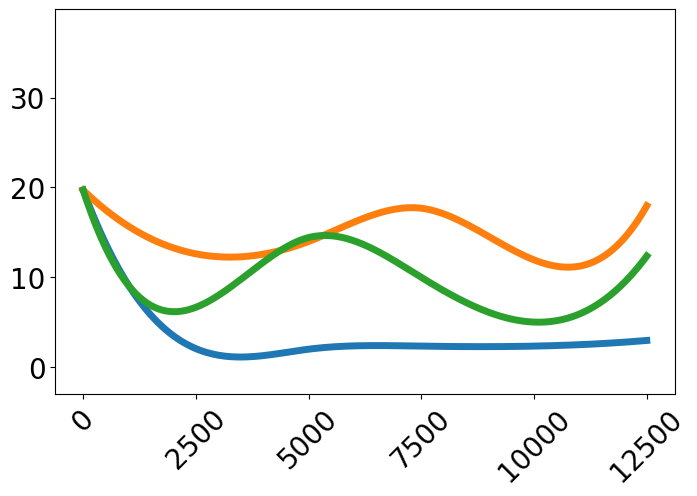} \\
    IWL score: Mistral-En-Ta-OOD & ICL score: Mistral-En-Ta-OOD & Copy score: Mistral-En-Ta-OOD \\[0.5em]
\end{tabular}
\end{figure*}

\begin{figure*}[!htbp]
\centering
    \includegraphics[width=0.8\textwidth]{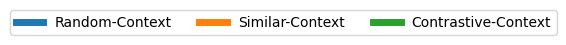}
    \caption{\label{fig:additonal_learning_dynamics} Learning dynamics plots for other models and datasets. X-axis is training steps and Y-axis denotes scores of one of the three probes.}
\end{figure*}

\newpage
\subsection{Learning Dynamics (Unsmoothed)}
\label{sec:appendix:fig3_raw}
\begin{figure*}[!h]
\centering
\begin{tabular}{ccc}
    \includegraphics[width=0.32\textwidth,height=0.16\textwidth]{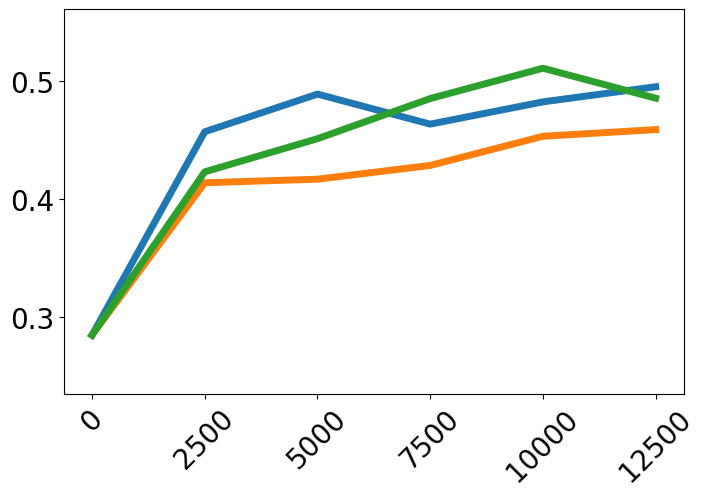} & \includegraphics[width=0.32\textwidth,height=0.16\textwidth]{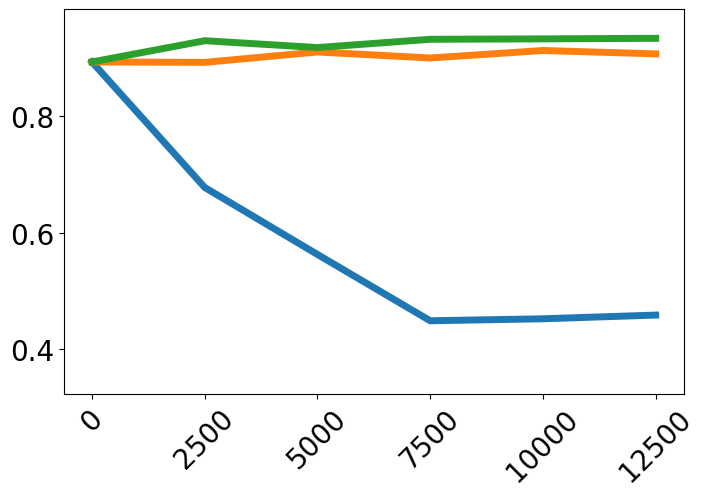} & \includegraphics[width=0.32\textwidth,height=0.16\textwidth]{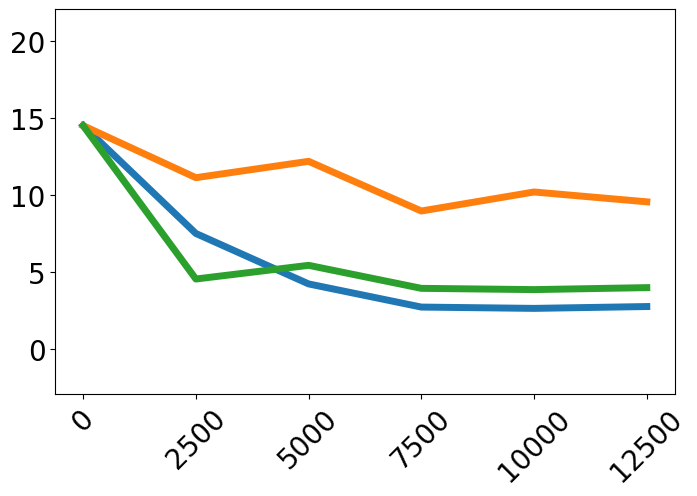} \\
    \iwlscore\ Llama1B-En-Ta-ID & \iclscore\ Llama1B-En-Ta-ID & \copyscore\ Llama1B-En-Ta-ID \\
    \includegraphics[width=0.32\textwidth,height=0.16\textwidth]{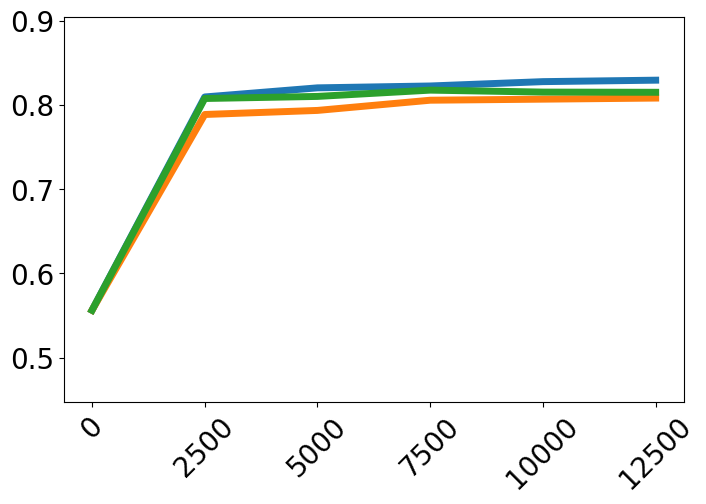} & \includegraphics[width=0.32\textwidth,height=0.16\textwidth]{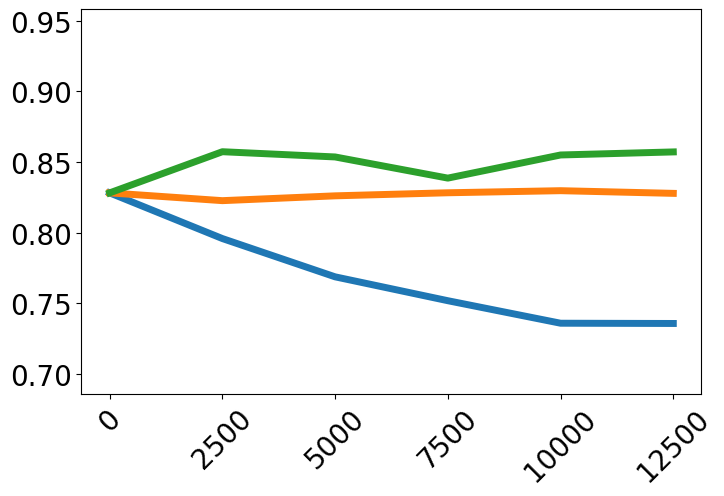} & \includegraphics[width=0.32\textwidth,height=0.16\textwidth]{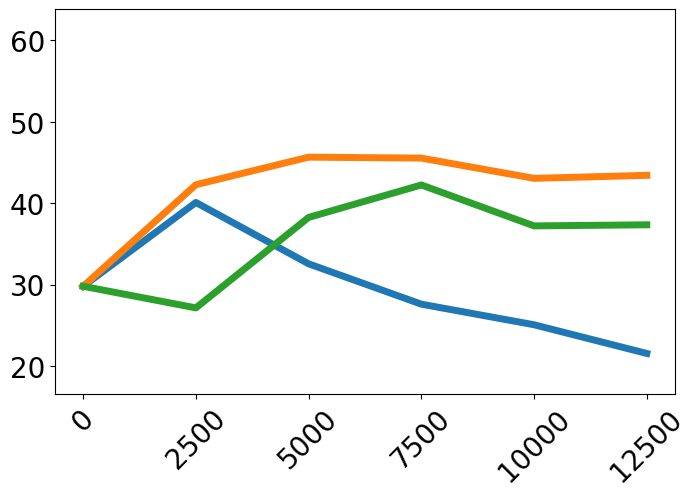} \\
    \iwlscore\ Llama1B-En-De-OOD & \iclscore\ Llama1B-En-De-OOD & \copyscore\ Llama1B-En-De-OOD \\
    \includegraphics[width=0.32\textwidth,height=0.16\textwidth]{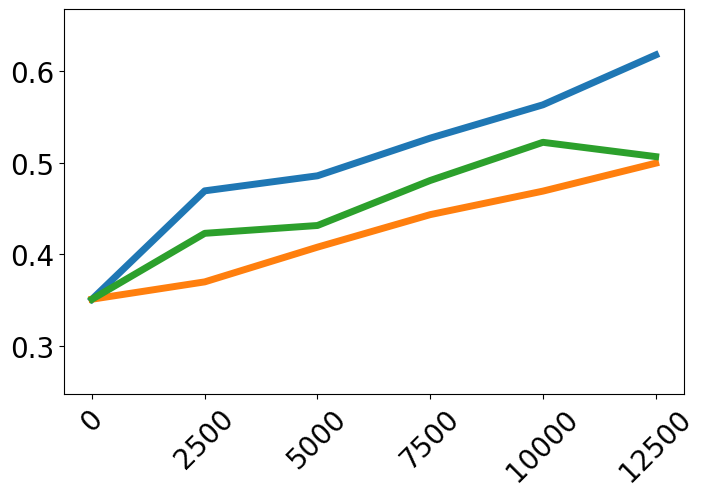} & \includegraphics[width=0.32\textwidth,height=0.16\textwidth]{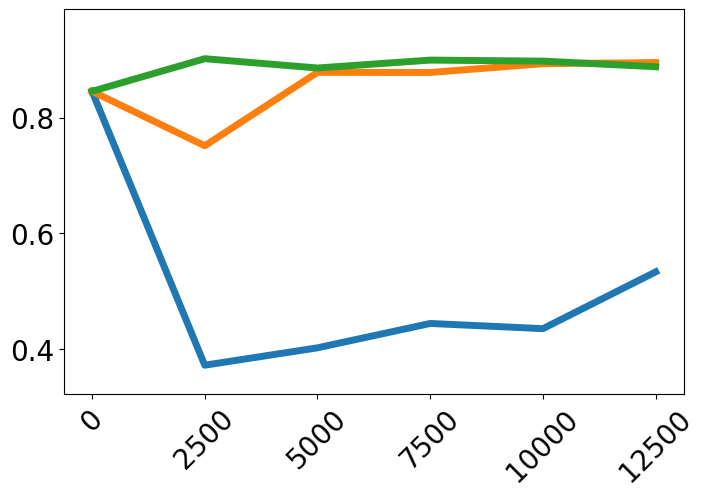} & \includegraphics[width=0.32\textwidth,height=0.16\textwidth]{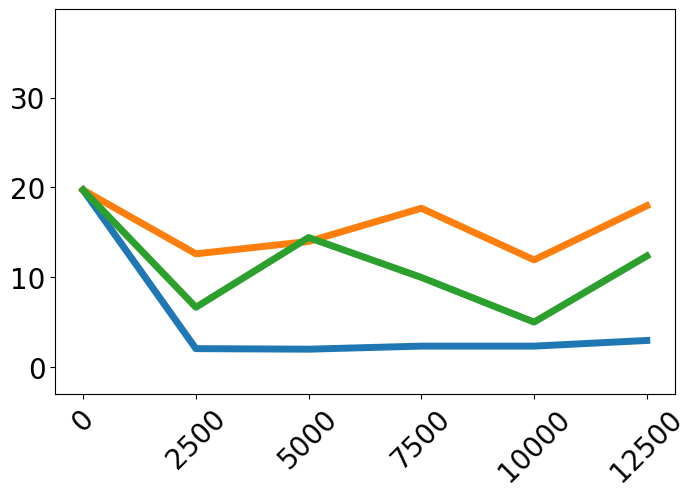} \\
    \iwlscore\ Mistral-En-Ta-OOD & \iclscore\ Mistral-En-Ta-OOD & \copyscore\ Mistral-En-Ta-OOD \\
   \multicolumn{3}{c}{\includegraphics[width=0.8\textwidth]{figures/training_dynamics_legend.png}} \\
\end{tabular}
   \caption{\label{fig:} Unsmoothed learning dynamics plots for a few models and datasets. X-axis is training steps and Y-axis denotes scores of one of the three probes.}
\end{figure*}

\newpage

\end{document}